\documentclass[10pt,journal,compsoc,x11names]{IEEEtran}

\DeclareUnicodeCharacter{FF0C}{,}
\usepackage{amsmath, amsfonts}
\usepackage{algorithmic}
\usepackage{algorithm}
\usepackage{array}
\usepackage[caption=false,font=normalsize,labelfont=sf,textfont=sf]{subfig}
\usepackage{textcomp}
\usepackage{stfloats}
\usepackage{url}
\usepackage{verbatim}
\usepackage{graphicx}
\usepackage{cite}
\usepackage{multirow}
\usepackage[T1]{fontenc}
\usepackage[utf8]{inputenc}

\usepackage{hyperref}
\hypersetup{
    colorlinks=true,      
    linkcolor=blue,       
    citecolor=red,        
    filecolor=magenta,    
    urlcolor=blue        
}

\usepackage{forest}
\usetikzlibrary{shadows}
\usepackage{amsmath}

\usepackage{amssymb}
\usepackage{cleveref}

\definecolor{deepred}{rgb}{0.698,0.133,0.133}
\definecolor{blue}{rgb}{0,0,1}
\usepackage{tikz}
\usepackage{amssymb}
\newcommand\encircle[2][]{\tikz[overlay]\node[fill=blue!20,inner sep=2pt, anchor=text, rectangle, rounded corners=1.5mm,#1] {#2};\phantom{#2}}

\definecolor{lightcoral}{rgb}{0.94, 0.5, 0.5}
\definecolor{lightgreen}{rgb}{0.56, 0.93, 0.56}
\definecolor{harvestgold}{rgb}{0.85, 0.57, 0.0}
\definecolor{brightlavender}{rgb}{0.75, 0.58, 0.89}
\definecolor{capri}{rgb}{0.0, 0.75, 1.0}
\definecolor{carminepink}{rgb}{0.92, 0.3, 0.26}
\definecolor{celadon}{rgb}{0.67, 0.88, 0.69}
\definecolor{darkpastelgreen}{rgb}{0.01, 0.75, 0.24}

\definecolor{deepred}{rgb}{0.698,0.133,0.133}
\definecolor{blue}{rgb}{0,0,1}
\definecolor{lightergray}{RGB}{230,230,230}
\definecolor{DarkRed}{RGB}{130,25,0}
\definecolor{PurpleRed}{RGB}{204,0,102}
\definecolor{DarkGreen}{RGB}{30,130,30}
\definecolor{DarkBlue}{RGB}{0,0,250}
\definecolor{DarkYellow}{RGB}{255,128,0}
\definecolor{light-gray}{gray}{0.95}
\definecolor{lightgreen}{RGB}{231,255,219}
\definecolor{lightred}{RGB}{252,231,234}
\definecolor{lightyellow}{RGB}{250,253,191}
\definecolor{lightpurple}{RGB}{229,204,255}
\definecolor{lightblue}{RGB}{229,246,254}
\definecolor{value-modification}{RGB}{250, 217, 86}
\definecolor{digit-expansion}{RGB}{216, 194, 104}
\definecolor{integer-decimal-fraction}{RGB}{240, 133, 51}
\definecolor{semantic-paraphrasing}{RGB}{85, 157, 63}
\definecolor{complexity-increasing}{RGB}{58, 120, 175}
\definecolor{question-transformation}{RGB}{174, 205, 225}
\definecolor{interference-injection}{RGB}{255,204,229}
\definecolor{remove-constrain}{RGB}{204,204,255}
\definecolor{myGreen}{RGB}{127,210,85}
\definecolor{myOrange}{RGB}{242,154,66}
\definecolor{myYellow}{RGB}{247,223,65}
\definecolor{myRed}{RGB}{232,80,43}
\definecolor{myViolet}{RGB}{162,57,102}
\definecolor{myBlue}{HTML}{4686f3}
\definecolor{myYellowv2}{HTML}{E6C802}
\definecolor{myOrangev2}{HTML}{ED8E55}
\definecolor{MyGreenv2}{HTML}{009B55}
\definecolor{MyRedv2}{HTML}{c22f2f}

\definecolor{lightcoral}{rgb}{0.94, 0.5, 0.5}
\definecolor{lightgreen}{rgb}{0.56, 0.93, 0.56}
\definecolor{harvestgold}{rgb}{0.85, 0.57, 0.0}
\definecolor{brightlavender}{rgb}{0.75, 0.58, 0.89}
\definecolor{capri}{rgb}{0.0, 0.75, 1.0}
\definecolor{carminepink}{rgb}{0.92, 0.3, 0.26}
\definecolor{celadon}{rgb}{0.67, 0.88, 0.69}
\definecolor{darkpastelgreen}{rgb}{0.01, 0.75, 0.24}

\definecolor{deepred}{rgb}{0.698,0.133,0.133}
\definecolor{blue}{rgb}{0,0,1}

\crefformat{section}{\S#2#1#3} 
\crefformat{subsection}{\S#2#1#3}
\crefformat{subsubsection}{\S#2#1#3}

\usepackage{amsthm}
\usepackage{amsmath}%
\usepackage{MnSymbol}%

\makeatletter
\newtheorem*{rep@theorem}{\rep@title}
\newcommand{\newreptheorem}[2]{%
\newenvironment{rep#1}[1]{%
 \def\rep@title{#2 \ref{##1}}%
 \begin{rep@theorem}}%
 {\end{rep@theorem}}}
\makeatother

\newreptheorem{theorem}{Theorem}

\newreptheorem{lemma}{Lemma}

\newreptheorem{proposition}{proposition}

\usepackage{bm}

\usepackage{booktabs}
\usepackage{xcolor,pifont}
\usepackage{bbding}
\usepackage{colortbl}

\usepackage[inline,shortlabels]{enumitem}

\usepackage{tikz,xcolor,hyperref}
\definecolor{lime}{HTML}{A6CE39}
\DeclareRobustCommand{\orcidicon}{%
    \begin{tikzpicture}
    \draw[lime, fill=lime] (0,0) 
    circle [radius=0.16] 
    node[white] {{\fontfamily{qag}\selectfont \tiny ID}};    \draw[white, fill=white] (-0.0625,0.095) 
    circle [radius=0.007];    \end{tikzpicture}
    \hspace{-2mm}}
\foreach \x in {A, ..., Z}{%
    \expandafter\xdef\csname orcid\x\endcsname{\noexpand\href{https://orcid.org/\csname orcidauthor\x\endcsname}{\noexpand\orcidicon}}
}

\begin{document}

\title{From System 1 to System 2: A Survey of\\ Reasoning Large Language Models}

\author{
Zhong-Zhi~Li$^{*}$,
Duzhen~Zhang$^{*}$,
Ming-Liang~Zhang$^{\S}$,
Jiaxin~Zhang$^{\S}$,
Zengyan~Liu$^{\S}$,
Yuxuan~Yao$^{\S}$,\\
Haotian~Xu, 
Junhao~Zheng, 
Pei-Jie Wang, 
Xiuyi~Chen,
Yingying~Zhang,
Fei~Yin, 
Jiahua~Dong, \\
Zhiwei Li, Bao-long Bi, Ling-rui Mei, Jun-Feng Fang, Xiao Liang \\
Zhijiang~Guo$^{\dagger}$,
Le~Song$^{\dagger}$,
Cheng-Lin~Liu$^{\dagger}$\orcidA{},~\IEEEmembership{Fellow,~IEEE}
\thanks{Version: v1 (major update on February 24, 2025)}
\thanks{$^{*}$Core contribution. $^{\S}$Significant contribution. $^\dagger$Corresponding author.}
\thanks{Duzhen Zhang, Jiahua Dong, and Le Song are with the Mohamed bin Zayed University of Artificial Intelligence, Abu Dhabi, UAE (E-mail: bladedancer957@gmail.com; dongjiahua1995@gmail.com; le.song@mbzuai.ac.ae).}
\thanks{Zhong-Zhi Li, Pei-Jie Wang, Bao-Long Bi, Ling-Rui Mei, Xiuyi Chen, Fei Yin, and Cheng-Lin Liu are with the Institute of Automation, Chinese Academy of Sciences, Beijing, China (E-mail: lizhongzhi2022@ia.ac.cn; wangpeijie2023@ia.ac.cn, bibaolong23z@ict.ac.cn, meilingrui25b@ict.ac.cn hugheren.chan@gmail.com; fyin@nlpr.ia.ac.cn; liucl@nlpr.ia.ac.cn).}
\thanks{Ming-Liang Zhang is with the Alibaba Group, Beijing, China (E-mail: luxing.zml@taobao.com).}
\thanks{Bao-Long Bi, Ling-Rui Mei are with the Chinese Academy of Sciences, Beijing, China (E-mail: bibaolong23z@ict.ac.cn, meilingrui25b@ict.ac.cn).}
\thanks{Xiao Liang is with the University of California，Los Angles, USA (E-mail: luxing.zml@taobao.com).}
\thanks{Zengyan Liu, Yuxuan Yao, Zhiwei Li, and Zhijiang Guo are with City University of Hong Kong and the Hong Kong University of Science and Technology (Guangzhou), China (E-mail: zengyaliu2-c@my.cityu.edu.hk; yuxuanyao3-c@my.cityu.edu.hk; zhiweil466@gmail.com; zhijiangguo@hkust-gz.edu.cn).}
\thanks{Jiaxin Zhang is with the University of Strathclyde, Glasgow, UK (E-mail: jiaxin.zhang@strath.ac.uk).}

\thanks{Haotian Xu is with the Xiaohongshu Inc, Beijing, China (E-mail: xuhaotian@xiaohongshu.com).}
\thanks{Yingying Zhang is with the East China Normal University, Shanghai, China (E-mail: yyzhang@fem.ecnu.edu.cn).}
\thanks{Junfeng Fang is with the Nanyang Technological University, Singapore (E-mail: junhaozheng47@outlook.com).}
\thanks{Junhao Zheng is with the South China University of Technology, Guangzhou, China (E-mail: junhaozheng47@outlook.com).}
}

\markboth{Journal of \LaTeX\ Class Files, January 2025}%
{Shell \MakeLowercase{\textit{et al.}}: A Sample Article Using IEEEtran.cls for IEEE Journals}


\IEEEtitleabstractindextext{

\begin{abstract}
Achieving human-level intelligence requires refining the transition from the fast, intuitive System 1 to the slower, more deliberate System 2 reasoning. 
While System 1 excels in quick, heuristic decisions, System 2 relies on logical reasoning for more accurate judgments and reduced biases. 
Foundational Large Language Models (LLMs) excel at fast decision-making but lack the depth for complex reasoning, as they have not yet fully embraced the step-by-step analysis characteristic of true System 2 thinking. 
Recently, reasoning LLMs like OpenAI's o1/o3 and DeepSeek's R1 have demonstrated expert-level performance in fields such as mathematics and coding, closely mimicking the deliberate reasoning of System 2 and showcasing human-like cognitive abilities. 
This survey begins with a brief overview of the progress in foundational LLMs and the early development of System 2 technologies, exploring how their combination has paved the way for reasoning LLMs. 
Next, we discuss how to construct reasoning LLMs, analyzing their features, the core methods enabling advanced reasoning, and the evolution of various reasoning LLMs. 
Additionally, we provide an overview of reasoning benchmarks, offering an in-depth comparison of the performance of representative reasoning LLMs. 
Finally, we explore promising directions for advancing reasoning LLMs and maintain a real-time \href{https://github.com/zzli2022/Awesome-Slow-Reason-System}{{GitHub Repository}} to track the latest developments. 
We hope this survey will serve as a valuable resource to inspire innovation and drive progress in this rapidly evolving field.



\end{abstract}






}

\maketitle


\section{Introduction}
\label{sec:introduction}

\begin{flushleft}
\leftskip=1cm\emph{``Don't teach. Incentivize.''} \\
\vspace{.3em}
\leftskip=4.55cm---\emph{Hyung Won Chung, OpenAI}
\end{flushleft}

\IEEEPARstart{A}{chieving} human-level intelligence requires refining the transition from \textit{System 1} to \textit{System 2} reasoning \cite{hua2022system, wei2022chain, wangself, zhouleast, zelikman2024star}. 
Dual-system theory suggests that human cognition operates through two modes: \textit{System 1}, which is fast, automatic, and intuitive, enabling quick decisions with minimal effort, and \textit{System 2}, which is slower, more analytical, and deliberate \cite{evans1984heuristic, kahneman2003maps}. 
While \textit{System 1} is efficient for routine tasks, it is prone to cognitive biases, especially in complex or uncertain situations, leading to judgment errors. 
In contrast, \textit{System 2} relies on logical reasoning and systematic thinking, resulting in more accurate and rational decisions \cite{huang2023towards, qiao2023reasoning, wang2023towards, shaikh2023second}. 
By mitigating the biases of \textit{System 1}, \textit{System 2} provides a more refined approach to problem-solving \cite{shao2024visual, zhangautomatic, hao2023reasoning, zhang2023meta}.


The development of foundational Large Language Models (LLMs)\footnote{In this paper, ``reasoning'' refers to answering questions involving complex, multi-step processes with intermediate steps. \textbf{Foundational LLMs:} LLMs with basic reasoning abilities, handling simple or single-step tasks. \textbf{Reasoning LLMs:} LLMs that excel in complex tasks like coding and mathematical proofs, incorporating a ``thinking'' process\textendash tasks that foundational LLMs struggle with.} has marked a major milestone in Artificial Intelligence (AI). 
Models such as GPT-4o \cite{gpt4o-0513} and DeepSeek-v3 \cite{liu2024deepseek} have demonstrated impressive capabilities in text generation, language translation, and a variety of perception tasks \cite{vaswani2017attention, DBLP:conf/naacl/DevlinCLT19, DBLP:journals/corr/abs-1907-11692, radford2018improving, radford2019language, brown2020language, ouyang2022training, touvron2023llama, zhao2023survey, liu2023llava, DBLP:conf/acl/ZhangY0L0C024}. 
These models, trained on extensive datasets and utilizing advanced algorithms, excel in understanding and generating human-like responses. 
However, despite their impressive achievements, foundational LLMs operate in a manner similar to \textit{System 1} reasoning, relying on fast, heuristic-driven decision-making. 
While they perform exceptionally well in providing rapid responses, they often fall short in scenarios requiring deep, logical analysis and precision in complex reasoning tasks. 
This limitation becomes especially clear in situations involving intricate problem-solving, logical analysis, or nuanced understanding, where these models do not yet match human cognitive abilities.

\begin{figure*}[t]
    \centering
    \includegraphics[width=0.94\linewidth]{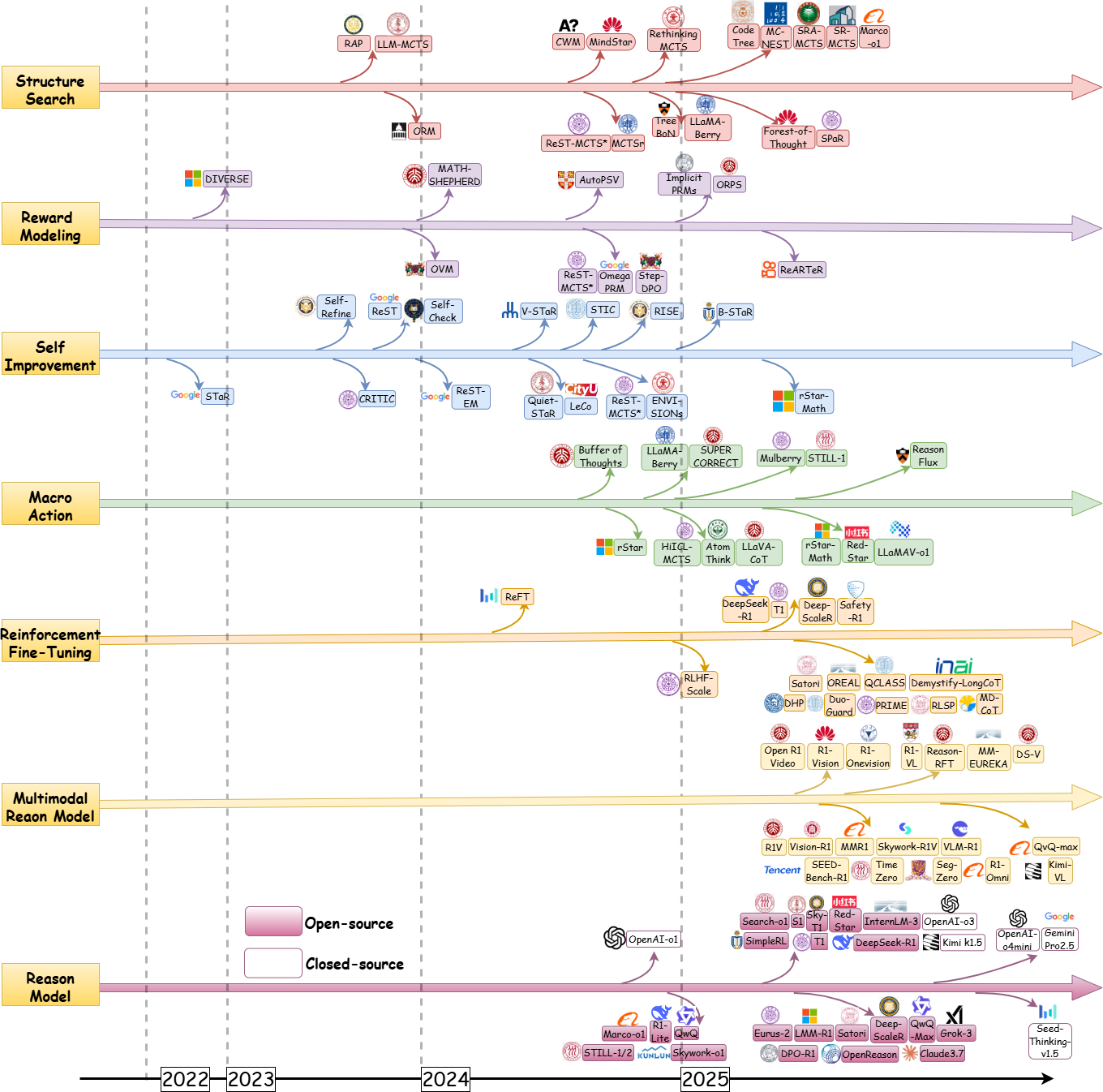}
    \caption{The recent timeline of reasoning LLMs, covering core methods and the release of open-source and closed-source reproduction projects.}
        \label{fig:timeline}
\end{figure*}

In contrast, reasoning LLMs represent a significant advancement in the evolution of language models. 
Models like OpenAI's o1/o3 \cite{openai_o1, o3-mini} and DeepSeek's R1 \cite{Deepseek-R1} are designed to emulate the slower, more deliberate reasoning associated with \textit{System 2} thinking. 
Unlike foundational LLMs, reasoning LLMs are equipped with mechanisms for processing information step-by-step, allowing them to make more accurate and rational decisions. 
This shift from fast-thinking, intuitive processes to more methodical, reasoning-driven models enables reasoning LLMs to tackle complex tasks, such as advanced mathematics \cite{cobbe2021training, kojima2022large, liu2023improving, zhu2023solving, ludynamic, lightmanlet}, logical reasoning \cite{yao2023thinking, yao2023beyond, wen2023mindmap, lei2023boosting, jin2024impact, besta2024graph, cheng2024self}, and multimodal reasoning \cite{you2023idealgpt, wu2024v, chen2024genome}, with expert-level performance, exhibiting human-like cognitive abilities. 
As a result, reasoning LLMs are increasingly seen as capable of achieving the deep, logical thinking needed for tasks that were once considered beyond AI's reach. 
The recent timeline of reasoning LLMs is presented in Figure \ref{fig:timeline}.


\begin{figure*}[t]
\centering
\resizebox{0.95\textwidth}{!}{\tikzset{
        my node/.style={
            draw,
            align=left,
            thin,
            text width=2.5cm, 
            rounded corners=3,
        },
        my leaf/.style={
            draw,
            align=left,
            thin,
            text width=4.5cm, 
            rounded corners=3,
        }
}
\forestset{
  every leaf node/.style={
    if n children=0{#1}{}
  },
  every tree node/.style={
    if n children=0{minimum width=1em}{#1}
  },
}
\begin{forest}
    for tree={%
        every leaf node={my leaf, font=},
        every tree node={my node, font=\small, l sep-=4.5pt, l-=1.pt},
        anchor=west,
        inner sep=2pt,
        l sep=10pt, 
        s sep=5pt, 
        fit=tight,
        grow'=east,
        edge={thick},
        parent anchor=east,
        child anchor=west,
        if n children=0{tier=last}{},
        edge path={
            \noexpand\path [draw, \forestoption{edge}] (!u.parent anchor) -- +(5pt,0) |- (.child anchor)\forestoption{edge label};
        },
        if={isodd(n_children())}{
            for children={
                if={equal(n,(n_children("!u")+1)/2)}{calign with current}{}
            }
        }{}
    }
    [{Organizational Structure}, draw=gray, color=gray!100, fill=gray!15, very thick, text=black, text width=2.1cm,
        [\cref{early_srs} Foundations of Reasoning LLMs, color=cyan!100, fill=cyan!15, very thick, text=black, text width=5.1cm
            [\cref{f_llm} Foundational LLMs, color=cyan!100, fill=cyan!15, very thick, text=black, text width=4.5cm
            ]
            [\cref{symb_exp} Symbolic Logic Systems, color=cyan!100, fill=cyan!15, very thick, text=black, text width=4.5cm
            ]
            [\cref{mcts} Monte Carlo Tree Search, color=cyan!100, fill=cyan!15, very thick, text=black, text width=4.5cm
            ]
            [\cref{rl} Reinforcement Learning, color=cyan!100, fill=cyan!15, very thick, text=black, text width=4.5cm
            ]
        ]
        [\cref{replication} Blueprinting Reasoning LLMs, color=lightcoral!100, fill=lightcoral!15, very thick, text=black, text width=5.1cm
            [\cref{o1_features} Feature Analysis, color=lightcoral!100, fill=lightcoral!15, very thick, text=black, text width=4.9cm
                [\cref{output_behaviour} Output Behaviour, color=lightcoral!100, fill=lightcoral!15, very thick, text=black, tier=Task, text width=5.1cm
                ]
                [\cref{dynamic_perspective} Training Dynamics, color=lightcoral!100, fill=lightcoral!15, very thick, text=black, tier=Task, text width=5.1cm
                ]
            ]
            [\cref{foundations} Core Method, color=lightcoral!100, fill=lightcoral!15, very thick, text=black, text width=4.9cm
                [\cref{structure_search} Structure Search, color=lightcoral!100, fill=lightcoral!15, very thick, text=black, tier=Task, text width=5.1cm
                ]
                [\cref{prm} Reward Modeling, color=lightcoral!100, fill=lightcoral!15, very thick, text=black, tier=Task, text width=5.1cm
                ]
                [\cref{self-improve} Self Improvement, color=lightcoral!100, fill=lightcoral!15, very thick, text=black, tier=Task, text width=5.1cm
                ]
                [\cref{macro_action} Macro Action, color=lightcoral!100, fill=lightcoral!15, very thick, text=black, tier=Task, text width=5.1cm
                ]
                [\cref{rl_supervise} Reinforcement Fine-Tuning, color=lightcoral!100, fill=lightcoral!15, very thick, text=black, tier=Task, text width=5.1cm
                ]
            ]
            [\cref{evolutionary} Reasoning LLMs Evolution, color=lightcoral!100, fill=lightcoral!15, very thick, text=black, text width=4.9cm
            ]
        ]
        [\cref{benchmark} Benchmarking Reasoning LLMs, color=darkpastelgreen!100, fill=darkpastelgreen!15, very thick, text=black, text width=5.1cm
            [\cref{benchmark_category} Benchmark Categories, color=darkpastelgreen!100, fill=darkpastelgreen!15, very thick, text=black, text width=4.55cm
            ]
            [\cref{metrics} Evaluation Metrics, color=darkpastelgreen!100, fill=darkpastelgreen!15, very thick, text=black, text width=4.55cm
            ]
            [\cref{performance_compare} Performance Comparison, color=darkpastelgreen!100, fill=darkpastelgreen!15, very thick, text=black, text width=4.55cm
            ]
        ]
        [\cref{extend_techniques} Extended Techniques, color=yellow!100, fill=yellow!20, very thick, text=black, text width=5.1cm
            [\cref{advanced_infra} Advanced Infrastructure, color=yellow!100, fill=yellow!20, very thick, text=black, text width=4.55cm
            ]
            [\cref{srs_safety_of_lrm} Trustworthiness of Reasoning LLMs, color=yellow!100, fill=yellow!20, very thick, text=black, text width=4.55cm
            ]
            [\cref{agent_reason} Reasoning LLM as Agent, color=yellow!100, fill=yellow!20, very thick, text=black, text width=4.55cm
            ]
            [\cref{efficient_lrm} Efficient Reasoning LLM, color=yellow!100, fill=yellow!20, very thick, text=black, text width=4.55cm
            ]
        ]
    ]
\end{forest}}
\caption{The primary organizational structure of the survey.}
\label{fig:structure}
\end{figure*}
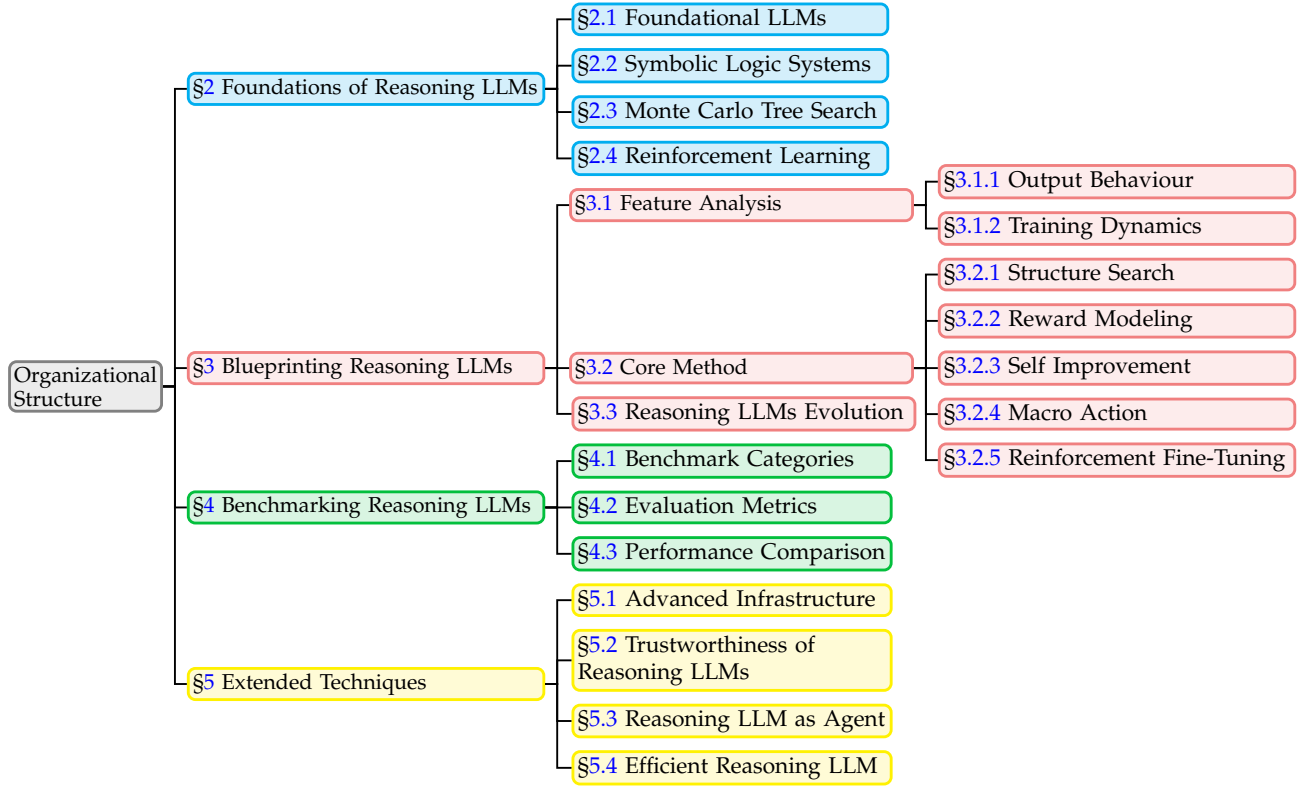

\subsection{Structure of the Survey}

This survey offers a comprehensive overview of the key concepts, methods, and challenges involved in the development of reasoning LLMs. 
As illustrated in Figure \ref{fig:structure}, this survey is organized as follows:
\begin{enumerate}[itemindent=0em]

\item Section \ref{early_srs} offers a concise overview of the progress in foundational LLMs (Section \ref{f_llm}) and the early development of key \textit{System 2} technologies, including symbolic logic systems (Section \ref{symb_exp}), Monte Carlo Tree Search (MCTS) (Section \ref{mcts}), and Reinforcement Learning (RL) (Section \ref{rl}), highlighting how their combination has paved the way for reasoning LLMs. 

\item Section \ref{replication} introduces reasoning LLMs and outlines their construction process. 
Specifically, Section \ref{o1_features} presents the characteristics of reasoning LLMs from two perspectives: output behavior (Section \ref{output_behaviour}) and training dynamics (Section \ref{dynamic_perspective}), emphasizing their differences from foundational LLMs. 
Section \ref{foundations} identifies the core methods necessary for achieving advanced reasoning capabilities, focusing on five aspects: Structure Search (Section \ref{structure_search}), Reward Modeling (Section \ref{prm}), Self Improvement (Section \ref{self-improve}), Macro Action (Section \ref{macro_action}), and Reinforcement Fine-Tuning (Section \ref{rl_supervise}). 
Each section delves into the specific characteristics of these methods and introduces representative reasoning LLMs for each approach. 
Section \ref{evolutionary} traces the evolutionary stages of reasoning LLMs.

\item Section \ref{benchmark} evaluates representative reasoning LLMs. Specifically, Section \ref{benchmark_category} reviews current mainstream reasoning benchmarks, covering both plain text and multimodal benchmarks across various task types. Section \ref{metrics} outlines the current evaluation metrics, while Section \ref{performance_compare} analyzes and compares the performance of mainstream reasoning LLMs with their foundational counterparts based on these benchmarks.

\item Section \ref{extend_techniques} summarizes several recent technical areas related to Reasoning LLMs. Section \ref{advanced_infra} summarizes technologies related to large-scale RL training. Section \ref{srs_safety_of_lrm} discusses several safety issues related to LRM. Section \ref{agent_reason} summarizes the field of integrating Reasoning LLMs with Agents. Section \ref{efficient_lrm} summarizes the technologies of Adaptive Reasoning LLMs and Efficient Reasoning LLMs.

\item Section \ref{future} highlights the limitations of existing reasoning LLMs and outlines several promising future development directions for these models.

\item Finally, we conclude the paper in Section \ref{conclusion} and provide a real-time tracking \href{https://github.com/zzli2022/Awesome-Slow-Reason-System}{{GitHub Repository}} to monitor the latest developments in the field. 

\end{enumerate}
We hope this survey serves as a valuable resource, fostering innovation and progress in this rapidly evolving domain.

\subsection{Contribution of the Survey}


Recently, several analyses and replications of specific technical approaches have been conducted \cite{Compare_o1, TowardsSystem2ReasoninLLM, o1_Journey_Part1, o1_Journey_Part2, huang2025o1, Slow_Thinking_with_LLMs_2, RedStar, Scaling_of_Search_Learning}, yet there remains a lack of systematic analysis and organization. 
Research \cite{Test_Time_Computing} has focused only on slow-thinking methods during testing. 
Meanwhile, studies \cite{besta2025reasoning, zhang2024llm, LargeReasonModel} have primarily concentrated on training or achieving reasoning LLMs, often from the perspective of RL.

Our survey distinguishes itself from and contributes to the existing literature in the following ways:
\begin{enumerate}[itemindent=0em]
\item Rather than focusing on a single technical approach, we offer a comprehensive overview of the key concepts, methods, and challenges involved in reasoning LLMs.

\item We summarize the key advancements of early \textit{System 2} and how they have paved the way for reasoning LLMs, specifically in combination with foundational LLMs\textendash a crucial aspect often overlooked in previous works.

\item We present a more thorough and inclusive summary of the core methods necessary for constructing reasoning LLMs, including but not limited to RL.

\end{enumerate}



\section{Foundations of Reasoning LLMs}\label{early_srs}


In this section, we provide a concise overview of the progress in foundational LLMs and the early development of key \textit{System 2} technologies, highlighting critical advancements that, when combined with foundational LLMs, have paved the way for reasoning LLMs. These advancements include symbolic logic systems, MCTS, and RL.

\subsection{Foundational LLMs}\label{f_llm}

The development of foundational LLMs saw significant advancements with the introduction of pretrained Transformers \cite{vaswani2017attention} in 2018-2019, notably through BERT \cite{DBLP:conf/naacl/DevlinCLT19} and GPT \cite{radford2018improving}. 
These models leveraged unsupervised pretraining on vast text corpora, followed by fine-tuning for task-specific applications. This approach enabled them to develop a broad language understanding before specializing in tasks such as sentiment analysis, entity recognition, and question answering. 
BERT's bidirectional context processing improved word understanding, while GPT excelled in text generation with its unidirectional design.

The release of GPT-2 \cite{radford2019language} in 2019, with 1.5 billion parameters, marked a significant leap in generative performance, though it also raised ethical concerns. 
GPT-3 \cite{brown2020language}, with 175 billion parameters, further demonstrated the power of unsupervised pretraining, excelling in few-shot learning and performing well across a wide range of NLP tasks. 
In subsequent years, multimodal models like CLIP \cite{radford2021learning} and DALL-E \cite{ramesh2021zero} emerged, integrating text and visual inputs. 
These models enabled new tasks, such as generating images from text, and enhanced human-computer interaction.

By 2023-2024, models such as GPT-4/4o \cite{openai2023gpt4,gpt4o-0513}, LLaMA \cite{touvron2023llama}, and LLaVA \cite{liu2023llava} demonstrated advanced capabilities in reasoning, contextual understanding, and multimodal reasoning, processing both text and images \cite{alayrac2022flamingo, DBLP:conf/icml/0008LSH23, DBLP:journals/corr/abs-2305-06500}. 
DeepSeek-V3 \cite{liu2024deepseek}, featuring a 671B Mixture-of-Expert architecture \cite{he2021fastmoe, du2022glam, dai2024deepseekmoe}, outperforms several other LLMs on key benchmarks while offering significant improvements in efficiency and processing speed. 
The evolution of foundational LLMs has revolutionized AI, enabling more sophisticated applications in language comprehension, problem-solving, and human-machine collaboration.

\noindent\textbf{Summary:} The development of foundational LLMs has progressed from pretrained transformers like BERT to multimodal models such as GPT-4, enhancing language understanding, text generation, and image processing. 
This advancement has led to significant breakthroughs in AI, improving language comprehension, problem-solving, and human-computer interaction. Building on deep learning advancements \cite{rumelhart1986learning, lecun1995convolutional, hochreiter1997long, hinton2006fast, hinton2006reducing, hinton2012deep, krizhevsky2012imagenet, DBLP:conf/emnlp/ChoMGBBSB14, sutskever2014sequence, srivastava2014dropout, kingma2014adam, lecun2015deep, he2016deep, huang2017densely,vaswani2017attention, goodfellow2020generative}, foundational LLMs can learn extensive world knowledge and semantic relationships from vast textual or multimodal data. This enables them to exhibit emergent capabilities such as In-Context Learning (ICL) \cite{min2022rethinking, dong2024survey}, prompt engineering \cite{white2023prompt, lester2021power}, and Chain-of-Thought (CoT) reasoning \cite{wei2022chain}, significantly enhancing their adaptability and creative problem-solving abilities.

Despite this progress, foundational LLMs operate similarly to \textit{System 1} reasoning, relying on fast, heuristic-driven decision-making and lacking the step-by-step analysis characteristic of \textit{System 2}. However, their developments lay a solid foundation for future reasoning LLMs\textendash especially when integrated with the following early \textit{System 2} technologies. This combination paves the way for more versatile, flexible, and human-like reasoning models.

\subsection{Symbolic Logic Systems}\label{symb_exp}

Symbolic logic systems mark the earliest phase of AI, utilizing rules and logical principles to represent knowledge and draw conclusions \cite{lewis1959symbolic, carnap2012introduction}. 
They are particularly effective in structured domains, where formal logic ensures precision.

Prolog, a logic programming language based on first-order logic, allows users to define facts, rules, and reason through queries. 
It has been pivotal in symbolic reasoning systems, especially in NLP and expert systems \cite{colmerauer1990introduction, clocksin2003programming, apt1997logic}. Logic-based systems like Prolog employ propositional and predicate logic for formal reasoning \cite{singh1999formal, jeroslow1988computation}. 
From the 1960s to the early 1980s, this approach dominated AI, with systems like IBM's LISP \cite{mccarthy1978history} for symbolic computation and Resolution Theorem Provers \cite{bachmair2001resolution} for automated reasoning. 
In the 1970s, Marvin Minsky introduced Frames, which organized knowledge into structured frameworks, influencing both expert systems and cognitive science \cite{minsky1974framework}.

\noindent\textbf{Summary:} 
Symbolic logic systems were pivotal milestones in early AI development. 
Based on formal logic, they excelled in well-defined problems, particularly in structured environments. 
However, they also exposed the limitations of rigid, rule-based systems. Despite these constraints, symbolic logic remains foundational to the progress of AI.

Recent advancements in reasoning LLMs have greatly enhanced the emulation of human-like \textit{System 2} cognitive processes through sophisticated thought architectures, known as Macro Action frameworks (Section \ref{macro_action}). By combining symbolic templates or rules with foundational LLMs, macro actions have significantly improved their reasoning capabilities. 
Integrating macro actions into foundational LLMs has transformed their ability to handle complex reasoning tasks, as hierarchical planning allows models to make high-level decisions before delving into specific problem details, mirroring symbolic logic's structured approach.

\subsection{Monte Carlo Tree Search}\label{mcts}

MCTS is a simulation-based search algorithm for decision-making and planning \cite{browne2012survey}. 
It constructs a search tree through four steps: \textit{Selection}, which chooses the child node with the highest priority using the UCB1 formula:
\begin{equation}
\footnotesize
\text{UCB1} = \frac{w_i}{n_i} + c \sqrt{\frac{\ln N}{n_i}} \text{, }
\end{equation}
where $w_i$ is the total reward of node $i$, $n_i$ is its visit count, $N$ is the parent node's visit count, and $c$ balances exploration and exploitation. \textit{Expansion} adds new nodes, \textit{Simulation} performs random rollouts to evaluate them, and \textit{Backpropagation} updates node statistics. 
MCTS has been widely used in tasks such as optimizing strategies in board games like Go \cite{gelly2011monte} and in robotic path planning, where it helps robots navigate dynamic environments effectively \cite{swiechowski2023monte}.

\noindent\textbf{Summary:} 
MCTS has played a crucial role in the development of reasoning LLMs, particularly in Structural Search (Section \ref{structure_search}). 
By simulating potential future reasoning paths and backpropagating estimated rewards, MCTS helps foundational LLMs efficiently identify the most promising, high-reward paths. 
This process mirrors human-like planning, where future consequences of decisions are considered before taking action. 
By dynamically exploring multiple reasoning trajectories, MCTS enables models to avoid getting stuck in suboptimal paths, making it easier to navigate complex decision spaces. 
This integration has significantly enhanced the ability of LLMs to handle intricate and dynamic reasoning problems, such as those requiring long-term planning or multi-step logical inferences. 
It has allowed LLMs to make more strategic and informed decisions, improving their overall performance in tasks that involve nuanced reasoning and strategic exploration.




\begin{figure*}[t]
    \centering
    \includegraphics[width=0.95\linewidth]{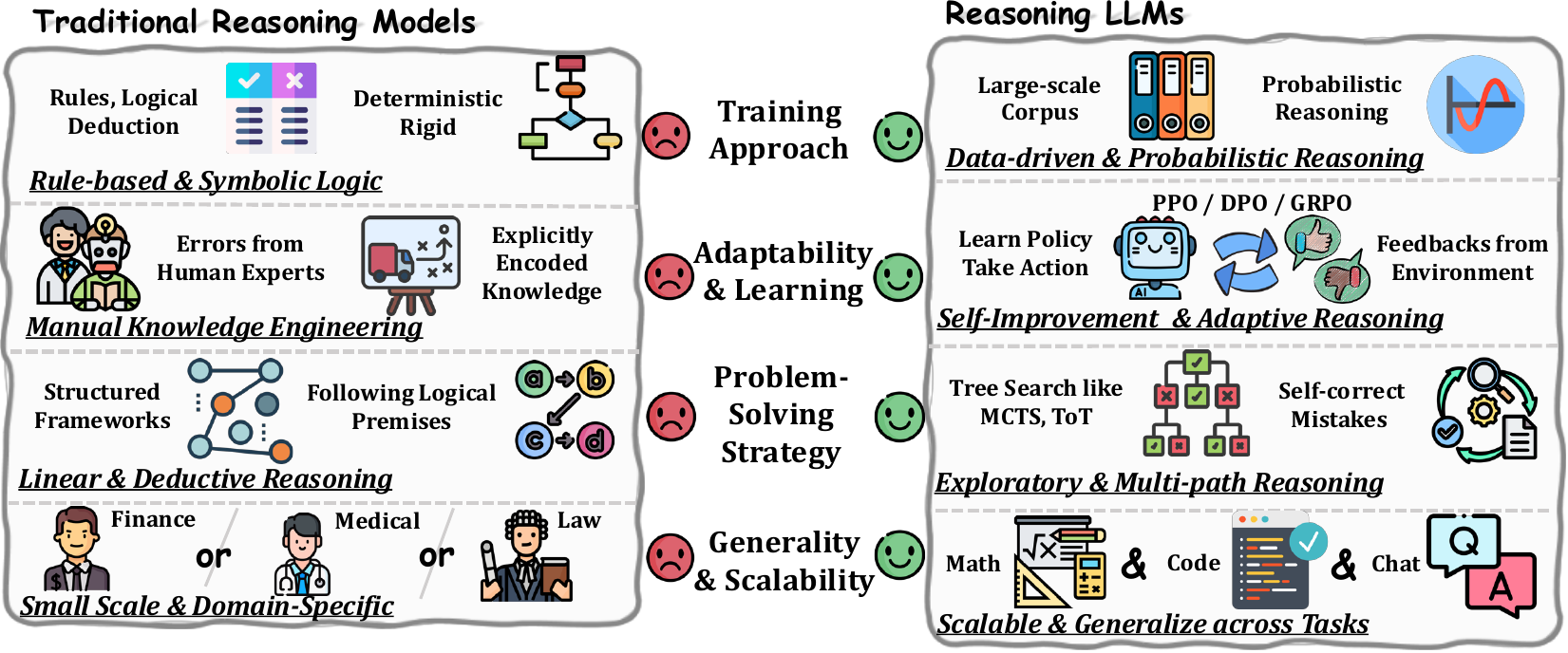}
    \caption{A comprehensive comparison of traditional reasoning models and reasoning LLMs. Reasoning LLMs offer significant advantages over traditional models in areas such as training approaches, adaptability and learning, problem-solving strategies, and generality and scalability.}
    \label{fig:compare}
\end{figure*}

\subsection{Reinforcement Learning}\label{rl}

RL is a type of machine learning where an agent learns to make decisions by interacting with an environment and receiving feedback in the form of rewards, aiming to maximize cumulative rewards over time \cite{sutton1998reinforcement}. 
Early breakthroughs in RL, such as Q-learning \cite{watkins1992q} and DQNs \cite{mnih2015human}, revolutionized the field by enabling the handling of complex state spaces using Deep Neural Networks (DNNs) \cite{torrado2018deep}. These methods paved the way for scaling RL to real-world tasks, where traditional tabular approaches fell short. 
The advent of deep RL marked a significant step forward, combining the power of deep learning with RL to process high-dimensional inputs, such as images and unstructured data.

A landmark achievement in deep RL was AlphaGo, which demonstrated RL's potential by defeating a world champion in the complex game of Go through self-play \cite{silver2016mastering}. 
This success highlighted deep RL's ability to thrive in environments with large, continuous action spaces and uncertainty. 
Building on this, AlphaZero advanced the approach by mastering multiple board games—chess, Go, and Shogi—using self-play, MCTS, and DNNs \cite{silver2017mastering}. 
AlphaZero's ability to learn entirely from scratch, without prior human knowledge, showcased RL's power in environments requiring long-term strategy and planning.

AlphaStar further expanded the boundaries of deep RL by excelling in the real-time strategy game StarCraft II. 
Unlike board games, StarCraft II presents dynamic, partially observable environments and demands multi-step, real-time decision-making \cite{vinyals2019grandmaster}. 
AlphaStar's success in this domain demonstrated deep RL's capacity to adapt to complex decision-making scenarios that require both strategic planning and tactical execution. 
These advancements in RL and deep RL have greatly expanded AI's potential, transitioning from well-defined, static environments to dynamic, complex settings that demand continuous learning and adaptation.

\noindent\textbf{Summary:} 
Deep RL has proven highly effective in solving complex decision-making tasks. AlphaGo exemplifies this by learning strategies through self-play and defeating the world champion in Go. 
This self-play concept laid the foundation for Self Improvement technology (Section \ref{self-improve}) in reasoning LLMs, both relying on continuous feedback and adjustments to optimize strategies.

In RL, reward shaping has been crucial, especially for multi-step reasoning tasks \cite{ng1999policy}. 
By adjusting the reward signal to provide more granular feedback during intermediate steps, it helps agents navigate complex decision-making paths. 
This concept inspired the development of Reward Modeling (Section \ref{prm}), particularly the process reward model, in reasoning LLMs. 
This model offers step-by-step supervision to identify and correct errors in the reasoning process. 
By mimicking human reasoning, the process reward model ensures more robust and interpretable results, especially in tasks like mathematical problem-solving and code generation, where step-by-step evaluation is critical.

Moreover, RL itself is a powerful tool for reasoning LLMs (Section \ref{rl_supervise}). 
With a reward mechanism, RL guides foundational LLMs to find optimal solutions, especially in dynamic reasoning problems. 
Its simplicity and efficiency make RL invaluable for training and optimizing reasoning LLMs, enhancing the intelligence and self-evolution of AI models. 
The integration of RL has led to significant advancements in reasoning LLMs, as demonstrated by DeepSeek-R1 \cite{Deepseek-R1}, offering more flexible and efficient solutions.









\section{Blueprinting Reasoning LLMs}\label{replication}

In this section, we first analyze the features of reasoning LLMs from both output behavior and training dynamics perspectives. We then provide a detailed overview of the core methods that enable their advanced reasoning capabilities. Finally, we summarize the evolution of reasoning LLMs. 
A comprehensive comparison of traditional reasoning models and reasoning LLMs is shown in Figure \ref{fig:compare}.

\subsection{Analysis of the Features of Reasoning LLMs}\label{o1_features}

\subsubsection{Output Behaviour Perspective}\label{output_behaviour}

\textbf{Explore and Planning Structure:} Recent empirical studies have revealed that reasoning LLMs demonstrate a strong tendency for exploratory behavior in their output structures, especially when compared to models such as WizardMath \cite{wizardmath} and DeepSeekMath \cite{deepseekmath}, which primarily rely on conventional CoT reasoning approaches. 
This exploratory behavior is evident in the formulation of novel hypotheses and the pursuit of alternative solution paths. 
Research by \cite{TowardsSystem2ReasoninLLM} suggests that slow-thinking models engage in a latent generative process, particularly noticeable during the prediction of subsequent tokens. 
This claim is supported by \cite{Deepseek-R1}, which observes that similar behaviors naturally arise during RL scale training. 
Furthermore, the Quiet-STaR framework \cite{QuietStar} introduces an auxiliary pre-training phase focused on next-token prediction, highlighting the critical role of internal deliberation and exploratory mechanisms prior to content generation. 
Collectively, these findings underscore the complex and dynamic nature of reasoning processes in advanced LLMs, emphasizing the interaction between exploration and structured reasoning within their operational frameworks.

\noindent\textbf{Verification and Check Structure:} Analysis of OpenAI's o1 \cite{openai_o1} and o3 \cite{o3-mini} models indicates that their reasoning frameworks incorporate both macro-level actions for long-term strategic planning and micro-level actions, including ``\textit{Wait}'', ``\textit{Hold on}'', ``\textit{Alternatively}'', and ``\textit{Let’s pause}''. 
These micro actions facilitate meticulous verification and iterative checking processes, ensuring precision in task execution. 
Such a dual-layered approach underscores the models' capacity to balance overarching goals with granular, detail-oriented operations, thereby enhancing their overall functionality and reliability. 
To emulate this characteristic, Marco-o1 \cite{Marco_o1}, during the MCTS process for constructing Long-CoT, assigns each tree node the state of ``\textit{Wait! Maybe I made some mistakes! I need to rethink from scratch}'', thereby facilitating the reflective nature of Long-CoT. 
Huatuo-o1 \cite{Huatuo-o1} employs a multi-agent framework to address the issue of incorrect CoT generation during validation. 
This is achieved by incorporating a prompt with ``\textit{Backtracking}'' and ``\textit{Correction}'' functionalities,  which enables the correction process.

\noindent\textbf{Longer Inference Length \& Time:} 
Recent research \cite{TowardsSystem2ReasoninLLM, o1_Journey_Part1, o1_Journey_Part2, huang2025o1, reflection_window, atom-of-thougth} indicates that reasoning LLMs often generate outputs exceeding 2000 tokens to tackle complex problems in coding and mathematics. 
However, this extended output length can sometimes lead to overthinking, where the model spends excessive time on a problem without necessarily improving the solution. 
Studies \cite{TowardsSystem2ReasoninLLM} highlight that while autoregressive generation and Classic CoT can effectively solve simpler problems, they struggle with more complex tasks. 
Research \cite{VisualSlowAgent, SlowPerception} shows that in multimodal domains, many problems demand careful observation, comparison, and deliberation. 
Additionally, Search-o1 \cite{li2025search} suggests that slow-thinking mechanisms are particularly beneficial in areas requiring external knowledge or where potential knowledge conflicts arise. 
In medical scenarios \cite{medagent_bench}, complex problems, such as those requiring test-time scaling techniques, demonstrate significant improvements \cite{huang2025o1}.

\noindent\textbf{Overly Cautious \& Simple Problem Trap:} Currently, reasoning LLMs have demonstrated strong performance in domains such as competitive-level mathematics \cite{Deepseek-R1, qwq-32b-preview, RedStar, sky_t1_2025}, complex coding \cite{o1_coder}, medical question answering \cite{Huatuo-o1, huang2025o1}, and multilingual translation \cite{Marco_o1, DRT-o1}. 
These scenarios require the model to perform fine-grained analysis of the problem and execute careful logical reasoning based on the given conditions. 
Interestingly, even for straightforward problems like ``\textit{2+3=?}'', reasoning LLMs can exhibit overconfidence or uncertainty. 
Recent research \cite{Tecent_2_plus_3} notes that o1-like models tend to generate multiple solution rounds for easier math problems, often exploring unnecessary paths. 
This behavior contrasts with the lack of diverse exploratory actions for simpler questions, indicating a potential inefficiency in the model's reasoning process.

\subsubsection{Training Dynamic Perspective}\label{dynamic_perspective}

\textbf{Amazing Data Efficiency:} Unlike traditional approaches that focus on expanding instruction sets with uniformly distributed difficulty levels, Studies \cite{huang2025o1, RedStar} suggest that constructing Slow-thinking CoT datasets with a focus on hard samples leads to better generalization in fields like medicine and mathematics. 
This approach diverges from the conventional practice of collecting diverse and evenly distributed instruction datasets.

\noindent\textbf{Sparse Training Method:} 
Contrary to conventional wisdom, the development of effective reasoning LLMs does not require extensive datasets or dense reward signals. 
For example, STILL2 \cite{o1_Journey_Part2} demonstrated impressive performance using only 5,000 distilled samples, while Sky-T1 \cite{sky_t1_2025} achieved performance parity with QwQ \cite{qwq-32b-preview} using just 17,000 LongCoT samples. 
Similarly, RedStar \cite{RedStar} achieved exceptional results across both textual and multimodal tasks with only 4,000 core LongCoT samples. 
In comparison to simple CoT, Slow-thinking Supervised Fine-Tuning (SFT) data exhibits remarkable sample efficiency, often delivering comparable results with just 1/100th of the sample size. 
Additionally, research \cite{simplerl_reason_blob} emphasizes the significant training potential of online RL scaling algorithms, suggesting that non-dense RL supervision and even rule-based reward structures are sufficient for achieving high performance.

\begin{figure}[t]
    \centering
    \includegraphics[width=0.95\linewidth]{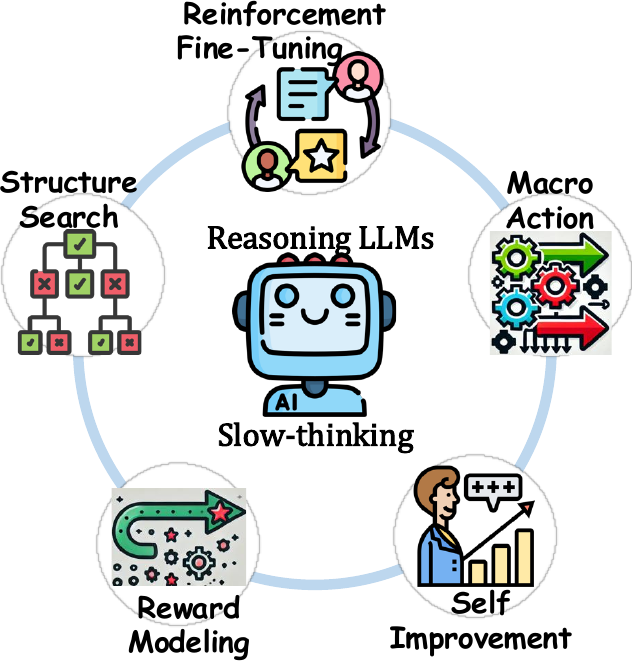}
    \caption{The core methods enabling reasoning LLMs.}
    \label{fig:techs}
\end{figure}

\noindent\textbf{Parameter Characteristic:} 
Training LLMs for slow-thinking, as characterized by the LongCoT approach, results in relatively uniform gradient norms across different layers. 
In contrast, fast-thinking, exemplified by the simplified CoT method, generates larger gradient magnitudes in the earlier layers, along with significant variability in gradient norms across layers. 
Empirical evidence suggests that larger models, particularly those exceeding 30 billion parameters, are more compatible with reasoning LLMs training due to their enhanced capacity for complex reasoning. 
Additionally, experiments conducted by RedStar \cite{RedStar} show that the benefits of data scaling vary across model sizes, with scaling effects being more pronounced and effective in larger models. 
This finding is supported by Deepseek-R1's research \cite{Deepseek-R1}, which demonstrates that a 670-billion-parameter model achieves performance metrics closely approximating those of the o1 benchmark, highlighting the scalability advantages of larger architectures in advanced reasoning tasks.

\subsection{Core Method}\label{foundations}

In this section, we provide an overview of the core methods that drive the advanced reasoning capabilities of reasoning LLMs, as shown in Figure \ref{fig:techs}. These include Structure Search, Reward Modeling, Self Improvement, Macro Action, and Reinforcement Fine-Tuning. 
We also highlight representative reasoning LLMs for each method.

\begin{table*}[tbp]
\centering
\caption{
Summary of tree search-based methods. \textbf{Tag:} ~$\encircle[fill=blue, text=white]{N}$=\underline{N}atural language, ~$\encircle[fill=yellow, text=white]{C}$=\underline{C}ode, ~$\encircle[fill=pink, text=white]{E}$=Math \underline{E}xpression, or ~$\encircle[fill=orange, text=white]{A}$=\underline{A}ction.
}
\resizebox{\textwidth}{!}{%
\begin{tabular}{@{}lccllll@{}}
\toprule
\multicolumn{1}{c}{} &
  \multicolumn{2}{c}{\textbf{Node}} &
  \multicolumn{1}{c}{} &
  \multicolumn{1}{c}{} &
  \multicolumn{1}{c}{} &
  \multicolumn{1}{c}{} \\ \cmidrule(lr){2-3}
\multicolumn{1}{l}{\multirow{-2}{*}{\textbf{Method}}} &
  \textbf{Format} &
  \textbf{Partial} &
  \multicolumn{1}{l}{\multirow{-2}{*}{\textbf{Evaluate}}} &
  \multicolumn{1}{l}{\multirow{-2}{*}{\textbf{Rollout}}} &
  \multicolumn{1}{l}{\multirow{-2}{*}{\textbf{Tasks}}} &
  \multicolumn{1}{l}{\multirow{-2}{*}{\textbf{LLM Models}}} \\ \midrule
\rowcolor[HTML]{EFEFEF} 
RAP \cite{hao2023reasoning} &
  \encircle[fill=blue, text=white]{N} &
  \textcolor{green}{\ding{51}} &
  LLM Self-correction &
  LLM-based Prediction &
  \begin{tabular}[c]{@{}l@{}}Planning, Reasoning\end{tabular} &
  LLaMA-33B \\
ORM \cite{wan2024alphazero} &
  \encircle[fill=blue, text=white]{N} &
  \textcolor{green}{\ding{51}} &
  Value/Reward Function &
  N/A &
  \begin{tabular}[c]{@{}l@{}}Multiple Tasks\end{tabular} &
  \begin{tabular}[c]{@{}l@{}}LLaMA-2-7B, GPT-2-small\end{tabular} \\
\rowcolor[HTML]{EFEFEF} 
Forest-of-Thought \cite{DBLP:journals/corr/abs-2412-09078} &
  \encircle[fill=blue, text=white]{N}
 &
  \textcolor{green}{\ding{51}} &
  LLM Self-correction &
  Self-refinement \& Iterative Improvement &
  \begin{tabular}[c]{@{}l@{}}Planning, Reasoning\end{tabular} &
  \begin{tabular}[c]{@{}l@{}}LLaMA-3, Mistral7B, GLM-4-9B\end{tabular} \\
CodeTree \cite{DBLP:journals/corr/abs-2411-04329} &
  \encircle[fill=blue, text=white]{N}~~\encircle[fill=yellow, text=white]{C} &
  \textcolor{green}{\ding{51}} &
  Execution Accuracy + LLM Self-correction &
  Code Execution &
  Code Generation &
  GPT-4 \\
\rowcolor[HTML]{EFEFEF} 
TreeBoN \cite{DBLP:journals/corr/abs-2410-16033} &
  \encircle[fill=blue, text=white]{N} &
  \textcolor{green}{\ding{51}} &
  Value/Reward Function &
  Speculative \& Dynamic Strategies &
  \begin{tabular}[c]{@{}l@{}}Planning, Reasoning\end{tabular} &
  LLaMA-3-8B \\
CWM \cite{DBLP:journals/corr/abs-2405-15383} &
  \encircle[fill=yellow, text=white]{C} &
  \textcolor{green}{\ding{51}} &
  Compare with Golden Data &
  Code Execution &
  Alignment Task &
  N/A \\
\rowcolor[HTML]{EFEFEF} 
LLM-MCTS \cite{DBLP:conf/nips/ZhaoLH23} &
  \encircle[fill=blue, text=white]{N} &
  \textcolor{red}{\ding{55}} &
  LLM Self-correction + Policy &
  LLM-Based Prediction &
  Household Environments &
  \begin{tabular}[c]{@{}l@{}}GPT-2, GPT-3.5\end{tabular} \\
RethinkMCTS \cite{DBLP:journals/corr/abs-2409-09584} &
  \encircle[fill=yellow, text=white]{C} &
  \textcolor{red}{\ding{55}} &
  Execution Accuracy &
  Code Execution &
  Code Generation &
  \begin{tabular}[c]{@{}l@{}}GPT-3.5-turbo, GPT-4o-mini\end{tabular} \\
\rowcolor[HTML]{EFEFEF} 
MCTSr \cite{DBLP:journals/corr/abs-2406-07394} &
  \encircle[fill=blue, text=white]{N} &
  \textcolor{red}{\ding{55}} &
  LLM Self-correction &
  Self-Refinement \& Iterative Improvement &
  Mathematical Reasoning &
  LLaMA-3-8B \\
MC-NEST \cite{DBLP:journals/corr/abs-2411-15645} &
  \encircle[fill=blue, text=white]{N} &
  \textcolor{red}{\ding{55}} &
  LLM Self-correction &
  Reasoning Path Generation &
  Mathematical Reasoning &
  \begin{tabular}[c]{@{}l@{}}GPT-4o, Phi-3-mini\end{tabular} \\
\rowcolor[HTML]{EFEFEF} 
SRA-MCTS \cite{DBLP:journals/corr/abs-2411-11053} &
  \encircle[fill=blue, text=white]{N} &
  \textcolor{green}{\ding{51}} &
  Execution Accuracy &
  Reasoning Path Generation &
  Code Generation &
  LLaMA-3-70B-Instruct \\
SPaR \cite{DBLP:journals/corr/abs-2412-11605} &
  \encircle[fill=blue, text=white]{N} &
  \textcolor{green}{\ding{51}} &
  LLM Self-correction &
  Self-Refinement \& Iterative Improvement &
  Instruction Following &
  LLaMA-3-8B \\
\rowcolor[HTML]{EFEFEF} 
MindStar \cite{DBLP:journals/corr/abs-2405-16265} &
  \encircle[fill=blue, text=white]{N} &
  \textcolor{green}{\ding{51}} &
  Value/Reward Function &
  Reasoning Path Generation &
  Mathematical Reasoning &
  \begin{tabular}[c]{@{}l@{}}LLaMA-2-13B, Mistral-7B\end{tabular} \\
SR-MCTS \cite{DBLP:journals/corr/abs-2411-04459} &
  \encircle[fill=pink, text=white]{E} &
  \textcolor{green}{\ding{51}} &
  Compare with Golden Data &
  Math Expression Generation &
  Financial Fraud Detection &
  GPT \\
\rowcolor[HTML]{EFEFEF} 
LLaMA-Berry \cite{DBLP:journals/corr/abs-2410-02884} &
  \encircle[fill=blue, text=white]{N} &
  \textcolor{red}{\ding{55}} &
  Compare with Other Solutions &
  Math Expression Generation &
  Mathematical Reasoning &
  LLaMA-3.1-8B \\
Macro-o1 \cite{Marco_o1} &
  \encircle[fill=blue, text=white]{N}~~\encircle[fill=orange, text=white]{A} &
  \textcolor{red}{\ding{55}} &
  LLM's Output Probabilities &
  Reasoning Path Generation &
  \begin{tabular}[c]{@{}l@{}}Multiple Tasks\end{tabular} &
  Qwen2-7B-Instruct \\
\rowcolor[HTML]{EFEFEF} 
ReST-MCTS* \cite{DBLP:journals/corr/abs-2406-03816} &
  \encircle[fill=blue, text=white]{N} &
  \textcolor{green}{\ding{51}} &
  Probability to Correct Answer &
  Reasoning Path Generation &
  Mathematical Reasoning &
  \begin{tabular}[c]{@{}l@{}}LLaMA-3, Mistral-7B, SciGLM-6B\end{tabular} \\ 
  CoMCTS \cite{yao2024mulberryempoweringmllmo1like} &
\encircle[fill=blue, text=white]{N}&
 \textcolor{green}{\ding{51}}&
 Compare with Other Solutions &
 Reasoning Path Generation&
 Multiple Tasks &
   \begin{tabular}[c]{@{}l@{}}GPT-4o, Qwen2-VL-7B, LLaMA-3.2-11B\end{tabular} \\
\rowcolor[HTML]{EFEFEF}  
C-MCTS  \cite{lin2025leveragingconstrainedmontecarlo}  &
\encircle[fill=blue, text=white]{N}&
\textcolor{green}{\ding{51}}&
  Compare with Golden Data &
  Math Expression Generation &
Mathematical Reasoning &
 Qwen2.5  \\
 rStar-Math \cite{guan2025rstarmathsmallllmsmaster} &
 \encircle[fill=pink, text=white]{E}&
\textcolor{green}{\ding{51}}&
 Compare with Other Solutions &
  Math Expression Generation &
Mathematical Reasoning &
\begin{tabular}[c]{@{}l@{}}Phi3-mini-Instruct, Qwen2.5-Math-1.5B, Qwen2.5-Math-7B\end{tabular} \\
\rowcolor[HTML]{EFEFEF}  
 AStar\cite{wu2025boostingmultimodalreasoningmctsautomated}&
 \encircle[fill=blue, text=white]{N}&
 \textcolor{green}{\ding{51}}&
 Compare with Other Solutions &
 Reasoning Path Generation&
 Multiple Tasks &
\begin{tabular}[c]{@{}l@{}}Qwen2.5-7B, Qwen2-VL-2B, Qwen2-VL-7B\end{tabular} \\
DeepSolution \cite{li2025deepsolutionboostingcomplexengineering} &
\encircle[fill=blue, text=white]{N}~~\encircle[fill=orange, text=white]{A} &
\textcolor{red}{\ding{55}} &
Compare with Golden Data &
Reasoning Path Generation &
Multiple Tasks&
Qwen2.5-7B-Instruct \\
\rowcolor[HTML]{EFEFEF} 
VisuoThink\cite{wang2025visuothinkempoweringlvlmreasoning}&
\encircle[fill=blue, text=white]{N}&
\textcolor{green}{\ding{51}}&
Compare with Golden Data &
Reasoning Path Generation &
Multiple Tasks&
\begin{tabular}[c]{@{}l@{}}GPT-4o, Qwen2-VL-72B-Instruct, Claude-3.5-sonnet\end{tabular} \\
TongGeometry \cite{zhang2024proposingsolvingolympiadgeometry}&
 \encircle[fill=pink, text=white]{E}&
\textcolor{green}{\ding{51}}&
Compare with Golden Data  &
  Math Expression Generation &
Mathematical Reasoning &
 GPT-4\\
 \rowcolor[HTML]{EFEFEF} 
PPO-MCTS \cite{liu2024dont}&
\encircle[fill=blue, text=white]{N}&
\textcolor{green}{\ding{51}}&
Compare with Other Solutions &
Reasoning Path Generation &
Alignment Task &
LLaMA-7B \\
  
  \bottomrule
\end{tabular}%
}
\label{tab:structure_search}
\end{table*}

\subsubsection{Structure Search}\label{structure_search}
 
Reasoning LLMs aim to achieve high accuracy and depth in solving complex problems by emulating the deliberate nature of human reasoning. 
However, despite recent advancements, current foundational LLMs face inherent limitations when addressing intricate reasoning tasks. 
These limitations arise from their lack of an internal world model to simulate environmental states, their inability to predict the long-term outcomes of reasoning paths, and their failure to iteratively refine reasoning steps based on future states or rewards \cite{huang2023towards}. 
As a result, these shortcomings hinder foundational LLMs from effectively balancing exploration and exploitation in vast reasoning spaces, creating challenges in tasks that require multi-step reasoning, such as complex mathematics, logical inference, or strategic decision-making \cite{DBLP:journals/corr/abs-2405-00451}.

MCTS, a powerful search and optimization algorithm, effectively addresses these challenges by providing a structured framework to explore and evaluate reasoning paths systematically. 
It operates by constructing a reasoning tree, where each node represents a reasoning state, and actions expand the tree by considering potential next steps. 
Through the simulation of future states and the iterative backpropagation of estimated rewards, MCTS allows foundational LLMs to efficiently identify high-reward reasoning paths, mirroring human planning processes. 
This approach aligns with the core principles of reasoning LLMs, where thorough analysis and deliberate exploration are essential for generating well-reasoned outputs. 
Recent methods, such as RAP \cite{hao2023reasoning}, enhance foundational LLMs by integrating MCTS with a world model, enabling the system to iteratively refine intermediate reasoning steps and improve future predictions. 
Similarly, Forest-of-Thought \cite{DBLP:journals/corr/abs-2412-09078} utilizes MCTS to dynamically explore multiple reasoning trajectories, revisiting flawed paths and refining outcomes. We list the recent methods in Table \ref{tab:structure_search}.

The application of MCTS in reasoning tasks extends beyond traditional problem-solving to highly specialized domains. 
For example, frameworks like SRA-MCTS \cite{DBLP:journals/corr/abs-2411-11053} and MC-NEST \cite{DBLP:journals/corr/abs-2411-15645} showcase the utility of MCTS in tackling technical challenges such as code generation and mathematical reasoning, where intermediate steps are iteratively evaluated and refined. 
In fields like instructional alignment, frameworks such as SPaR \cite{DBLP:journals/corr/abs-2412-11605} and Marco-o1 \cite{Marco_o1} leverage MCTS to refine responses and align reasoning trajectories with human preferences or desired outcomes. 
Additionally, task-specific implementations like HuatuoGPT-o1 \cite{Huatuo-o1} underscore MCTS's crucial role in navigating highly specialized domains, such as medical reasoning, where accuracy and robustness are paramount.

MCTS also enables models to go beyond single-pass reasoning methods, such as CoT or Tree-of-Thought, by incorporating mechanisms to revisit, critique, and refine reasoning steps dynamically \cite{DBLP:journals/corr/abs-2407-01476, DBLP:journals/corr/abs-2409-09584}. 
This iterative capability is essential for tackling tasks with vast decision spaces or those requiring long-term planning, where earlier decisions can significantly impact final outcomes. 
By allowing LLMs to simulate, evaluate, and refine multiple reasoning paths, MCTS introduces a level of adaptability and strategic exploration that traditional approaches lack. 
As shown by AlphaZero-like tree-search \cite{wan2024alphazero} and Search-o1 \cite{li2025search}, MCTS enables reasoning LLMs to not only achieve better performance on specific tasks but also exhibit enhanced generalization capabilities across diverse domains.

The integration of MCTS into LLMs depends on defining actions and rewards to guide reasoning path exploration and assess quality. We classify the actions in prior work into four categories: 
\begin{enumerate}[itemindent=0em]
\item \textbf{Reasoning Steps as Nodes:} Actions represent intermediate reasoning steps or decisions, such as selecting rules, applying transformations, or generating sub-questions \cite{hao2023reasoning, DBLP:journals/corr/abs-2405-00451, wan2024alphazero, DBLP:journals/corr/abs-2412-09078}.

\item \textbf{Token-level Decisions:} Actions involve generating tokens or sequences (\emph{e.g.}, the next word, phrase, or code snippet) \cite{liu2024dontthrowawayvalue, DBLP:journals/corr/abs-2411-04329, DBLP:journals/corr/abs-2412-11605, DBLP:journals/corr/abs-2410-16033}.

\item \textbf{Task-specific Structures:} Actions are domain-specific, such as moving blocks in blocksworld, constructing geometry in geometry problem-solving, or modifying workflows in task planning \cite{DBLP:journals/corr/abs-2412-10673, DBLP:journals/corr/abs-2405-15383, DBLP:conf/nips/ZhaoLH23}.

\item \textbf{Self-correction and Exploration:} Actions focus on revisiting, refining, or backtracking to improve previous reasoning steps \cite{jiang2024intrinsicselfcorrectionenhancementmonte, DBLP:journals/corr/abs-2409-09584, DBLP:journals/corr/abs-2406-07394}.
\end{enumerate}

Additionally, we classify the reward design into five categories:
\begin{enumerate}[itemindent=0em]
\item \textbf{Outcome-based Rewards:} Rewards focus on the correctness or validity of the final outcome or solution, including the validation of reasoning paths or task success \cite{DBLP:journals/corr/abs-2405-00451, DBLP:journals/corr/abs-2411-15645, DBLP:journals/corr/abs-2412-10673}.

\item \textbf{Stepwise Evaluations:} Rewards are assigned at intermediate steps based on the quality of each step or its contribution toward the final outcome \cite{hao2023reasoning, DBLP:journals/corr/abs-2411-11053, wan2024alphazero}.

\item \textbf{Self-evaluation Mechanisms:} Rewards rely on the model’s own confidence or self-assessment (\emph{e.g.}, likelihood, next-word probability, or confidence scores) \cite{DBLP:journals/corr/abs-2412-11605, DBLP:journals/corr/abs-2410-16033, DBLP:journals/corr/abs-2405-16265}.

\item \textbf{Domain-specific Criteria:} Rewards are tailored to specific tasks, such as symmetry and complexity in geometry or alignment with human preferences in text generation \cite{DBLP:journals/corr/abs-2412-10673, DBLP:conf/nips/ZhaoLH23, DBLP:journals/corr/abs-2411-04459}.

\item \textbf{Iterative Preference Learning:} Rewards are derived from comparing multiple solutions or reasoning paths, guiding learning dynamically \cite{DBLP:journals/corr/abs-2410-02884, Marco_o1, DBLP:journals/corr/abs-2406-03816}.

\end{enumerate}

\noindent\textbf{Summary:} Despite its advantages, structure search-based (\emph{i.e.}, MCTS) reasoning LLMs often suffer from substantial computational overhead due to the large number of simulations required. This makes them less suitable for tasks that demand real-time decision-making or operate under resource constraints \cite{DBLP:journals/corr/abs-2309-03224}. 
Additionally, the effectiveness of MCTS is highly dependent on well-designed reward mechanisms and action definitions, which can vary significantly across different domains, thus posing challenges to its generalizability \cite{DBLP:journals/apin/KemmerlingLS24}.

\begin{table*}[t!]
\centering
\renewcommand\arraystretch{1.2}
\caption{Summary of Reward Modeling method.}
\resizebox{0.98\linewidth}{!}{
\begin{tabular}{llllllll}
\toprule[1.2pt]
\multicolumn{2}{c}{\multirow{2}{*}{\textbf{Category}}} & \multirow{2}{*}{\textbf{Methods}} & \multirow{2}{*}{\textbf{Data Source}} & \multicolumn{2}{c}{\textbf{Model Refinement}} & \multirow{2}{*}{\textbf{Applications}} & \multirow{2}{*}{\textbf{Characteristic}}  \\ 
\cmidrule(lr){5-6}
& & & & \textbf{Strategy} & \textbf{Learning} & & \\ 

\midrule

\multicolumn{2}{c}{\multirow{4}{*}{ORM}} & \cellcolor[rgb]{ .949,  .949,  .949}Verifier Training\cite{cobbe2021training} & \cellcolor[rgb]{ .949,  .949,  .949}Existing Data & \cellcolor[rgb]{ .949,  .949,  .949}Verification & \cellcolor[rgb]{ .949,  .949,  .949}SL & \cellcolor[rgb]{ .949,  .949,  .949}Math Reasoning & \cellcolor[rgb]{ .949,  .949,  .949}GSM8K, ORM  \\
                                           \multicolumn{2}{c}{} & ORM PRM Comparison\cite{uesato2022solving} & Human Annotation & Feedback-guided & SFT \& RL & Math Reasoning & PRM ORM Analysis \\
                                           \multicolumn{2}{c}{} & \cellcolor[rgb]{ .949,  .949,  .949}OVM\cite{yu2024ovm} & \cellcolor[rgb]{ .949,  .949,  .949}Sampling & \cellcolor[rgb]{ .949,  .949,  .949}Feedback-guided & \cellcolor[rgb]{ .949,  .949,  .949}SFT & \cellcolor[rgb]{ .949,  .949,  .949}Math Reasoning & \cellcolor[rgb]{ .949,  .949,  .949}Guided Decoding  \\ 
                                           \multicolumn{2}{c}{} & \cellcolor[rgb]{ .949,  .949,  .949}ENCORE\cite{ENCORE} & \cellcolor[rgb]{ .949,  .949,  .949}Existing Data & \cellcolor[rgb]{ .949,  .949,  .949}Entropy-guided & \cellcolor[rgb]{ .949,  .949,  .949}Training-free & \cellcolor[rgb]{ .949,  .949,  .949}Safety Tasks & \cellcolor[rgb]{ .949,  .949,  .949}Entropy of Safety Attribute  \\ 
                                           
                                           \midrule

\multirow{22}{*}{PRM} & \multirow{11}{*}{Outcome-based} & DIVERSE\cite{li2022making} & Prompting & Fine-tuning & SFT & Multiple Reasoning Tasks & Weighted Voting Verifier  \\
                                               \multicolumn{2}{c}{} & \cellcolor[rgb]{ .949,  .949,  .949}MATH-SHEPHERD\cite{wang2024math} & \cellcolor[rgb]{ .949,  .949,  .949}Sampling & \cellcolor[rgb]{ .949,  .949,  .949}Feedback-guided & \cellcolor[rgb]{ .949,  .949,  .949}SFT \& RL & \cellcolor[rgb]{ .949,  .949,  .949}Math Reasoning & \cellcolor[rgb]{ .949,  .949,  .949}Correctness Score Assignment  \\
                                               \multicolumn{2}{c}{} & AutoPSV\cite{lu2024autopsv} & Prompting & Feedback-guided & SFT & Math / Commonsense Reasoning & Automated Process Supervision  \\
                                               \multicolumn{2}{c}{} & \cellcolor[rgb]{ .949,  .949,  .949}Implicit PRMs\cite{yuan2024free} & \cellcolor[rgb]{ .949,  .949,  .949}Sampling & \cellcolor[rgb]{ .949,  .949,  .949}Fine-tuning & \cellcolor[rgb]{ .949,  .949,  .949}SFT \& RL & \cellcolor[rgb]{ .949,  .949,  .949}Math Reasoning & \cellcolor[rgb]{ .949,  .949,  .949}Obtaining PRM from ORM \\
                                               \multicolumn{2}{c}{} & ORPS\cite{yu2024outcome} & Sampling & Feedback-guided & SFT & Code Generation & Supervising Outcome Refinement \\
                                               \multicolumn{2}{c}{} & \cellcolor[rgb]{ .949,  .949,  .949}R-PRM\cite{rprm} & \cellcolor[rgb]{ .949,  .949,  .949}Sampling & \cellcolor[rgb]{ .949,  .949,  .949}Feedback-guided & \cellcolor[rgb]{ .949,  .949,  .949}SFT \& RL & \cellcolor[rgb]{ .949,  .949,  .949}Math Reasoning & \cellcolor[rgb]{ .949,  .949,  .949}Reasoning-Driven Supervision \\
                                               \multicolumn{2}{c}{} & BiRM\cite{BiRM} & Prompting  & Feedback-guided & SFT & Math Reasoning & Bidirectional Reward Signals \\
                                               \multicolumn{2}{c}{} & \cellcolor[rgb]{ .949,  .949,  .949}DeepSeek-GRM\cite{DeepSeek-GRM} & \cellcolor[rgb]{ .949,  .949,  .949}Sampling & \cellcolor[rgb]{ .949,  .949,  .949}Feedback-guided & \cellcolor[rgb]{ .949,  .949,  .949}SFT \& RL & \cellcolor[rgb]{ .949,  .949,  .949}Multiple Reasoning Tasks & \cellcolor[rgb]{ .949,  .949,  .949}Inference-Time Scalability \\
                                               \multicolumn{2}{c}{} & RewardAgent\cite{RewardAgent} & Existing Data & Feedback-guided & SFT \& RL & NLP Tasks & Human \& Verifiable Signals \\
                                               \multicolumn{2}{c}{} & \cellcolor[rgb]{ .949,  .949,  .949}PAR \cite{PAR} & \cellcolor[rgb]{ .949,  .949,  .949}Sampling  & \cellcolor[rgb]{ .949,  .949,  .949}Feedback-guided & \cellcolor[rgb]{ .949,  .949,  .949}SFT \& RL & \cellcolor[rgb]{ .949,  .949,  .949}NLP Tasks & \cellcolor[rgb]{ .949,  .949,  .949}Centered Reward Shaping \\
                                               \multicolumn{2}{c}{} & SCIR  \cite{SCIR} & Sampling  & Feedback-guided & SFT \& RL & NLP Tasks & Self-Consistency Enforcement \\
                                               \cmidrule(lr){2-8}
                      &\multirow{4}{*}{MCTS} & \cellcolor[rgb]{ .949,  .949,  .949}ReST-MCTS$^{*}$\cite{zhang2024rest} & \cellcolor[rgb]{ .949,  .949,  .949}Sampling & \cellcolor[rgb]{ .949,  .949,  .949}Self-training & \cellcolor[rgb] { .949,  .949,  .949}SFT \& RL & \cellcolor[rgb]{ .949,  .949,  .949}Multiple Reasoning Tasks & \cellcolor[rgb]{ .949,  .949,  .949}MCTS and Self-training  \\
                                               \multicolumn{2}{c}{}& OmegaPRM\cite{luo2024improve} & MCTS with Binary Search & Feedback-guided & SFT & Math Reasoning & Divide-and-Conquer MCTS \\
                                               \multicolumn{2}{c}{}& \cellcolor[rgb]{ .949,  .949,  .949}Consensus Filtering\cite{zhang2025lessons} & \cellcolor[rgb]{ .949,  .949,  .949}MCTS Data Construction & \cellcolor[rgb]{ .949,  .949,  .949}Feedback-guided & \cellcolor[rgb]{ .949,  .949,  .949}SFT & \cellcolor[rgb]{ .949,  .949,  .949}Math Reasoning & \cellcolor[rgb]{ .949,  .949,  .949}Consensus Filtering Mechanism  \\
                                               \multicolumn{2}{c}{}& ReARTeR\cite{sun2025rearter} & Sampling & Feedback-guided & SFT \& RL & QA & Retrieval-Augmented Generation  \\
                                               \cmidrule(lr){2-8}
                                               
                      &\multirow{3}{*}{Multimodal} & \cellcolor[rgb]{ .949,  .949,  .949}MSTaR\cite{liu2024diving} & \cellcolor[rgb]{ .949,  .949,  .949}Sampling & \cellcolor[rgb]{ .949,  .949,  .949}Self-training & \cellcolor[rgb]{ .949,  .949,  .949}SFT & \cellcolor[rgb]{ .949,  .949,  .949}Multiple Reasoning Tasks & \cellcolor[rgb]{ .949,  .949,  .949}Adaptive Temperature Adjustment  \\
                                                      \multicolumn{2}{c}{}& VisualPRM\cite{wang2025visualprm} & Sampling & Fine-tuning & SFT & Math Reasoning & Multimoda PRM, BoN \\
                                                     \multicolumn{2}{c}{}&\cellcolor[rgb]{ .949,  .949,  .949}UnifiedReward\cite{UnifiedReward} & \cellcolor[rgb]{ .949,  .949,  .949}Sampling & \cellcolor[rgb]{ .949,  .949,  .949}Feedback-guided & \cellcolor[rgb]{ .949,  .949,  .949}SFT \& RL &\cellcolor[rgb]{ .949,  .949,  .949}Multimodal Tasks &\cellcolor[rgb]{ .949,  .949,  .949}Unified Multimodal Reward Modeling\\
                                                      \cmidrule(lr){2-8}
                      &\multirow{4}{*}{Others} & Pro-Out Feedback\cite{uesato2022solvingmathwordproblems} & Existing Data \& Annotation & Feedback-guided & SFT \& RL & Math Reasoning & Process \& Outcome Supervision \\
                      \multicolumn{2}{c}{}&\cellcolor[rgb]{ .949,  .949,  .949}Verify Step-by-Step\cite{lightman2023let} & \cellcolor[rgb]{ .949,  .949,  .949}Human Annotation & \cellcolor[rgb]{ .949,  .949,  .949}Feedback-guided & \cellcolor[rgb] { .949,  .949,  .949}SFT & \cellcolor[rgb]{ .949,  .949,  .949}Math Reasoning & \cellcolor[rgb]{ .949,  .949,  .949}Process Reward Annotation  \\
                                               \multicolumn{2}{c}{}& Step-DPO\cite{lai2024step} & Sampling & Feedback-guided & SFT \& RL & Math Reasoning & Step-wise Preference Pairs \\
                                               \multicolumn{2}{c}{}& \cellcolor[rgb]{ .949,  .949,  .949}AdaptiveStep\cite{liu2025adaptivestep}  & \cellcolor[rgb]{ .949,  .949,  .949}Response Dividing & \cellcolor[rgb]{ .949,  .949,  .949}Feedback-guided & \cellcolor[rgb]{ .949,  .949,  .949}SFT & \cellcolor[rgb]{ .949,  .949,  .949}Math Reasoning, Code Generation & \cellcolor[rgb]{ .949,  .949,  .949}Dividing   Reasoning Steps  \\

\bottomrule[1.2pt]
\end{tabular}}
\label{table:PRM}
\end{table*}

\subsubsection{Reward Modeling}\label{prm}

Two primary training paradigms are used to tackle multi-step reasoning tasks: outcome supervision and process supervision. 
Outcome supervision emphasizes the correctness of the final answer at a higher level of granularity, and the resulting model is referred to as the Outcome Reward Model (ORM) \cite{cobbe2021training, yu2023outcome, chen2025judgelrm}. 
In contrast, process supervision provides step-by-step labels for the solution trajectory, evaluating the quality of each reasoning step. 
The resulting model is known as the Process Reward Model (PRM) \cite{uesato2022solving, lightmanlet, li2023making}. 


PRM offers significant advantages \cite{wu2023fine, wang2024math} in complex reasoning tasks for several key reasons. First, it provides fine-grained, step-wise supervision, allowing for the identification of specific errors within a solution path. This feature is especially valuable for RL and automated error correction. Second, PRM closely mirrors human reasoning behavior, which relies on accurate intermediate steps to reach correct conclusions. Unlike ORM, PRM avoids situations where incorrect reasoning can still lead to a correct final answer, thus ensuring more robust and interpretable reasoning. While PRM has primarily been applied to complex mathematical problems, its benefits have recently driven applications in other fields. For instance, ORPS \cite{yu2024outcome} utilizes PRM to address complex code generation challenges, while Step-DPO \cite{lai2024step} combines process supervision with the Direct Preference Optimization (DPO) algorithm \cite{rafailov2024direct} to improve long-chain mathematical reasoning. 
The framework has also shown promise in multimodal settings, with M-STAR \cite{liu2024diving} demonstrating how self-evolving training can optimize PRM for vision-language tasks, and VisualPRM \cite{wang2025visualprm} establishing effective process reward modeling for multimodal reasoning through its Best-of-N evaluation approach.
A summary of Reward Modeling method is presented in Table \ref{table:PRM}.

\noindent\textbf{Summary:} 
Despite the advantages of PRMs, they present several challenges. 
The primary difficulty is obtaining process supervision-labeled data, which is often both costly and time-consuming. 
To address concerns related to scalability, efficiency, and accuracy, researchers have explored various automated annotation methods. 
For example, MATH-SHEPHERD \cite{wang2024math} utilizes the correctness of the final answer to define the quality of intermediate steps based on their potential to lead to the correct outcome, automating the step-wise data collection process. ReST-MCTS$^{*}$ \cite{zhang2024rest} combines process reward guidance with MCTS to generate higher-quality reasoning traces through extensive rollouts. 
Similarly, OmegaPRM \cite{luo2024improve} employs the MCTS framework while introducing a divide-and-conquer algorithm for automated process supervision data generation. 
Another novel approach involves using ORM to train a PRM. Yuan et al. \cite{yuan2024free} propose training a PRM implicitly by leveraging ORM training on cheaper datasets, under mild reward parameterization assumptions. 
They also provide theoretical guarantees for the performance of this implicit PRM, demonstrating its practicality and cost-effectiveness.

In addition to data collection, PRMs face challenges related to trustworthiness \cite{sun2025rearter}, categorized as follows: 
\begin{enumerate}[itemindent=0em]
\item \textbf{Lack of Explanations:} Current PRMs often generate scores for reasoning steps without sufficient explanations, limiting interpretability and hindering their usefulness in refining reasoning during test-time. 

\item \textbf{Bias in Training Data:} Data collection methods, such as MCTS, tend to introduce distributional biases, assigning disproportionately higher scores to the majority of questions. As a result, PRMs struggle to effectively identify erroneous reasoning steps. 

\item \textbf{Early-Step Bias:} PRMs show lower accuracy in predicting rewards for earlier reasoning steps compared to those closer to the final answer. This issue stems from the increased randomness and uncertainty associated with the initial steps in the reasoning process.

\end{enumerate}

\begin{table*}[t!]
\centering
\renewcommand\arraystretch{1.2}
\caption{Summary of Self Improvement method.}


\resizebox{0.98\linewidth}{!}{
\begin{tabular}{llllll}
\toprule[1.2pt]
\multirow{2}{*}{\textbf{Stage}}   & \multirow{2}{*}{\textbf{Methods}}   & \multirow{2}{*}{\textbf{Data Source}}  & \multicolumn{2}{c}{\textbf{Model Refinement}} & \multirow{2}{*}{\textbf{Application}}\\\cmidrule(lr){4-5}
& &  & \multicolumn{1}{l}{\textbf{Feedback}} & \multicolumn{1}{l}{\textbf{Strategy}}                            \\ 

\midrule

\multirow{20}{*}{Training} & \cellcolor[rgb]{ .949,  .949,  .949}STaR \cite{DBLP:conf/nips/ZelikmanWMG22}                   & \cellcolor[rgb]{ .949,  .949,  .949}Few-shot  & \cellcolor[rgb]{ .949,  .949,  .949}Language Model & \cellcolor[rgb]{ .949,  .949,  .949}SFT                       & \cellcolor[rgb]{ .949,  .949,  .949}QA, Arithmetic Reasoning                                                        \\

&  Quiet-STaR\cite{QuietStar} &  Token-level Exploration &Language Model &RL                   &  QA, Arithmetic Reasoning                                                                     \\
& \cellcolor[rgb]{ .949,  .949,  .949}V-STaR \cite{hosseini2024vstartrainingverifiersselftaught} & \cellcolor[rgb]{ .949,  .949,  .949}Sampling & \cellcolor[rgb]{ .949,  .949,  .949}Verifier & \cellcolor[rgb]{ .949,  .949,  .949}SFT                   & \cellcolor[rgb]{ .949,  .949,  .949}Arithmetic Reasoning, Code Generation \\

&  B-STaR \cite{zeng2024bstarmonitoringbalancingexploration}  & Sampling &Reward Model &SFT &  Arithmetic Reasoning, Code Generation    \\

& \cellcolor[rgb]{ .949,  .949,  .949}rStar-Math \cite{guan2025rstarmathsmallllmsmaster}         & \cellcolor[rgb]{ .949,  .949,  .949}MCTS Data Construction & \cellcolor[rgb]{ .949,  .949,  .949}Reward Model & \cellcolor[rgb]{ .949,  .949,  .949}SFT                            & \cellcolor[rgb]{ .949,  .949,  .949}Arithmetic Reasoning \\

&  ReST \cite{DBLP:journals/corr/abs-2308-08998}              & Sampling &Reward Model &RL                                          &Machine Translation      \\

& \cellcolor[rgb]{ .949,  .949,  .949}ReST-EM \cite{singh2024humandatascalingselftraining}       & \cellcolor[rgb]{ .949,  .949,  .949}Sampling & \cellcolor[rgb]{ .949,  .949,  .949}Language Model & \cellcolor[rgb]{ .949,  .949,  .949}EM for RL                                   & \cellcolor[rgb]{ .949,  .949,  .949}Arithmetic Reasoning, Code Generation            \\

&  ReST-MCTS* \cite{zhang2024rest}  & Sampling  &Reward Model   &SFT, RL                    &  Reasoning                                              \\

& \cellcolor[rgb]{ .949,  .949,  .949}ENVISIONS \cite{DBLP:journals/corr/abs-2406-11736}         & \cellcolor[rgb]{ .949,  .949,  .949}Sampling & \cellcolor[rgb]{ .949,  .949,  .949}Environment Guided & \cellcolor[rgb]{ .949,  .949,  .949}SFT                      & \cellcolor[rgb]{ .949,  .949,  .949}Web Agents, Reasoning    \\

&  RISE \cite{qu2024recursiveintrospectionteachinglanguage}   &  Sampling &Reward Function &Weighted SFT                                &  Arithmetic Reasoning   \\

& \cellcolor[rgb]{ .949,  .949,  .949}STIC \cite{deng2024enhancinglargevisionlanguage}           & \cellcolor[rgb]{ .949,  .949,  .949}Few-shot & \cellcolor[rgb]{ .949,  .949,  .949}Language Model & \cellcolor[rgb]{ .949,  .949,  .949}SFT                       & \cellcolor[rgb]{ .949,  .949,  .949}Vision Language Model Tasks            \\ 

&  SIRLC \cite{DBLP:conf/iclr/PangWLC0Z024}   &Question Answeing  &Language Model &RL    &Reasoning, Translation, Summary   \\

& \cellcolor[rgb]{ .949,  .949,  .949}AlpacaFarm \cite{DBLP:conf/nips/DuboisLTZGBGLH23}           & \cellcolor[rgb]{ .949,  .949,  .949}Existing Data & \cellcolor[rgb]{ .949,  .949,  .949}Language Model & \cellcolor[rgb]{ .949,  .949,  .949}SFT                       & \cellcolor[rgb]{ .949,  .949,  .949}None (Intrinsic Evaluation)            \\ 

& \cellcolor[rgb]{ .949,  .949,  .949}PSRLM \cite{DBLP:journals/corr/abs-2403-19094}              & \cellcolor[rgb]{ .949,  .949,  .949}Sampling  & \cellcolor[rgb]{ .949,  .949,  .949}Language Model                    & \cellcolor[rgb]{ .949,  .949,  .949}RL  & \cellcolor[rgb]{ .949,  .949,  .949}Reasoning                       \\ 

&  SCRIT \cite{SCRIT}   &Sampling  &Language Model &SFT     &Arithmetic Reasoning   \\

&  \cellcolor[rgb]{ .949,  .949,  .949}S²R \cite{S²R}   &\cellcolor[rgb]{ .949,  .949,  .949}Sampling  &\cellcolor[rgb]{ .949,  .949,  .949}Language Model &\cellcolor[rgb]{ .949,  .949,  .949}SFT \& RL     &\cellcolor[rgb]{ .949,  .949,  .949}Arithmetic Reasoning   \\

&  Self-Training \cite{Self-Training}   &Sampling  &Language Model &SFT     &Arithmetic Reasoning   \\

&  \cellcolor[rgb]{ .949,  .949,  .949}STL \cite{STL}   &\cellcolor[rgb]{ .949,  .949,  .949}Sampling  &\cellcolor[rgb]{ .949,  .949,  .949}Language Model &\cellcolor[rgb]{ .949,  .949,  .949}SFT \& RL     &\cellcolor[rgb]{ .949,  .949,  .949}State-Value Estimation   \\

&  Genius \cite{Genius}   & Sampling  & Language Model & RL  &Arithmetic Reasoning   \\

&  \cellcolor[rgb]{ .949,  .949,  .949}START \cite{START}   &\cellcolor[rgb]{ .949,  .949,  .949}Sampling  &\cellcolor[rgb]{ .949,  .949,  .949}Language Model &\cellcolor[rgb]{ .949,  .949,  .949}SFT \& RL     &\cellcolor[rgb]{ .949,  .949,  .949}Arithmetic Reasoning, Code Generation   \\

&  AlphaMath \cite{alphamath}   &Sampling  & Language Model & RL  &Arithmetic Reasoning   \\

&  \cellcolor[rgb]{ .949,  .949,  .949}HS-STAR \cite{HS-STAR}   &\cellcolor[rgb]{ .949,  .949,  .949}Sampling  &\cellcolor[rgb]{ .949,  .949,  .949}Language Model &\cellcolor[rgb]{ .949,  .949,  .949}SFT \& RL   &\cellcolor[rgb]{ .949,  .949,  .949}Arithmetic Reasoning  \\

\midrule

\multirow{10}{*}{Inference} & Self-Refine \cite{DBLP:conf/nips/MadaanTGHGW0DPY23}   &Independent of Training Data     & Language Model        & Few-shot Demonstration                        & Code Generation, Sentiment Reversal, Acronym Generation                     \\

& \cellcolor[rgb]{ .949,  .949,  .949}Self-Check \cite{DBLP:conf/iclr/MiaoTR24}       & \cellcolor[rgb]{ .949,  .949,  .949}Independent of Training Data            & \cellcolor[rgb]{ .949,  .949,  .949}Language Model           & \cellcolor[rgb]{ .949,  .949,  .949}Step Check               & \cellcolor[rgb]{ .949,  .949,  .949}QA, Arithmetic Reasoning   \\

&  CRITIC \cite{DBLP:conf/iclr/GouSGSYDC24}               &Independent of Training Data    &Language Model  & External Tools                             &  QA, Arithmetic Reasoning, Detoxification          \\

&  EffiLearner \cite{EffiLearner24}               &Independent of Training Data    &Language Model  & External Tools                             &  Code Generation          \\

& \cellcolor[rgb]{ .949,  .949,  .949}ROSE\cite{zhong2024rosedoesntthatboosting}              & \cellcolor[rgb]{ .949,  .949,  .949}Independent of Training Data & \cellcolor[rgb]{ .949,  .949,  .949}Language Model                    & \cellcolor[rgb]{ .949,  .949,  .949}Distributed Prompt                     & \cellcolor[rgb]{ .949,  .949,  .949}Safety, Knowledge    \\ 

&  Self-Verification \cite{DBLP:conf/emnlp/WengZX0HLSLZ23}               &Independent of Training Data    &Language Model  & Re-Ranking                             & Arithmetic Reasoning          \\

& \cellcolor[rgb]{ .949,  .949,  .949}SelfEval-Decoding \cite{xie2023selfevaluationguidedbeamsearch}              & \cellcolor[rgb]{ .949,  .949,  .949}Independent of Training Data & \cellcolor[rgb]{ .949,  .949,  .949}Language Model                    & \cellcolor[rgb]{ .949,  .949,  .949}Beam Search                     & \cellcolor[rgb]{ .949,  .949,  .949}Aritnmetic/Symbolic Reasoning                      \\ 

&  IPS \cite{yao2023finegrainedconversationaldecodingisotropic}               &Independent of Training Data    &Language Model  &Constrained Decoding    &Dialogue \\

& \cellcolor[rgb]{ .949,  .949,  .949}Control-DAG\cite{chen2024controldagconstraineddecodingnonautoregressive}              & \cellcolor[rgb]{ .949,  .949,  .949}Independent of Training Data & \cellcolor[rgb]{ .949,  .949,  .949}Language Model                    & \cellcolor[rgb]{ .949,  .949,  .949}Constrained Decoding                     & \cellcolor[rgb]{ .949,  .949,  .949}Dialogue, Open-domain Generation    \\ 

&  Look-Back \cite{xu2023lookbackdecodingopenendedtext}   &Independent of Training Data    &Language Model  &Contrastive Decoding    &Alleviating Repetitions \\

& \cellcolor[rgb]{ .949,  .949,  .949}LeCo \cite{DBLP:journals/corr/abs-2403-19094}              & \cellcolor[rgb]{ .949,  .949,  .949}Independent of Training Data & \cellcolor[rgb]{ .949,  .949,  .949}Language Model                    & \cellcolor[rgb]{ .949,  .949,  .949}Constrained Decoding                     & \cellcolor[rgb]{ .949,  .949,  .949}QA, Reasoning                       \\ 

&  ProgCo \cite{ProgCo}   &Independent of Training Data    &Language Model  &Program-driven Verification    &Instruction-following, Arithmetic Reasoning \\

\bottomrule[1.2pt]
\end{tabular}
}
\label{table:self-improve}
\end{table*}

\subsubsection{Self Improvement}\label{self-improve}

Reasoning LLMs exemplify a progression from weak to strong supervision, while traditional CoT fine-tuning faces challenges in scaling effectively. 
Self improvement, using the model's exploration capabilities for self-supervision, gradually enhances LLMs performance \cite{Genius} in tasks such as translation \cite{DBLP:journals/corr/abs-2308-08998}, mathematics \cite{DBLP:conf/nips/ZelikmanWMG22, singh2024humandatascalingselftraining}, and multimodal perception \cite{deng2024enhancinglargevisionlanguage}. 
This approach fosters exploration and application within reasoning LLMs \cite{anthony2017thinkingfastslowdeep, tong2024can, tong2024optimizing, guan2025rstarmathsmallllmsmaster}. 
A summary of Self Improvement method is presented in Table \ref{table:self-improve}.

Training-based self improvement in LLMs can be categorized based on exploration and improvement strategies. 
The exploration phase focuses on data collection to facilitate subsequent training improvements, with notable variations in approach. 
STaR \cite{DBLP:conf/nips/ZelikmanWMG22} uses few-shot examples for data gathering, while ReST \cite{DBLP:journals/corr/abs-2308-08998}, ReST-EM \cite{singh2024humandatascalingselftraining}, and ENVISIONS \cite{DBLP:journals/corr/abs-2406-11736} rely on multiple samplings of complete trajectories. 
Quiet-STaR \cite{QuietStar} explores at the token level, introducing concepts like meta-tokens and non-myopic loss to enhance supervision. 
Additionally, ReST-MCTS* \cite{zhang2024rest} and rStar-Math \cite{guan2025rstarmathsmallllmsmaster} generate training data through MCTS.

Improvement strategies also exhibit significant diversity. 
For instance, STaR and its derivatives, such as V-STaR \cite{hosseini2024vstartrainingverifierselftaught} and B-STaR \cite{zeng2024bstarmonitoringbalancingexploration}, combine filtering with SFT. 
ReST and its variants typically introduce innovative reward calculation methods to enhance RL training for policy models. 
RISE \cite{qu2024recursiveintrospectionteachinglanguage} incorporates external feedback, recording rewards and refining responses through distillation during the improvement process. 
Notably, rStar-Math \cite{guan2025rstarmathsmallllmsmaster} demonstrates that small models have achieved \textit{System 2} reflective capabilities through self-evolving training approaches.

Test-time self improvement leverages the consistency of a model's internal knowledge to correct hallucinations during inference. 
These approaches can be categorized into three main types: methods that refine answers using prompts \cite{DBLP:conf/nips/MadaanTGHGW0DPY23, DBLP:conf/iclr/MiaoTR24}, approaches that utilize external tools \cite{DBLP:conf/iclr/GouSGSYDC24}, and techniques that leverage logits without the need for external tools or prompts \cite{DBLP:journals/corr/abs-2403-19094, xu2023lookbackdecodingopenendedtext}.

\subsubsection{Macro Action}\label{macro_action}

Recent advancements in LLMs have driven progress in emulating human-like \textit{System 2} cognitive processes via sophisticated thought architectures, often referred to as macro action frameworks. 
These structured reasoning systems go beyond traditional token-level autoregressive generation by introducing hierarchical cognitive phases, such as strategic planning, introspective verification, and iterative refinement. 
This approach not only enhances the depth of reasoning but also broadens the solution space, enabling more robust and diverse problem-solving pathways. A summary of Macro Action method is presented in Table \ref{table:macro_action}.

\begin{table*}[!t]
\centering
\caption{Summary of Macro Action method. \textbf{Tag}: \encircle[fill=harvestgold, text=white]{I} = \underline{I}mage, \encircle[fill=lightcoral, text=white]{T} = \underline{T}ext, \encircle[fill=DarkGreen, text=white]{V} = \underline{V}ideo. Action Category: AD: Analysis and Decomposition, IPR: Information Processing and Reasoning, VC: Verification and Correction, GO: Generation and Optimization, EB: Exploration and Backtracking.}  

\resizebox{0.98\textwidth}{!}{
   \begin{tabular}%
   {lllllccc}
    \toprule
        \multirow{2}{*}{\textbf{Methods}} & \multirow{2}{*}{\textbf{Usage}} & \multicolumn{5}{c}{\textbf{Action Attribute}} & \multirow{2}{*}{\textbf{Main Action Category}} \\ \cmidrule(lr){3-7}
          & & \textbf{Action Source} & \textbf{Action Number} & \textbf{Learning} & \textbf{Reflection} & \textbf{Modality} & \\ \midrule
       \rowcolor[rgb]{ .949,  .949,  .949} Self-Check\cite{miao2023selfcheck} & Verification & Human-Designed & 4 & ICL & \textcolor{green}{\ding{51}} & \encircle[fill=lightcoral, text=white]{T} & AD, VC \\
       
        LeMa\cite{LeMa} & Synthetic Data & Human-Designed & 3 & ICL \& SFT & \textcolor{green}{\ding{51}} & \encircle[fill=lightcoral, text=white]{T} & VC, IPR \\
        
       \rowcolor[rgb]{ .949,  .949,  .949} REFINER\cite{li2024refiner} & Verification/Exploration & Human-Designed & 2 & ICL \& SFT & \textcolor{green}{\ding{51}} & \encircle[fill=lightcoral, text=white]{T} & VC, AD \\
        
        HiICL-MCTS\cite{HiICL-MCTS} & Exploration & Human-Designed & 5 & ICL & \textcolor{green}{\ding{51}} & \encircle[fill=lightcoral, text=white]{T} & VC, EB, AD  \\
        
         \rowcolor[rgb]{ .949,  .949,  .949} SUPERCORRECT\cite{yang2024supercorrect} & Distill & In-Context Learning & Dynamic & SFT \& RL & \textcolor{red}{\ding{55}} & \encircle[fill=lightcoral, text=white]{T} & AD, IPR \\
        
        ReasonFlux\cite{ReasonFlux}  & Synthetic Data/Exploration & Human-Designed & $\sim$500 & ICL \& SFT \& RL & \textcolor{red}{\ding{55}} &  \encircle[fill=lightcoral, text=white]{T} & AD, IPR \\
        
         \rowcolor[rgb]{ .949,  .949,  .949} rStar\cite{rSTaR} & Exploration & Human-Designed & 5 & ICL \& RL & \textcolor{green}{\ding{51}} & \encircle[fill=lightcoral, text=white]{T} & VC, GO, EB \\
         
        LLaMA-Berry\cite{zhang2024llama} & Exploration & Human-Designed & 2 & ICL \& RL & \textcolor{green}{\ding{51}} & \encircle[fill=lightcoral, text=white]{T} & VC, EB \\
    
        \rowcolor[rgb]{ .949,  .949,  .949} Huatuo-o1\cite{Huatuo-o1} & Synthetic Data & Human-Designed & 4 & ICL \& SFT & \textcolor{green}{\ding{51}} & \encircle[fill=lightcoral, text=white]{T} & VC \\

        Marco-o1\cite{Marco_o1} & Verification & Human-Designed & 1 & ICL \& SFT & \textcolor{green}{\ding{51}} & \encircle[fill=lightcoral, text=white]{T} & VC \\
        
         \rowcolor[rgb]{ .949,  .949,  .949}BoT\cite{yang2024buffer} & Exploration & In-Context Learning & Dynamic & ICL & \textcolor{red}{\ding{55}} & \encircle[fill=lightcoral, text=white]{T} & AD, IPR \\
         
        rStar-Math\cite{guan2025rstarmathsmallllmsmaster} & Exploration &  In-Context Learning & $1$ & ICL \& RL & \textcolor{green}{\ding{51}} & \encircle[fill=lightcoral, text=white]{T} & AD, IPR \\
        
         \rowcolor[rgb]{ .949,  .949,  .949}Mulberry\cite{yao2024mulberry} & Synthetic Data & In-Context Learning & 1 & ICL \& SFT & \textcolor{green}{\ding{51}}  & \encircle[fill=lightcoral, text=white]{T} & VC, EB \\
         
        LLaVA-CoT\cite{xu2024llava} & Synthetic Data/Exploration & Human-Designed & 4 & SFT & \textcolor{red}{\ding{55}} & \encircle[fill=harvestgold, text=white]{I} \space \encircle[fill=lightcoral, text=white]{T} & AD, IPR \\
        
         \rowcolor[rgb]{ .949,  .949,  .949}LLaMAV-o1\cite{thawakar2025llamav} & Verification/Exploration & Human-Designed & 4173 & Curriculum Learning & \textcolor{green}{\ding{51}} &  \encircle[fill=harvestgold, text=white]{I} \space \encircle[fill=lightcoral, text=white]{T} & AD, IPR \\
         
        AtomThink\cite{atomthink} & Synthetic Data/Exploration & In-Context Learning & $>$100 & SFT \& RL & \textcolor{green}{\ding{51}} & \encircle[fill=harvestgold, text=white]{I} \space \encircle[fill=lightcoral, text=white]{T} & AD, IPR, EB \\

         \rowcolor[rgb]{ .949,  .949,  .949}RedStar\cite{RedStar} & Distill & Human-Designed & 2 & SFT & \textcolor{green}{\ding{51}} & \encircle[fill=harvestgold, text=white]{I} \space \encircle[fill=lightcoral, text=white]{T} & AD, VC \\
        
        Auto-CoT \cite{zhangautomatic}  & Exploration & In-Context Learning & 2 & ICL & \textcolor{red}{\ding{55}} & \encircle[fill=lightcoral, text=white]{T} & AD, IPR, GO \\
        
         \rowcolor[rgb]{ .949,  .949,  .949}PoT\cite{chen2022program} & Verification & In-Context Learning & 1 & ICL & \textcolor{red}{\ding{55}} & \encircle[fill=lightcoral, text=white]{T} & AD, IPR, GO \\
        
        PAL\cite{gao2023pal} & Verification & In-Context Learning & 1 & ICL & \textcolor{red}{\ding{55}} & \encircle[fill=lightcoral, text=white]{T} & AD, IPR, GO \\
        
         \rowcolor[rgb]{ .949,  .949,  .949} Decomposed Prompt\cite{decomposed} & Exploration & Human-Designed & 3 & ICL & \textcolor{red}{\ding{55}} & \encircle[fill=lightcoral, text=white]{T} & AD, IPR \\
        
        Least-to-Most\cite{least2most} & Exploration & Human-Designed & 2 & ICL & \textcolor{red}{\ding{55}} & \encircle[fill=lightcoral, text=white]{T} & AD, IPR \\

        CoR-Math \cite{CoR} & Synthetic Data & Human-Designed & 3 & SFT & \textcolor{green}{\ding{51}} & \encircle[fill=lightcoral, text=white]{T} & AS, SR, NLR \\
        \bottomrule
   \end{tabular}
}
    \label{table:macro_action}
\end{table*}

We classify the progress of macro action into two aspects: 
\begin{enumerate}[itemindent=0em]
\item \textbf{Test-time Scaling through Macro Action Operationalization:} Recent research identifies two key methodologies for improving reasoning performance during inference and test-time scaling. 
HiICL-MCTS \cite{HiICL-MCTS} employs a deliberate search through seed data to generate action-chain templates consisting of macro actions, thereby facilitating an action-chain-guided approach to test-time reasoning. ReasonFlux \cite{ReasonFlux} utilizes an iterative test-time scaling framework, harnessing external high-level thought templates to iteratively refine and update the current CoT.


\item \textbf{Macro Action-Enhanced Data Synthesis Paradigms:} A key application of macro actions in complex reasoning is in the synthesis of reasoning data \cite{llm_anotation_and_synthesis}. 
In data synthesis and training frameworks, macro action architectures enhance reasoning diversity and generalization. 
Recent research has shown that integrating or synthesizing a CoT process with macro actions within the reasoning sequence can significantly improve the data efficiency of the reasoning chain. 
For instance, LLaVA-CoT \cite{xu2024llava} enhances CoT data synthesis by externalizing intermediate reasoning steps across multiple modalities. 
AtomThink \cite{atomthink} generates the AMATH-SFT dataset using a structured g1 prompt \cite{g1}, achieving superior performance on long-horizon reasoning tasks compared to traditional CoT approaches. 
CoAct \cite{hou2024coact} introduces a dual-agent collaborative reasoning framework, where a global planning agent executes overarching macro-actions, while a local execution agent carries out specific sub-actions within those broader actions.

\end{enumerate}

Macro actions also play a crucial role in enhancing self improvement frameworks. 
rStar-Math \cite{guan2025rstarmathsmallllmsmaster} utilizes high-level deliberate search through Code-augmented CoT, generating diverse and reliable solutions while achieving proactive search capabilities. 
Satori \cite{Satori} integrates CoT with RL, incorporating ``$<$\textit{reflect}$>$''-style macro actions to diversify exploration and alleviate policy saturation in online RL environments. 
Huatuo-o1 \cite{Huatuo-o1} combines hierarchical planning with domain-specific knowledge bases to improve medical reasoning. 
Additionally, ReasonFlux \cite{ReasonFlux} dynamically reconfigures reasoning templates (\emph{e.g.}, breaking down calculus problems into symbolic and numeric phases) to align with the problem structure.

\subsubsection{Reinforcement Fine-Tuning}\label{rl_supervise}

Reinforcement Fine-Tuning (RFT) \cite{openaiRFT} is an innovative technique recently introduced by OpenAI, designed to enable developers and engineers to fine-tune existing models for specific domains or complex tasks. Unlike general SFT, RFT focuses on optimizing the model's reasoning process by using a reward mechanism to guide the model's evolution, thereby enhancing its reasoning capabilities and accuracy. 
The core of RFT lies in improving the model's performance in a specific domain with minimal high-quality training data \cite{chang2024survey}, an appropriate reward model \cite{trung2024reft}, and a stable optimization process in long-context \cite{blobRFT, codepmp, nextlong, quest}. 
A summary of RFT method is presented in Table \ref{table:rl_supervise}.

DeepSeek-R1 \cite{Deepseek-R1}, which employs a verifier reward-based strategy, has shown significant performance improvements compared to traditional methods like SoS \cite{SoS}. Key advantages include:
\begin{enumerate}[itemindent=0em]
    \item \textbf{Simplified Training Pipeline:} RL supervision streamlines data construction and training processes, eliminating the need for complex stepwise search mechanisms. 
    
    \item \textbf{Enhanced Scalability:} Online RL training facilitates efficient scaling on large datasets, particularly for complex reasoning tasks.
    
    \item \textbf{Emergent Properties:} DeepSeek-R1 \cite{Deepseek-R1} demonstrates unique emergent capabilities, such as Long-CoT reasoning, which are difficult to achieve through SFT alone.
\end{enumerate}

The success of R1 in unimodal reasoning has spurred its adaptation to MLLM, achieving SOTA performance across diverse domains (math \cite{MMR1-Math2025,meng2025mm},  medical imaging \cite{pan2025medvlm}, segmentation \cite{liu_seg-zero_2025}). Key strategies include rule-based reward systems to incentivize structured reasoning (\emph{e.g.}, step-wise validity/accuracy rewards in R1-VL \cite{zhang_r1-vl_2025}), modality-agnostic training frameworks (\emph{e.g.}, Vision-R1’s \cite{huang_vision-r1_2025} cold-start CoT data generation), and efficient parameter utilization (\emph{e.g.}, MMR1 \cite{MMR1-Math2025}’s 7B models rivaling larger proprietary counterparts). The work emphasizes transparency through explicit reasoning paths (\emph{e.g.}, MedVLM-R1’s \cite{pan2025medvlm} interpretable medical analysis) and open-source contributions (code \cite{zheng2025easyr1}, data \cite{meng2025mm}, benchmarks \cite{yang_r1-onevision_2025,chen_exploring_2025}), fostering reproducibility and MLLM-community improvement.

\begin{table*}[!t]
\centering
\caption{Summary of RFT method. \textbf{Tag}: \encircle[fill=harvestgold, text=white]{I} = \underline{I}mage, \encircle[fill=lightcoral, text=white]{T} = \underline{T}ext, \encircle[fill=DarkGreen, text=white]{V} = \underline{V}ideo.}

\resizebox{0.98\textwidth}{!}{
   \begin{tabular}%
   {lllclllll}
    \toprule
        \multirow{2}{*}{\textbf{Category}}&\multirow{2}{*}{\textbf{Methods}} & \multicolumn{2}{c}{\textbf{Model Attribute}} & \multicolumn{4}{c}{\textbf{Incentivize Attribute}} & \multirow{2}{*}{\textbf{Application \& Benchmark}} \\
        
        \cmidrule(lr){3-4} \cmidrule(lr){5-8}

         && \textbf{Foundational LLMs} & \textbf{Feedback Modality} & \textbf{Reward Type} & \textbf{Algorithm} & \textbf{Learning} & \textbf{Incentivize Sample} & \\ 
         
         \midrule
        
        \multicolumn{9}{c}{\textbf{Reason RFT Project}} \\ 
        
        \midrule
        
         \multirow{21}{*}{LLM}&\cellcolor[rgb]{ .949,  .949,  .949}DeepSeek-R1-Zero\cite{Deepseek-R1} & \cellcolor[rgb]{ .949,  .949,  .949}DeepSeek-V3 & \cellcolor[rgb]{ .949,  .949,  .949}\encircle[fill=lightcoral, text=white]{T} & \cellcolor[rgb]{ .949,  .949,  .949}Rule-Outcome-Reward & \cellcolor[rgb]{ .949,  .949,  .949}GPRO & \cellcolor[rgb]{ .949,  .949,  .949}RL & \cellcolor[rgb]{ .949,  .949,  .949}800K  & \cellcolor[rgb]{ .949,  .949,  .949}Multiple Tasks \\
        
        &DeepSeek-R1\cite{Deepseek-R1} & DeepSeek-V3 & \encircle[fill=lightcoral, text=white]{T} & Rule-Outcome-Reward & GPRO & RL \& SFT & 800K  & Multiple Tasks \\
        
         &\cellcolor[rgb]{ .949,  .949,  .949}Kimi k1.5\cite{team2025kimi} & \cellcolor[rgb]{ .949,  .949,  .949}-- & \cellcolor[rgb]{ .949,  .949,  .949}\encircle[fill=harvestgold, text=white]{I} \space\encircle[fill=lightcoral, text=white]{T} & \cellcolor[rgb]{ .949,  .949,  .949}Rule-Outcome-Reward & \cellcolor[rgb]{ .949,  .949,  .949}$\text{PPO}^{*}$ & \cellcolor[rgb]{ .949,  .949,  .949}RL \& SFT & \cellcolor[rgb]{ .949,  .949,  .949}-- & \cellcolor[rgb]{ .949,  .949,  .949}Multiple Tasks \\
         
        &ReFT\cite{trung2024reft}  & Galactica, CodeLLama & \encircle[fill=lightcoral, text=white]{T} & Rule-Outcome-Reward & $\text{PPO}^{*}$ & RL \& SFT & 3k/7k/8k/15k & GSM8k/SVAMP/MathQA \\
        
         &\cellcolor[rgb]{ .949,  .949,  .949}RFTT\cite{RFTT} & \cellcolor[rgb]{ .949,  .949,  .949}LLaMA-3-3/8B-Instruct, Qwen-2.5-7B-Instruct & \cellcolor[rgb]{ .949,  .949,  .949}\encircle[fill=lightcoral, text=white]{T} & \cellcolor[rgb]{ .949,  .949,  .949}Rule-Outcome-Reward &  \cellcolor[rgb]{ .949,  .949,  .949}Reinforce++ & \cellcolor[rgb]{ .949,  .949,  .949}RL \& SFT & \cellcolor[rgb]{ .949,  .949,  .949}1.2K & \cellcolor[rgb]{ .949,  .949,  .949}Multiple Math Task \\
         
        &Satori\cite{Satori} & Qwen-2.5-Math-7B & \encircle[fill=lightcoral, text=white]{T} & Rule-Outcome-Reward & PPO & RL \& SFT & 66K & Multiple Math Task  \\

        &\cellcolor[rgb]{ .949,  .949,  .949}QCLASS\cite{QCLASS} & \cellcolor[rgb]{ .949,  .949,  .949}Llama-2-7B-Chat & \cellcolor[rgb]{ .949,  .949,  .949}\encircle[fill=lightcoral, text=white]{T} & \cellcolor[rgb]{ .949,  .949,  .949}Process-Reward & \cellcolor[rgb]{ .949,  .949,  .949}QNet & \cellcolor[rgb]{ .949,  .949,  .949}RL \& SFT & \cellcolor[rgb]{ .949,  .949,  .949}1.9K/1.5K/3.3K & \cellcolor[rgb]{ .949,  .949,  .949}WebShop, ALFWorld, SciWorld \\
        
        &PRIME\cite{cui2025process} & Qwen2.5-Math-7B & \encircle[fill=lightcoral, text=white]{T} & Rule-Process-Outcome-Reward & PPO & RL \& SFT  & 150K & Math, Code Tasks  \\
        
        &\cellcolor[rgb]{ .949,  .949,  .949}DeepScaleR\cite{deepscaler2025} & \cellcolor[rgb]{ .949,  .949,  .949}DeepSeek-R1-Distill-Qwen-1.5B & \cellcolor[rgb]{ .949,  .949,  .949}\encircle[fill=lightcoral, text=white]{T} & \cellcolor[rgb]{ .949,  .949,  .949}Rule-Outcome-Reward & \cellcolor[rgb]{ .949,  .949,  .949}Iteratively GPRO & \cellcolor[rgb]{ .949,  .949,  .949}RL & \cellcolor[rgb]{ .949,  .949,  .949}40K & \cellcolor[rgb]{ .949,  .949,  .949}Multiple Math Task \\

        &PURE\cite{cheng2025pure} & Qwen2.5-Math-7B &  \encircle[fill=lightcoral, text=white]{T} & Rule-Process-Outcome-Reward & PPO+RLOO & RL & 8K & Multiple Math Task \\
        
        &\cellcolor[rgb]{ .949,  .949,  .949}SimpleRL\cite{simplerl_reason_blob} & \cellcolor[rgb]{ .949,  .949,  .949}Qwen2.5-Math-7B & \cellcolor[rgb]{ .949,  .949,  .949}\encircle[fill=lightcoral, text=white]{T} & \cellcolor[rgb]{ .949,  .949,  .949}Rule-Outcome-Reward & \cellcolor[rgb]{ .949,  .949,  .949}PPO & \cellcolor[rgb]{ .949,  .949,  .949}RL & \cellcolor[rgb]{ .949,  .949,  .949}8K & \cellcolor[rgb]{ .949,  .949,  .949}Multiple Math Task \\
        
        &Open-R1\cite{openr1} & Qwen2.5-1.5B-Instruct & \encircle[fill=lightcoral, text=white]{T} & Rule-Outcome-Reward & GPRO & RL \& SFT & 8K & Multiple Math, Code Task \\
        
        &\cellcolor[rgb]{ .949,  .949,  .949}TinyZero\cite{tinyzero} & \cellcolor[rgb]{ .949,  .949,  .949}Qwen2.5-0.5B/3B & \cellcolor[rgb]{ .949,  .949,  .949}\encircle[fill=lightcoral, text=white]{T} & \cellcolor[rgb]{ .949,  .949,  .949}Rule-Outcome-Reward & \cellcolor[rgb]{ .949,  .949,  .949}GPRO & \cellcolor[rgb]{ .949,  .949,  .949}RL & \cellcolor[rgb]{ .949,  .949,  .949}-- & \cellcolor[rgb]{ .949,  .949,  .949}CountDown Task \\
        
        &Ota-Zero\cite{liu2025oatzero} & Qwen-2.5-Series, DeepSeek-Series, Rho, Llama-3.x & \encircle[fill=lightcoral, text=white]{T} & Rule-Outcome-Reward & GRPO & RL & 0.5K & CountDown Task \\
        
        &\cellcolor[rgb]{ .949,  .949,  .949}Ota\cite{liu2025oat} & \cellcolor[rgb]{ .949,  .949,  .949}RHO-1b/Qwen2.5-3B & \cellcolor[rgb]{ .949,  .949,  .949}\encircle[fill=lightcoral, text=white]{T} & \cellcolor[rgb]{ .949,  .949,  .949}Rule-Outcome-Reward & \cellcolor[rgb]{ .949,  .949,  .949}GPRO/PPO & \cellcolor[rgb]{ .949,  .949,  .949}RL & \cellcolor[rgb]{ .949,  .949,  .949}7.5K & \cellcolor[rgb]{ .949,  .949,  .949}GSM8K \\
        
        &LIMR\cite{limr} & Qwen-Math-7B & \encircle[fill=lightcoral, text=white]{T} & Rule-Outcome-Reward & PPO & RL & 1.3K & Multiple Math Task \\
        
        &\cellcolor[rgb]{ .949,  .949,  .949}Critic-RL\cite{critic_rl} & \cellcolor[rgb]{ .949,  .949,  .949}Qwen2.5-Coder-32B & \cellcolor[rgb]{ .949,  .949,  .949}\encircle[fill=lightcoral, text=white]{T} & \cellcolor[rgb]{ .949,  .949,  .949}Rule-Outcome-Reward & \cellcolor[rgb]{ .949,  .949,  .949}$\text{GPRO}^{*}$ & \cellcolor[rgb]{ .949,  .949,  .949}RL \& SFT & \cellcolor[rgb]{ .949,  .949,  .949}18.8K & \cellcolor[rgb]{ .949,  .949,  .949}Multiple Code Task \\
        
        &Logic-R1\cite{logicrl} & Qwen2.5-7B-Instruct-1M & \encircle[fill=lightcoral, text=white]{T} & Rule-Outcome-Reward & $\text{REINFORCE++}^{*}$ & RL & 5K & Multiple Math, Logic Task \\
        
        &\cellcolor[rgb]{ .949,  .949,  .949}Online-DPO-R1\cite{online_dpo_r1} & \cellcolor[rgb]{ .949,  .949,  .949}Qwen2.5-MATH-7B & \cellcolor[rgb]{ .949,  .949,  .949}\encircle[fill=lightcoral, text=white]{T} & \cellcolor[rgb]{ .949,  .949,  .949}Rule-Outcome-Reward  & \cellcolor[rgb]{ .949,  .949,  .949}DPO  & \cellcolor[rgb]{ .949,  .949,  .949}RL\& SFT & \cellcolor[rgb]{ .949,  .949,  .949}207.5K & \cellcolor[rgb]{ .949,  .949,  .949}Multiple Math Task \\
        
        &OpenReason-Zero\cite{OpenReasonerZero2025} & Qwen-2.5-7B/32B & \encircle[fill=lightcoral, text=white]{T} & Rule-Outcome-Reward   & PPO & RL & 57K & Multiple Math Task, GPQA, MMLU \\
        
        &\cellcolor[rgb]{ .949,  .949,  .949}RLAIF\cite{lee2023rlaif} & \cellcolor[rgb]{ .949,  .949,  .949}PaLM 2 Extra-Small & \cellcolor[rgb]{ .949,  .949,  .949}\encircle[fill=lightcoral, text=white]{T} & \cellcolor[rgb]{ .949,  .949,  .949}Rule-Outcome-Reward & \cellcolor[rgb]{ .949,  .949,  .949}RLAIF & \cellcolor[rgb]{ .949,  .949,  .949}RL & \cellcolor[rgb]{ .949,  .949,  .949}-- & \cellcolor[rgb]{ .949,  .949,  .949}Summary and Conversation Generation \\

        \midrule
        
        \multirow{24}{*}{MLLM}&RLHF-V\cite{yu2024rlhfv} & OmniLMM-12B & \encircle[fill=harvestgold, text=white]{I} \space\encircle[fill=lightcoral, text=white]{T} & Process-Reward &  DDPO & RL & 1.4K & Multiple Tasks \\

        &\cellcolor[rgb]{ .949,  .949,  .949}MM-RLHF\cite{MM_RLHF} & \cellcolor[rgb]{ .949,  .949,  .949}LLaVA-onevision-7B & \cellcolor[rgb]{ .949,  .949,  .949}\encircle[fill=harvestgold, text=white]{I} \space\encircle[fill=lightcoral, text=white]{T} \space\encircle[fill=DarkGreen, text=white]{V} & \cellcolor[rgb]{ .949,  .949,  .949}Process-Reward & \cellcolor[rgb]{ .949,  .949,  .949}MM-DPO & \cellcolor[rgb]{ .949,  .949,  .949}RL & \cellcolor[rgb]{ .949,  .949,  .949}120K & \cellcolor[rgb]{ .949,  .949,  .949}MM-RLHF-RewardBench/SafetyBench\\
        
        &Align-DS-V\cite{Align_DS_V}  & LLaVA-v1.5-7B,Qwen2-VL &\encircle[fill=harvestgold, text=white]{I} \space\encircle[fill=lightcoral, text=white]{T} \space\encircle[fill=DarkGreen, text=white]{V} & Process-Reward & PPO, DPO & RL \& SFT & 200K & Align-Anything, Eval-Anything \\
        
        &\cellcolor[rgb]{ .949,  .949,  .949}R1V\cite{chen2025r1v}  & \cellcolor[rgb]{ .949,  .949,  .949}Qwen2-VL,Qwen2.5-VL & \cellcolor[rgb]{ .949,  .949,  .949}\encircle[fill=harvestgold, text=white]{I} \space\encircle[fill=lightcoral, text=white]{T} & \cellcolor[rgb]{ .949,  .949,  .949}Rule-Outcome-Reward & \cellcolor[rgb]{ .949,  .949,  .949}GRPO & \cellcolor[rgb]{ .949,  .949,  .949}RL & \cellcolor[rgb]{ .949,  .949,  .949}70K/70K/8K & \cellcolor[rgb]{ .949,  .949,  .949}Multiple Tasks \\
        
        &VLM-R1\cite{vlmr1} & Qwen2.5-VL & \encircle[fill=harvestgold, text=white]{I} \space\encircle[fill=lightcoral, text=white]{T} & Rule-Outcome-Reward & GRPO & RL & 120K & Multiple Tasks \\
        
        &\cellcolor[rgb]{ .949,  .949,  .949}LMM-R1\cite{lmmr1} & \cellcolor[rgb]{ .949,  .949,  .949}Qwen2.5-VL & \cellcolor[rgb]{ .949,  .949,  .949}\encircle[fill=harvestgold, text=white]{I} \space\encircle[fill=lightcoral, text=white]{T} & \cellcolor[rgb]{ .949,  .949,  .949}Rule-Outcome-Reward & \cellcolor[rgb]{ .949,  .949,  .949}PPO/RLOO & \cellcolor[rgb]{ .949,  .949,  .949}RL & \cellcolor[rgb]{ .949,  .949,  .949}8K & \cellcolor[rgb]{ .949,  .949,  .949}Multiple Tasks \\
        
        &Open-R1-Video\cite{open-r1-video} & Qwen2-VL-7B & \encircle[fill=harvestgold, text=white]{I} \space\encircle[fill=lightcoral, text=white]{T} \space\encircle[fill=DarkGreen, text=white]{V} & Rule-Outcome-Reward & GRPO & RL & 4K & Multiple Tasks \\
        
        &\cellcolor[rgb]{ .949,  .949,  .949}Easy-R1 \cite{zheng2025easyr1} & \cellcolor[rgb]{ .949,  .949,  .949}Qwen2.5-VL & \cellcolor[rgb]{ .949,  .949,  .949}\encircle[fill=harvestgold, text=white]{I} \space\encircle[fill=lightcoral, text=white]{T} & \cellcolor[rgb]{ .949,  .949,  .949}Rule-Outcome-Reward & \cellcolor[rgb]{ .949,  .949,  .949}GRPO & \cellcolor[rgb]{ .949,  .949,  .949}RL & \cellcolor[rgb]{ .949,  .949,  .949}3K & \cellcolor[rgb]{ .949,  .949,  .949}Multiple Tasks \\
        
        &Efficient-R1-VLLM\cite{baiefficient} & DeepSeek-VL2 MoE & \encircle[fill=harvestgold, text=white]{I} \space\encircle[fill=lightcoral, text=white]{T} & Rule-Outcome-Reward & GRPO & RL & -- & Multiple Tasks \\
        
        &\cellcolor[rgb]{ .949,  .949,  .949}MMR1 \cite{MMR1-Math2025} & \cellcolor[rgb]{ .949,  .949,  .949}Qwen2.5-VL-7B & \cellcolor[rgb]{ .949,  .949,  .949}\encircle[fill=harvestgold, text=white]{I} \space\encircle[fill=lightcoral, text=white]{T} & \cellcolor[rgb]{ .949,  .949,  .949}Rule-Outcome-Reward & \cellcolor[rgb]{ .949,  .949,  .949}GRPO & \cellcolor[rgb]{ .949,  .949,  .949}RL & \cellcolor[rgb]{ .949,  .949,  .949}6K & \cellcolor[rgb]{ .949,  .949,  .949}Multiple Math Tasks \\
        
        &MedVLM-R1\cite{pan2025medvlm} & Qwen2-VL-2B & \encircle[fill=harvestgold, text=white]{I} \space\encircle[fill=lightcoral, text=white]{T} & Rule-Outcome-Reward & GRPO & RL & 17.3K & Medical VQA \\
        
        &\cellcolor[rgb]{ .949,  .949,  .949}MM-EUREKA \cite{meng2025mm} & \cellcolor[rgb]{ .949,  .949,  .949}InternVL2.5-8B/38B & \cellcolor[rgb]{ .949,  .949,  .949}\encircle[fill=harvestgold, text=white]{I} \space\encircle[fill=lightcoral, text=white]{T} & \cellcolor[rgb]{ .949,  .949,  .949}Rule-Outcome-Reward & \cellcolor[rgb]{ .949,  .949,  .949}RLOO & \cellcolor[rgb]{ .949,  .949,  .949}RL & \cellcolor[rgb]{ .949,  .949,  .949}54K & \cellcolor[rgb]{ .949,  .949,  .949}Multiple Math Tasks\\
        
        &R1-Omni\cite{zhao2025r1} & HumanOmni-0.5B & \encircle[fill=harvestgold, text=white]{I} \space\encircle[fill=lightcoral, text=white]{T} \space\encircle[fill=DarkGreen, text=white]{V} & Rule-Outcome-Reward & GRPO & RL \& SFT & 0.58K/15.3K & Multimodal Emotion Recognition \\
        
        &\cellcolor[rgb]{ .949,  .949,  .949}R1-Onevision \cite{yang_r1-onevision_2025} & \cellcolor[rgb]{ .949,  .949,  .949}Qwen2.5-VL-3B/7B & \cellcolor[rgb]{ .949,  .949,  .949}\encircle[fill=harvestgold, text=white]{I} \space\encircle[fill=lightcoral, text=white]{T} & \cellcolor[rgb]{ .949,  .949,  .949}Rule-Outcome-Reward & \cellcolor[rgb]{ .949,  .949,  .949}GRPO & \cellcolor[rgb]{ .949,  .949,  .949}RL \& SFT & \cellcolor[rgb]{ .949,  .949,  .949}155K & \cellcolor[rgb]{ .949,  .949,  .949}Multiple Tasks\\
        
        &R1-VL\cite{zhang_r1-vl_2025} & Qwen2-VL-2B/7B & \encircle[fill=harvestgold, text=white]{I} \space\encircle[fill=lightcoral, text=white]{T} & Rule-Outcome-Reward & stepGRPO & RL \& SFT & 260K & Multiple Tasks \\
        
        &\cellcolor[rgb]{ .949,  .949,  .949}VisualThinker R1 Zero \cite{zhou_r1-zeros_2025} & \cellcolor[rgb]{ .949,  .949,  .949}Qwen2-VL-2B, Qwen-2-VL-2B-Instruct & \cellcolor[rgb]{ .949,  .949,  .949}\encircle[fill=harvestgold, text=white]{I} \space\encircle[fill=lightcoral, text=white]{T} & \cellcolor[rgb]{ .949,  .949,  .949}Rule-Outcome-Reward & \cellcolor[rgb]{ .949,  .949,  .949}GRPO & \cellcolor[rgb]{ .949,  .949,  .949}RL & \cellcolor[rgb]{ .949,  .949,  .949}218K & \cellcolor[rgb]{ .949,  .949,  .949}CVBench\\
        
        &LALM AQA\cite{li_reinforcement_2025} & Qwen2-Audio-7B-Instruc & \encircle[fill=harvestgold, text=white]{I} \space\encircle[fill=lightcoral, text=white]{T} \space\encircle[fill=DarkGreen, text=white]{V} & Rule-Outcome-Reward & GRPO & RL & 38k & MMAU \\
        
        &\cellcolor[rgb]{ .949,  .949,  .949}Seg-Zero \cite{liu_seg-zero_2025} & \cellcolor[rgb]{ .949,  .949,  .949}Qwen2.5-VL-3B & \cellcolor[rgb]{ .949,  .949,  .949}\encircle[fill=harvestgold, text=white]{I} \space\encircle[fill=lightcoral, text=white]{T} & \cellcolor[rgb]{ .949,  .949,  .949}Rule-Outcome-Reward & \cellcolor[rgb]{ .949,  .949,  .949}GRPO & \cellcolor[rgb]{ .949,  .949,  .949}RL & \cellcolor[rgb]{ .949,  .949,  .949}9K & \cellcolor[rgb]{ .949,  .949,  .949}RefCOCO, ReasonSeg\\
        
        &Skywork R1V\cite{skywork2025r1v} & InternViT-6B, DeepSeek-R1-Distill-Qwen-32B & \encircle[fill=harvestgold, text=white]{I} \space\encircle[fill=lightcoral, text=white]{T} & Rule-Outcome-Reward & GRPO & RL \& SFT & -- & Multiple Tasks \\
        
        &\cellcolor[rgb]{ .949,  .949,  .949}TimeZero \cite{wang_timezero_2025} & \cellcolor[rgb]{ .949,  .949,  .949}Qwen2.5-VL-7B & \cellcolor[rgb]{ .949,  .949,  .949}\encircle[fill=harvestgold, text=white]{I} \space\encircle[fill=lightcoral, text=white]{T} \space\encircle[fill=DarkGreen, text=white]{V} & \cellcolor[rgb]{ .949,  .949,  .949}Rule-Outcome-Reward & \cellcolor[rgb]{ .949,  .949,  .949}GRPO & \cellcolor[rgb]{ .949,  .949,  .949}RL & \cellcolor[rgb]{ .949,  .949,  .949}37.4K & \cellcolor[rgb]{ .949,  .949,  .949}Charades-STA, ActivityNet\\
        
        &Vision-R1\cite{huang_vision-r1_2025} & Qwen-2.5-VL-7B-Instruct & \encircle[fill=harvestgold, text=white]{I} \space\encircle[fill=lightcoral, text=white]{T} & Rule-Outcome-Reward & GRPO & RL \& SFT & 10K/200K & Multiple Tasks \\
        
        &\cellcolor[rgb]{ .949,  .949,  .949}Visual-RFT \cite{liu_visual-rft_2025} & \cellcolor[rgb]{ .949,  .949,  .949}Qwen2-VL-2B/7B & \cellcolor[rgb]{ .949,  .949,  .949}\encircle[fill=harvestgold, text=white]{I} \space\encircle[fill=lightcoral, text=white]{T} & \cellcolor[rgb]{ .949,  .949,  .949}Rule-Outcome-Reward & \cellcolor[rgb]{ .949,  .949,  .949}GRPO & \cellcolor[rgb]{ .949,  .949,  .949}RL & \cellcolor[rgb]{ .949,  .949,  .949}-- & \cellcolor[rgb]{ .949,  .949,  .949}Visual Perception Tasks\\
        
        &Reason-RFT\cite{tan_reason-rft_2025} & Qwen-2-VL-2B/7B & \encircle[fill=harvestgold, text=white]{I} \space\encircle[fill=lightcoral, text=white]{T} & Rule-Outcome-Reward & GRPO & RL & -- & Multiple Tasks \\
        
        &\cellcolor[rgb]{ .949,  .949,  .949}SEED-Bench-R1 \cite{liu_visual-rft_2025} & \cellcolor[rgb]{ .949,  .949,  .949}Qwen2-VL-Instruct-7B & \cellcolor[rgb]{ .949,  .949,  .949}\encircle[fill=harvestgold, text=white]{I} \space\encircle[fill=lightcoral, text=white]{T} \space\encircle[fill=DarkGreen, text=white]{V} & \cellcolor[rgb]{ .949,  .949,  .949}Rule-Outcome-Reward & \cellcolor[rgb]{ .949,  .949,  .949}GRPO & \cellcolor[rgb]{ .949,  .949,  .949}RL & \cellcolor[rgb]{ .949,  .949,  .949}50.2K & \cellcolor[rgb]{ .949,  .949,  .949}Video Understanding\\

        &STAR-R1\cite{STAR_R1}  & Qwen2.5-VL-7B &\encircle[fill=harvestgold, text=white]{I} \space\encircle[fill=lightcoral, text=white]{T} \space\encircle[fill=DarkGreen, text=white]{V} & Rule-Outcome-Reward & GRPO & RL \& SFT & 9K & Visual Reasoning \\
        \midrule
        
        \multicolumn{9}{c}{\textbf{Analysis RFT Project}} \\ 
        
        \midrule
        
        \multirow{3}{*}{LLM}&Demystify-LongCoT\cite{yeo2025demystifying} & Llama-3.1-8B, Qwen2.5
-7B-Math & \encircle[fill=lightcoral, text=white]{T} & Rule-Outcome-Reward  & PPO/Reinforce++ &  RL \& SFT  & 7.5K & Multiple Math, MMLU \\ 

         &\cellcolor[rgb]{ .949,  .949,  .949}RLHF-Scale\cite{RLHF_Scaling} & \cellcolor[rgb]{ .949,  .949,  .949}GLM4-9B & \cellcolor[rgb]{ .949,  .949,  .949}\encircle[fill=lightcoral, text=white]{T} & \cellcolor[rgb]{ .949,  .949,  .949}Process-Reward  & \cellcolor[rgb]{ .949,  .949,  .949}PPO & \cellcolor[rgb]{ .949,  .949,  .949}RL  & \cellcolor[rgb]{ .949,  .949,  .949}11K & \cellcolor[rgb]{ .949,  .949,  .949}Multiple Tasks  \\ 
        
        &MD-CoT~\cite{metastabledynamicschainofthoughtreasoning} & -- & -- & --  & -- & --   & -- & --  \\ 

        &\cellcolor[rgb]{ .949,  .949,  .949}GraidientUnified~\cite{li2025instructionreasoningdatashape}  & \cellcolor[rgb]{ .949,  .949,  .949}-- & \cellcolor[rgb]{ .949,  .949,  .949}-- & \cellcolor[rgb]{ .949,  .949,  .949}--  & \cellcolor[rgb]{ .949,  .949,  .949}-- & \cellcolor[rgb]{ .949,  .949,  .949}--  & \cellcolor[rgb]{ .949,  .949,  .949}-- & \cellcolor[rgb]{ .949,  .949,  .949}-- \\
        
        \bottomrule
   \end{tabular}}
\label{table:rl_supervise}
\end{table*}

Despite the strengths of RFT, it still faces the following challenges:

\begin{enumerate}[itemindent=0em]

\item \textbf{Unclear Mechanism behind Reasoning:} The underlying mechanisms driving the reasoning improvements in DeepSeek-R1 remain poorly understood. 
For example, while DeepSeek-R1 exhibits emergent properties (\emph{e.g.}, ``Emergent Length Increasing'', ``Aha moments''), studies such as \cite{ThereMaybeNot} suggest that capabilities like Long-CoT might already exist in the base model, rather than solely emerging from RL training. 
Furthermore, performance gains observed in smaller models (\emph{e.g.}, Qwen-Math-2B/7B \cite{qwen2}) occur without noticeable ``Aha moments'', complicating causal interpretations.

\item \textbf{Reward Model Saturation:} Many existing RL algorithms face reward model saturation, typically manifested as exploration collapse after around 100 training steps. 
Although DeepSeek-R1 alleviates this issue through specialized reward formatting, methods like ReFT \cite{trung2024reft} and Satori \cite{Satori} propose alternating sampling and SFT distillation to combat reward hacking and exploration collapse.

\item \textbf{Unstable Long-CoT Generation:} Long reasoning chains generated by RFT are prone to instability, including context overflow, failure to return final answers, and sensitivity to reward shaping \cite{Tecent_2_plus_3}. 
For instance, methods like \cite{yeo2025demystifying} inadvertently introduce cosine reward functions, which degrade performance with increased iterations. 
O1-Prune \cite{o1_pruner} uses post-hoc length pruning techniques \cite{team2025kimi} (via RL/SFT) to stabilize outputs. 
Recent VL models such as Open-R1-Video \cite{open-r1-video} and Seg-Zero \cite{liu_seg-zero_2025} show instability in chain generation when handling long videos or fine-grained segmentation reasoning.

\end{enumerate}

\noindent \textbf{Reinforcement Learning without External Reward}. 
Reinforcement learning without explicit answer-based reward signals has recently been shown to be effective even in the absence of external rewards. 
INTUITOR~\cite{zhao2025learning} and RENT~\cite{shafayat2025can} propose leveraging the model’s confidence—specifically, the entropy of its self-generated responses—as an intrinsic reward signal during RL training, thereby utilizing internal feedback rather than external supervision. 
Genius~\cite{xu2025genius} demonstrates that using uncertainty from future sampling steps as a reward can further enhance model reasoning.
EMPO~\cite{zhang2025right} utilizes semantic clusters~\cite{farquhar2024detecting} to normalize rewards within groups of responses, thereby minimizing predictive entropy.
From another perspective, self-consistency-based methods~\cite{shafayat2025can,zuo2025ttrl} find that using self-consistent answers as pseudo-gold labels in RL enables self-evolution, and \cite{zuo2025ttrl} further demonstrates the effectiveness of self-consistency-based RL training for test-time scaling. Similarly, MM-UPT~\cite{wei2025unsupervised} introduces a voting-based reward for unsupervised RL training of MLLMs.
More surprisingly,~\cite{shao2025spurious} shows that a model's reasoning can be substantially improved using "spurious rewards," such as random or even completely incorrect rewards.
However, these alternative approaches still yield inferior results compared to reward signals from golden answers and exhibit limited compatibility with answer-based rewards, indicating that further exploration is needed.

\noindent \textbf{Entropy in RL Exploration}. Since the exploration plays a crucial role in RL~\cite{sutton1998reinforcement}, adequate exploration is necessary to achieve optimal performance.
DAPO~\cite{yu2025dapo} finds that increasing the clip-high ratio in GRPO helps prevent entropy collapse during training.
Skywork-OR1~\cite{he2025skywork} demonstrates that strategically tuning sampling-related parameters and introducing an entropy loss help maintain diversity during long-term RL training. 
\cite{gao2025one} states that entropy minimization (EM) shares the same objective as RLVR in unlocking the pretrained model’s latent potential, and shows that a single unsupervised sample in EM training can elicit reasoning in LLMs.
\cite{cui2025entropy} presents an empirical transformation between the entropy and model performance during RLVR training and proposes a covariance-based entropy control strategy. 
Nevertheless, research on the role of entropy during RL training remains empirical, and deeper theoretical analysis is urgently needed.\\ 


Future directions for RFT may include several exciting and innovative advancements, such as:

\begin{enumerate}[itemindent=0em]

\item \textbf{Efficient and Stable RL Frameworks:} There is a need to develop more robust RL algorithms that prevent reward saturation and exploration collapse. \cite{yeo2025demystifying} reveals that REINFORCE++ \cite{reinforce_plusplus} underperforms when combined with KL divergence regularization, suggesting the need for alternative methods. 
Future work should revisit classic RL algorithms in the context of modern LLMs training to optimize both stability and efficiency.

\item \textbf{Scaling RFT:} Current RL-Supervise models rely on curated, verifiable prompts selected from large-scale datasets. 
Future research should focus on synthesizing high-quality, diverse prompts to improve generalization. \cite{RLHF_Scaling} shows that merely scaling policy/reward models or increasing sample sizes results in diminishing returns, while expanding the scope of PRM and R1 training data holds greater promise. 
Hybrid approaches, such as combining RL with SFT or curriculum learning, should be explored to enhance scalability.

\item \textbf{Controlling Long-CoT Stability:} Adaptive reward shaping mechanisms are needed to balance reasoning length, coherence, and answer correctness. 
Techniques such as O1-Prune \cite{o1_pruner} demonstrate the value of post-hoc length regularization, but dynamic in-training controls are necessary. 
Hierarchical RL frameworks should be investigated to decompose long reasoning chains into manageable sub-tasks, reducing instability.

\item \textbf{Theoretical and Empirical Analysis:} It is essential to clarify the relationship between RL training and the capabilities of the base model. 
For instance, it should be determined whether emergent properties (\emph{e.g.}, Long-CoT) arise from RL optimization or are latent traits of the base model. 
However, this phenomenon of Long-CoT is often difficult to appear in multimodal situations (\emph{e.g.}, \cite{chen2025r1v,lmmr1,zheng2025easyr1, spatial_reason}). Systematic studies on reward design principles (\emph{e.g.}, sparse vs. dense rewards, multi-objective balancing) should be conducted to avoid unintended behaviors such as reward hacking.

\end{enumerate}

\noindent\textbf{Summary:} RFT presents a promising direction for advancing LLMs reasoning, as evidenced by DeepSeek-R1 \cite{Deepseek-R1}. 
However, challenges such as reward saturation, unstable long reasoning chains, and unclear emergent mechanisms require urgent attention. 
Future efforts should prioritize algorithmic innovation, scalable prompt synthesis, and theoretical grounding to fully unlock the potential of RL-driven reasoning LLMs.

\subsection{Evolutionary of Reasoning LLMs}\label{evolutionary}


The evolution of reasoning LLMs has progressed by several distinct stages, with various strategies developed to overcome the limitations of direct autoregressive inference and build more advanced slow-thinking reasoning architectures.

\definecolor{mygray}{rgb}{.949,  .949,  .949}

\begin{table*}[tbhp]
\scriptsize
\centering
\caption{Statistics of benchmarks for reasoning LLMs.}
\resizebox{0.95\linewidth}{!}{
\begin{tabular}{lllllll}
\toprule[1.2pt]
\textbf{Domain}                  & \textbf{Benchmark}                                                            & \textbf{Question Type} & \textbf{Venue}      & \textbf{Language}        & \textbf{Size}   & \textbf{Level}              \\
\hline
\multirow{17}{*}{Math}         

& \cellcolor[rgb]{ .949,  .949,  .949}AIME 2024 \cite{AIME2024}                        &  \cellcolor[rgb]{ .949,  .949,  .949}Open-End & \cellcolor[rgb]{ .949,  .949,  .949}-          & \cellcolor[rgb]{ .949,  .949,  .949}English         &\cellcolor[rgb]{ .949,  .949,  .949}30     & \cellcolor[rgb]{ .949,  .949,  .949}Competition        \\

& BBH \cite{suzgun2022challenging} & Hybrid & Findings ACL 2023 & English & 23 & Challenging \\

&  \cellcolor{mygray}MATH-500 \cite{lightmanlet}  & \cellcolor{mygray}Open-End & \cellcolor{mygray} ICLR 2024 & \cellcolor{mygray}English & \cellcolor{mygray}500  & \cellcolor{mygray}Competition        \\

& AMC 2023 \cite{AMC2023}                          & Open-End & --   & English         & 30     & Competition        \\

&  \cellcolor{mygray}Olympiad Bench   \cite{he2024olympiadbench}    & \cellcolor{mygray}Open-End  &  \cellcolor{mygray}ACL 2024     & \cellcolor{mygray}English/Chinese &  \cellcolor{mygray}8,476  & \cellcolor{mygray}Competition        \\

& Putnam-AXIOM \cite{gulati2024putnam} & Open-End &  NeurIPS 2024 & English & 236 & Competition \\

&  \cellcolor{mygray}PRM800K \cite{lightman2023let} & \cellcolor{mygray}Open-End & \cellcolor{mygray} ICLR 2024 & \cellcolor{mygray}English & \cellcolor{mygray}800,000 & \cellcolor{mygray}Hybrid \\

& FrontierMath \cite{glazer2024frontiermath} & Open-End & ArXiv 2024 & English & - & Expert \\

&  \cellcolor{mygray}ProcessBench \cite{zheng2024processbench} & \cellcolor{mygray}Open-End & \cellcolor{mygray} ArXiv 2024 & \cellcolor{mygray}English & \cellcolor{mygray}3400 & \cellcolor{mygray}Competition \\

& LiveBench \cite{white2025livebench} & Open-End & ICLR 2025 & English & Frequently Updated & Expert \\

&  \cellcolor{mygray}AIME 2025  & \cellcolor{mygray}Open-End & \cellcolor{mygray} Hugging Face & \cellcolor{mygray}English & \cellcolor{mygray}13 & \cellcolor{mygray}Competition \\

& ThinkBench \cite{huang2025thinkbench} & Hybrid & ArXiv 2025 & English & 2,912 & Expert \\

&  \cellcolor{mygray}MATH-Perturb \cite{huang2025math} & \cellcolor{mygray}Open-End & \cellcolor{mygray} ArXiv 2025 & \cellcolor{mygray}English & \cellcolor{mygray}279 & \cellcolor{mygray}Competition \\

& ZebraLogic \cite{lin2025zebralogic} & Open-End & ArXiv 2025 & English & 1,000 & Hybrid \\

&  \cellcolor{mygray}QuestBench \cite {li2025questbench} & \cellcolor{mygray}Choice & \cellcolor{mygray} ArXiv 2025 & \cellcolor{mygray}English & \cellcolor{mygray}38,882 & \cellcolor{mygray}Hybrid \\

& Math-RoB \cite {yu2025benchmarking} & Open-End & ArXiv 2025 & English & - & High School \\

&  \cellcolor{mygray}GSM-Ranges \cite {shrestha2025mathematical} & \cellcolor{mygray}Open-End & \cellcolor{mygray} ArXiv 2025 & \cellcolor{mygray}English & \cellcolor{mygray}- & \cellcolor{mygray}Middle School \\

\hline

\multirow{5}{*}{Code}          

& Codeforces                                         & Open-End   & -          & English         & -      &Expert             \\

& \cellcolor{mygray}CodeContests \cite{li2022competition} & \cellcolor{mygray}Open-End & \cellcolor{mygray}Science 2022 & \cellcolor{mygray}English & \cellcolor{mygray}13,610 & \cellcolor{mygray}Competition \\

&  SWE-bench \cite{jimenez2024swebench} & Open-End  &  ICLR 2024    &  English         &  2,294  &  Expert             \\

& \cellcolor[rgb]{ .949,  .949,  .949}LiveCodeBench   \cite{jain2024livecodebench}     & \cellcolor[rgb]{ .949,  .949,  .949}Open-End   &\cellcolor[rgb]{ .949,  .949,  .949}ArXiv 2024   & \cellcolor[rgb]{ .949,  .949,  .949}English         & \cellcolor[rgb]{ .949,  .949,  .949}-      & \cellcolor[rgb]{ .949,  .949,  .949}Expert             \\

& CodeCriticBench \cite {zhang2025codecriticbench} & Hybrid & ArXiv 2025 & English & - & Expert \\

\hline

\multirow{17}{*}{Science} 
&  \cellcolor{mygray}GPQA Diamond \cite{rein2024gpqa}   & \cellcolor{mygray}Choice &  \cellcolor{mygray}COLM 2024    &  \cellcolor{mygray}English         & \cellcolor{mygray}448    &  \cellcolor{mygray}University  \\

& MR-Ben \cite{ZengLWLCDYXQZSL24}        & Hybrid & NeurIPS 2024 & English         & 5,975 & Hybrid             \\

 & \cellcolor[rgb]{ .949,  .949,  .949}MMLU-Pro \cite{wang2024mmlu} & \cellcolor[rgb]{ .949,  .949,  .949}Choice  & \cellcolor[rgb]{ .949,  .949,  .949}NeurIPS 2024 & \cellcolor[rgb]{ .949,  .949,  .949}English         & \cellcolor[rgb]{ .949,  .949,  .949}12,032 & \cellcolor[rgb]{ .949,  .949,  .949}Hybrid             \\

& MHPP \cite{dai2024mhpp} & Open-End & ArXiv 2024 & English & 210 & Expert \\

& \cellcolor{mygray}RewardBench \cite {lambert2024rewardbench} & \cellcolor{mygray}Hybrid & \cellcolor{mygray}ArXiv 2024 & \cellcolor{mygray}English & \cellcolor{mygray}- & \cellcolor{mygray}Hybrid \\

& MR-Ben \cite{zeng2024mr} & Open-End & NeurIPS 2024 & English & 5,975 & Hybrid \\

& \cellcolor{mygray}ReaLMistake \cite {kamoi2024evaluating} & \cellcolor{mygray}Open-End & \cellcolor{mygray}COLM 2024 & \cellcolor{mygray}English & \cellcolor{mygray}- & \cellcolor{mygray}Expert \\

& CriticBench \cite {lin-etal-2024-criticbench} & Open-End & ACL 2024 & English & - & Hybrid \\

& \cellcolor{mygray}JudgeBench \cite {tan2024judgebench} & \cellcolor{mygray}Open-End & \cellcolor{mygray}ArXiv 2024 & \cellcolor{mygray}English & \cellcolor{mygray}- & \cellcolor{mygray}Hybrid \\

& TPBench \cite{chung2025theoretical} & Open-End & ArXiv 2025 & English & 57 & University \\

& \cellcolor{mygray}ProBench \cite{yang2025probench} & \cellcolor{mygray}Open-End & \cellcolor{mygray}ArXiv 2025 & \cellcolor{mygray}English/Chinese & \cellcolor{mygray}790 & \cellcolor{mygray}Competition \\

& EquiBench \cite{wei2025equibench} & Open-End & ArXiv 2025 & English & 2,400 & Hybrid \\

& \cellcolor{mygray}SuperGPQA \cite {du2025supergpqa} & \cellcolor{mygray}Choice & \cellcolor{mygray}ArXiv 2025 & \cellcolor{mygray}English & \cellcolor{mygray}26,529 & \cellcolor{mygray}University \\

& Sys2Bench \cite {parashar2025inference} & Open-End & ArXiv 2025 & English & - & Hybrid \\

& \cellcolor{mygray}PRMBench \cite {song2025prmbench} & \cellcolor{mygray}Open-End & \cellcolor{mygray}ArXiv 2025 & \cellcolor{mygray}English & \cellcolor{mygray}6,216 & \cellcolor{mygray}Expert \\

& DeltaBench \cite {he2025can} & Open-End & ArXiv 2025 & English & - & Expert \\

& \cellcolor{mygray}FINEREASON \cite {chen2025finereason} & \cellcolor{mygray}Open-End & \cellcolor{mygray}ArXiv 2025 & \cellcolor{mygray}English & \cellcolor{mygray}- & \cellcolor{mygray}Expert \\

\hline

\multirow{13}{*}{Agent}   
&  ARC \cite{chollet2019measure} & Open-End & ArXiv 2019 & Symbolic & 1,000 & Expert \\
&  \cellcolor{mygray}WebShop \cite{yao2022webshop}  & \cellcolor{mygray}Open-End &  \cellcolor{mygray}NeurIPS 2022 &  \cellcolor{mygray}English  &  \cellcolor{mygray}1,600  & \cellcolor{mygray}Hybrid \\

&  SciWorld   \cite{chan2025mlebenchevaluatingmachinelearning}  & Open-End &  EMNLP 2022   & English   & 7,200  &  Hybrid  \\

& \cellcolor[rgb]{ .949,  .949,  .949}WebArena  \cite{zhou2023webarena} & \cellcolor[rgb]{ .949,  .949,  .949}Open-End  & \cellcolor[rgb]{ .949,  .949,  .949}ICLR 2024    & \cellcolor[rgb]{ .949,  .949,  .949}English         & \cellcolor[rgb]{ .949,  .949,  .949}812    & \cellcolor[rgb]{ .949,  .949,  .949}Hybrid             \\

& TextCraft   \cite{prasad2024adapt} & Open-End  & NAACL 2024   & English         & 200    & Hybrid             \\

& \cellcolor{mygray}Osworld \cite{xie2024osworld} & \cellcolor{mygray}Open-End & \cellcolor{mygray}NeurIPS 2024 & \cellcolor{mygray}English & \cellcolor{mygray}369 & \cellcolor{mygray}Hybrid \\

& GAMABench \cite{huang2024far} & Open-End & ArXiv 2024 & English & - & Hybrid \\

& \cellcolor{mygray}Mle-bench \cite{scienceworld2022} & \cellcolor{mygray}Open-End & \cellcolor{mygray}ArXiv 2025 & \cellcolor{mygray}English & \cellcolor{mygray}- & \cellcolor{mygray}Competition \\

& ToolComp \cite{nath2025toolcomp} & Open-End & ArXiv 2025 & English & - & Hybrid \\

& \cellcolor{mygray}Mobile-Agent-E \cite{wang2025mobile} & \cellcolor{mygray}Open-End & \cellcolor{mygray}ArXiv 2025 & \cellcolor{mygray}English & \cellcolor{mygray}25 & \cellcolor{mygray}Hybrid \\

& Text2World \cite{zhang2025physreason} & Open-End & ArXiv 2025 & English & - & Hybrid \\

& \cellcolor{mygray}WebGames \cite{thomas2025webgames} & \cellcolor{mygray}Open-End & \cellcolor{mygray}ArXiv 2025 & \cellcolor{mygray}English & \cellcolor{mygray}50 & \cellcolor{mygray}Hybrid \\

& Ui-r1 \cite{lu2025ui} & Open-End & ArXiv 2025 & English & 136 & Hybrid \\

\hline
\multirow{5}{*}{Medicine}      
& \cellcolor{mygray}JAMA Clinical \cite{chen2024benchmarking} & \cellcolor{mygray}Choice & \cellcolor{mygray}NAACL 2025 & \cellcolor{mygray}English & \cellcolor{mygray}1,524 & \cellcolor{mygray}Expert \\

& Medbullets \cite{chen2024benchmarking} & Choice & NAACL 2025 & English & 308 & Expert \\

& \cellcolor{mygray}MedQA \cite{jin2021disease} & \cellcolor{mygray}Choice & \cellcolor{mygray}ArXiv 2020 & \cellcolor{mygray}English/Chinese & \cellcolor{mygray}61,097 & \cellcolor{mygray}Expert \\

& MEDEC \cite{abacha2024medec} & Open-End & ArXiv 2024 & English & 3,848 & Expert \\

& \cellcolor{mygray}MedXpertQA \cite{zuo2025medxpertqa} & \cellcolor{mygray}Choice & \cellcolor{mygray}ArXiv 2025 & \cellcolor{mygray}English & \cellcolor{mygray}4,460 & \cellcolor{mygray}Expert \\

\bottomrule[1.2pt]
\end{tabular}
}
\label{table:benchmark_categories}
\end{table*}


\begin{table*}[tbhp]
\scriptsize
\centering
\caption{Statistics of benchmarks for reasoning MLLMs.}
\resizebox{0.95\linewidth}{!}{
\begin{tabular}{lllllll}
\toprule[1.2pt]
\textbf{Domain}                  & \textbf{Benchmark}                                                            & \textbf{Question Type} & \textbf{Venue}      & \textbf{Language}        & \textbf{Size}   & \textbf{Level}              \\
\hline
\multirow{31}{*}{Multimodality} 
& \cellcolor[rgb]{ .949,  .949,  .949}MMMU \cite{yue2024mmmu} & \cellcolor[rgb]{ .949,  .949,  .949}Hybrid & \cellcolor[rgb]{ .949,  .949,  .949}CVPR 2024    & \cellcolor[rgb]{ .949,  .949,  .949}English         & \cellcolor[rgb]{ .949,  .949,  .949}11,500 & \cellcolor[rgb]{ .949,  .949,  .949}Hybrid             \\

&  MathVista   \cite{lu2024mathvista}  & Hybrid &  ICLR 2024    &  English         &  6,141  &  Middle School      \\

& \cellcolor[rgb]{ .949,  .949,  .949}MathVision \cite{MathVision}   & \cellcolor[rgb]{ .949,  .949,  .949}Hybrid & \cellcolor[rgb]{ .949,  .949,  .949}NeurIPS 2024 & \cellcolor[rgb]{ .949,  .949,  .949}English         & \cellcolor[rgb]{ .949,  .949,  .949}3,040  & \cellcolor[rgb]{ .949,  .949,  .949}Middle/High School \\

&  CMMaTH \cite{li2024cmmath}  & Hybrid &  COLING 2025  & English/Chinese &  23,856 & Middle/High School \\

& \cellcolor[rgb]{ .949,  .949,  .949}PGPS9K \cite{Zhang2023PGPS}  & \cellcolor[rgb]{ .949,  .949,  .949}Hybrid & \cellcolor[rgb]{ .949,  .949,  .949}IJCAI 2023   & \cellcolor[rgb]{ .949,  .949,  .949}English         & \cellcolor[rgb]{ .949,  .949,  .949}9,023  & \cellcolor[rgb]{ .949,  .949,  .949}Middle School  \\

&  ZeroBench \cite{roberts2025zerobench}  & Open-End &  ArXiv 2025  & English & 100/334 & Impossible  \\

& \cellcolor[rgb]{ .949,  .949,  .949}MME-CoT \cite{jiang2025mme}  & \cellcolor[rgb]{ .949,  .949,  .949}Hybrid & \cellcolor[rgb]{ .949,  .949,  .949}ArXiv 2025   & \cellcolor[rgb]{ .949,  .949,  .949}English         & \cellcolor[rgb]{ .949,  .949,  .949}1,130  & \cellcolor[rgb]{ .949,  .949,  .949}Hybrid  \\ 

&  MM-IQ \cite{cai2025mm}  & Choice &  ArXiv 2025  & English/Chinese & 2,710 & Hybrid  \\

& \cellcolor[rgb]{ .949,  .949,  .949}Multimodal RewardBench \cite{yasunaga2025multimodal}  & \cellcolor[rgb]{ .949,  .949,  .949}Hybrid & \cellcolor[rgb]{ .949,  .949,  .949}ArXiv 2025   & \cellcolor[rgb]{ .949,  .949,  .949}English         & \cellcolor[rgb]{ .949,  .949,  .949}5,211  & \cellcolor[rgb]{ .949,  .949,  .949}Hybrid  \\ 

&  GRAB \cite{roberts2024grab}  & Open-End &  ArXiv 2024  & English & 2,710 & Hybrid  \\

& \cellcolor[rgb]{ .949,  .949,  .949}SciFIBench \cite{roberts2024scifibench}  & \cellcolor[rgb]{ .949,  .949,  .949}Choice & \cellcolor[rgb]{ .949,  .949,  .949}ArXiv 2024   & \cellcolor[rgb]{ .949,  .949,  .949}English         & \cellcolor[rgb]{ .949,  .949,  .949}2,000  & \cellcolor[rgb]{ .949,  .949,  .949}Challenging  \\ 

& MV-MATH \cite{mv-math}  & Hybrid & ArXiv 2025   & English         & 2,009  & Middle/High School  \\ 

& \cellcolor{mygray}ScienceQA \cite{lu2022learn} & \cellcolor{mygray}Choice & \cellcolor{mygray}NeurIPS 2022 & \cellcolor{mygray}English & \cellcolor{mygray}21,000 & \cellcolor{mygray}Hybrid \\

& Plot2Code \cite{wu2024plot2code} & Open-End & ArXiv 2024 & English & 132 & Hybrid \\

& \cellcolor{mygray}M3CoT \cite{chen2024m} & \cellcolor{mygray}Open-End & \cellcolor{mygray}ArXiv 2024 & \cellcolor{mygray}English & \cellcolor{mygray}- & \cellcolor{mygray}Hybrid \\

& PUZZLEVQA \cite{chia2024puzzlevqa} & Hybrid & ACL 2024 & English & - & Hybrid \\

& \cellcolor{mygray}MolPuzzle \cite{guo2024can} & \cellcolor{mygray}Hybrid & \cellcolor{mygray}NeurIPS 2024 & \cellcolor{mygray}English & \cellcolor{mygray}23,000 & \cellcolor{mygray}Expert \\

& HumanEval-V \cite{zhang2024humaneval} & Open-End & ArXiv 2024 & English & 108 & Hybrid \\

& \cellcolor{mygray}CoMT \cite{cheng2025comt} & \cellcolor{mygray}Choice & \cellcolor{mygray}AAAI 2025 & \cellcolor{mygray}English & \cellcolor{mygray}3,853 & \cellcolor{mygray}Hybrid \\

& ChartMimic \cite{yang2024chartmimic} & Open-End & ArXiv 2024 & English & 4,800 & Expert \\

& \cellcolor{mygray}OlympicArena \cite{huang2024olympicarena} & \cellcolor{mygray}Hybrid & \cellcolor{mygray}NeurIPS 2024 & \cellcolor{mygray}English/Chinese & \cellcolor{mygray}11,163 & \cellcolor{mygray}Competition \\

& CVQA\&CPVQA \cite{wang2025can} & Open-End & ArXiv 2025 & English & 1664 & Hybrid \\

& \cellcolor{mygray}ENIGMAEVAL \cite{wang2025enigmaeval} & \cellcolor{mygray}Open-End & \cellcolor{mygray}ArXiv 2025 & \cellcolor{mygray}English & \cellcolor{mygray}1184 & \cellcolor{mygray}Challenging \\

& CODE-VISION \cite{wang2025code} & Open-End & ArXiv 2025 & English & 438 & Competition \\

& MKRC \cellcolor{mygray}\cite{jia2025exploring} & \cellcolor{mygray}Open-End & \cellcolor{mygray}ArXiv 2025 & \cellcolor{mygray}English & \cellcolor{mygray}7010 & \cellcolor{mygray}Hybrid \\

& MMSciBench \cite{ye2025mmscibench} & Hybrid & ArXiv 2025 & English/Chinese & 4,482 & High School \\

& \cellcolor{mygray}LEGO-Puzzles \cite{tang2025lego} & \cellcolor{mygray}Hybrid & \cellcolor{mygray}ArXiv 2025 & \cellcolor{mygray}English & \cellcolor{mygray}1,100 & \cellcolor{mygray}Expert \\

& JustLogic \cite{chen2025justlogic} & Open-End & ArXiv 2025 & English & 7,000 & Expert \\

& \cellcolor{mygray}HUMANITY'S LAST EXAM \cite{phan2025humanity} & \cellcolor{mygray}Hybrid & \cellcolor{mygray}ArXiv 2025 & \cellcolor{mygray}English & \cellcolor{mygray}2,700 & \cellcolor{mygray}Challenging \\

& DivIL \cite{jiaqidivil} & Hybrid & TMLR & English & - & Hybrid \\

& \cellcolor{mygray}ErrorRadar \cite{yan2024errorradar} & \cellcolor{mygray}Open-End & \cellcolor{mygray}ArXiv 2024 & \cellcolor{mygray}English & \cellcolor{mygray}2,500 & \cellcolor{mygray}Middle/High School \\

\bottomrule[1.2pt]
\end{tabular}
}
\label{table:benchmark_categories_multimodal}
\end{table*}

In the early stages, reasoning LLMs primarily focused on enhancing pre-trained LLMs with external reasoning algorithms, without altering the underlying model parameters. Approaches such as Tree of Thoughts \cite{Tree_of_Thought, Thought-Propagation} and Reasoning via Planning \cite{hao2023reasoning} utilized LLMs-driven Breadth-First Search, Depth-First Search, and MCTS \cite{Renda_Report_Tree_Search, browne2012survey, DBLP:journals/corr/abs-2410-16033, DBLP:journals/corr/abs-2412-09078} to simulate human-like reasoning processes. 
These methods represented reasoning as tree or graph traversals, where intermediate reasoning states were depicted as nodes, and various reasoning strategies produced distinct reasoning paths. The final decision was made through additional voting mechanisms \cite{wangself} or Monte Carlo-based value estimation to identify the optimal path.

However, these externalized slow-reasoning approaches introduced several challenges:
\begin{enumerate}[itemindent=0em]

\item \textbf{Limited Exploration Space:} The search-based methods required predefined constraints on the breadth, depth, and granularity of the search space, which often restricted the LLM's exploration to a narrow reasoning space. Furthermore, the reasoning strategies across different child nodes of the same parent node frequently lacked sufficient diversity, further limiting exploration.

\item \textbf{Limited Experience Sharing:} Exploration experiences and reasoning information across different paths could only be assessed based on reward models or self-consistency among outcomes. Additionally, search-based methods significantly increased computational overhead, relying on reward models such as PRM/ORM for tree pruning or speculative decoding techniques to accelerate inference.

\end{enumerate}
To overcome these limitations, subsequent models such as rSTaR \cite{rSTaR}, LLaMAV-o1 \cite{thawakar2025llamav}, HiICL-MCTS \cite{HiICL-MCTS}, Mulberry \cite{yao2024mulberry}, g1 \cite{g1}, and Thinking-Claude \cite{Thinking-Claude} introduced richer action spaces. 
These enhanced action spaces offered high-level planning cues, broadening the model's exploration scope and enabling more comprehensive structured search processes. 
However, this approach necessitated careful design of the action spaces to ensure their effectiveness.

With the introduction of models like o1 \cite{openai_o1} and QwQ \cite{qwq-32b-preview}, external reasoning paradigms were internalized within the LLM's context. 
These models initially performed exploratory macro-planning to generate an initial reasoning path, followed by contextual exploration of alternative paths. 
Through mechanisms like ``Rethink'' and ``Verification'', these models produced extended reasoning chains. 
To replicate this internalized capability, STILL-1 \cite{Renda_Report_Tree_Search} linearized tree search outputs into long reasoning chains with attributes such as ``Rethink'', ``Wait'', and ``Explore New Path''. 
Similarly, STILL-2 \cite{Slow_Thinking_with_LLMs_2} and sky-T1 \cite{sky_t1_2025} synthesized long reasoning chains using distillation techniques. 
Approaches like Virgo \cite{virgo} have attempted to distill text-based slow-thinking reasoning into multimodal LLMs; their performance improvements in tasks such as MathVision \cite{MathVision}, which demand detailed visual understanding, have been marginal.
However, the linearized reasoning chains derived from search-based methods struggled to match the quality of those produced by distillation approaches. 
Additionally, extending slow-thinking reasoning capabilities from text-based domains to multimodal contexts remains a significant challenge, especially in tasks requiring fine-grained perception \cite{SlowPerception, vll_app, li2023lans, CapArena}.


Recent advancements, including DeepSeek-R1 \cite{Deepseek-R1} and Kimi-k1.5 \cite{team2025kimi}, have demonstrated the potential of RL to enhance models like DeepSeek-V3 \cite{liu2024deepseek}, resulting in the emergence of complex behaviors such as long reasoning chains, reflective reasoning, and advanced planning capabilities. 
Remarkably, these sophisticated behaviors were achieved through simple RL scaling. 
SimpleRL \cite{simplerl_reason_blob} sought to replicate these capabilities using a streamlined pipeline and minimal codebase, while R1V \cite{chen2025r1v} explored the development of multimodal reasoning models based on multimodal foundation architectures. 
However, though R1 has been proven to significantly enhance reasoning abilities in the LLMs field, there remain many challenges to explore in the MLLMs domain. These include difficulties in maintaining a consistent slow thinking process when handling complex visual inputs and achieving training benefits comparable to those acquired through unimodal RL.

\noindent\textbf{Summary:} The evolution of reasoning LLMs has shifted from externally augmented reasoning to internally embedded reasoning. 
Recent developments emphasize the potential of RL-based scaling to unlock advanced capabilities.

\begin{figure*}[t]
    \centering
    \includegraphics[width=0.95\linewidth]{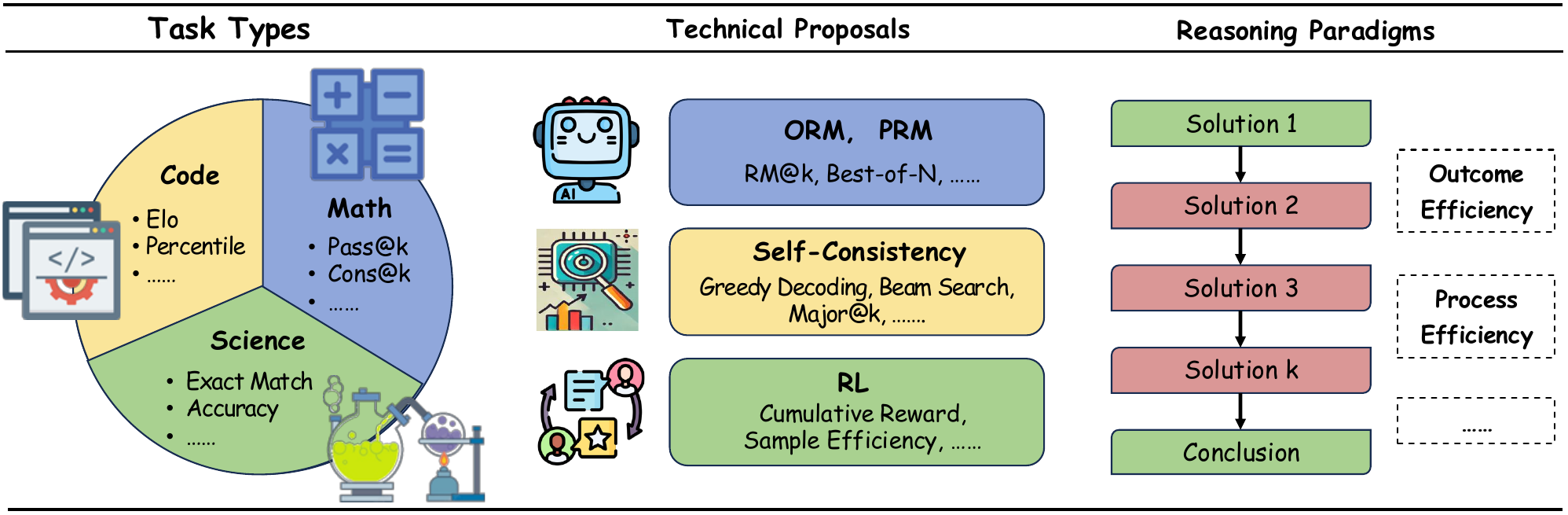}
    \caption{Various evaluation metrics of reasoning LLMs divided by task types, technical proposals, and reasoning paradigms.}
    \label{fig:evaluation_metrics}
\end{figure*}

\section{Benchmarking Reasoning LLMs}\label{benchmark}

The development of a robust benchmark is crucial for documenting the advancements in reasoning LLMs capabilities and for identifying promising research directions for future progress. 
Here, we review the benchmarks from three key aspects: categories, evaluation metrics, and performance comparisons, while offering our reflections and insights.

\subsection{Benchmark Categories}\label{benchmark_category}

We categorize reasoning benchmarks by task type, which can be broadly divided into math, code, scientific, agent \cite{Foundation-Agents-Survey}, medical, and multimodal reasoning. The detailed statistics for these benchmarks are presented in Table \ref{table:benchmark_categories}.

\subsubsection{Benchmark Introduction} 

\begin{enumerate}[itemindent=0em]
\item \textbf{Math Problems:} We document the current popular competition-level mathematical benchmarks to showcase the capabilities of reasoning LLMs, including AIME 2024 \cite{AIME2024}, MATH-500 \cite{lightmanlet}, AMC 2023 \cite{AMC2023}, and Olympiad Bench \cite{he2024olympiadbench}.

\item \textbf{Code Problems:} Code problems \cite{sun2023corex, sun2024survey_code} requires solid foundation and high logical thinking to evaluate the reasoning ability of reasoning LLMs such as \href{https://codeforces.com/}{Codeforces}, SWE-bench \cite{jimenez2024swebench}, and LiveCodeBench \cite{jain2024livecodebench}.

\item \textbf{Scientific Problems:} Scientific benchmarks, \emph{i.e.}, GPQA Diamond \cite{rein2024gpqa} and MMLU-Pro \cite{wang2024mmlu}, involve multi-domains reasoning about chemistry, biology, and physics, which requires extensive knowledge accumulation and integrated reasoning. 

\item \textbf{Agent Reasoning:} Realistic tasks often involve complex planning and tool usage, leading to the creation of agent reasoning benchmarks \cite{xi2024agentgym}. For example, WebShop \cite{yao2022webshop} and WebArena \cite{zhou2023webarena} focus on web operations, while SciWorld \cite{scienceworld2022} and TextCraft \cite{prasad2024adapt} are centered around scientific research.

\item \textbf{Medical Reasoning:} Medicine fundamentally involves complex reasoning, spanning tasks from diagnostic decision making to treatment planning. Benchmarks of JAMA Clinical Challenge \cite{chen2024benchmarking}, Medbullets \cite{chen2024benchmarking}, and MedQA \cite{jin2021disease} offer model measurements that mimic the doctor's disease diagnosis. 

\item \textbf{Multimodal Reasoning:} Multimodal reasoning, such as benchmarks of MMMU \cite{yue2024mmmu}
, MathVista \cite{lu2024mathvista}, MME-CoT \cite{jiang2025mme} and MM-IQ \cite{cai2025mm}, requires cross-modal thinking in combination with text and images. 
Especially for those visual-centered problems, in benchmarks MathVision \cite{MathVision}, MathVerse \cite{zhang2024mathverse}, CMMaTH \cite{li2024cmmath}
, PGPS9K \cite{Zhang2023PGPS}, SciFIBench \cite{roberts2024scifibench} and GRAB \cite{roberts2024grab}, put forward higher requirements for reasoning MMs.

\end{enumerate}

\subsubsection{Summary} 

The field of LLMs has advanced rapidly in recent years, with benchmark performance consistently improving. 
Simple reasoning benchmarks, such as GSM8K \cite{cobbe2021training}, MATH-500 \cite{lightmanlet}, and ScienceQA \cite{lu2022learn}, have approached performance saturation. 
Recent studies on reasoning LLMs \cite{guan2025rstarmathsmallllmsmaster,RedStar} show that models designed for long reasoning chains do not significantly outperform those designed for shorter chains on these benchmarks. 
This highlights the urgent need to establish new benchmarks that more effectively assess the reasoning capabilities of reasoning LLMs. 
Moreover, current benchmarks are limited, focusing mainly on solid reasoning tasks. 
Soft reasoning benchmarks, lacking explicitly defined correct answers, offer a more nuanced evaluation, better capturing the complexities and subtleties of human-like reasoning. Furthermore, it is essential to address the issue of data leakage in evaluation processes \cite{li2024open}. 
Ensuring the confidentiality and neutrality of evaluation data is critical to preserving the integrity and reliability of benchmark results.

\subsection{Evaluation Metrics}\label{metrics}

Depending on task types, technical proposals, and reasoning paradigms, various evaluation metrics have been introduced for reasoning LLMs as shown in Figure \ref{fig:evaluation_metrics}. These metrics are designed to more accurately assess the model's performance in handling complex reasoning tasks, ensuring that both the quality and coherence of the generated solutions are effectively measured. 

\begin{table*}[htbp]
\centering
\caption{Performance of Different Models, including Basic LLMs and Reasoning LLMs, on Plain Text Benchmarks. The \textcolor{deepred}{\textbf{red}} denotes the highest result, and the \textcolor{blue}{\textbf{blue}} denotes the second highest result.}
\label{tab:text_performance}
\resizebox{0.95\textwidth}{!}{
\begin{tabular}{l|lccccccc}
\toprule[1.2pt]
 \multicolumn{2}{c}{\multirow{3}{*}{\textbf{Model}}} & \multicolumn{2}{c}{\textbf{Math}} & \multicolumn{3}{c}{\textbf{Code}} & \multicolumn{2}{c}{\textbf{General}} \\
\cmidrule(lr){3-4} \cmidrule(lr){5-7} \cmidrule(lr){8-9}
 \multicolumn{2}{c}{}  & AIME 2024  & MATH-500  & LiveCodeBench & Codeforces  & SWE Verified & MMLU & GPQA-Diamond \\
 \multicolumn{2}{c}{} & (\textit{Pass@1}) & (\textit{Pass@1}) & (\textit{Pass@1-CoT}) & (\textit{Percentile}) & (\textit{Resolved}) & (\textit{Pass@1}) & (\textit{Pass@1}) \\
\hline
\fontsize{6.8}{9}\selectfont\multirow{6}{*}{ {\rotatebox{90}{\textbf{Basic LLMs}}}} 

&\cellcolor[rgb]{ .949,  .949,  .949}Gemini-2.5-Pro \cite{gemini2.5-pro} & \cellcolor[rgb]{ .949,  .949,  .949}\textcolor{deepred}{\textbf{92.0}} & \cellcolor[rgb]{ .949,  .949,  .949}- & \cellcolor[rgb]{ .949,  .949,  .949}70.4 & \cellcolor[rgb]{ .949,  .949,  .949}- & \cellcolor[rgb]{ .949,  .949,  .949}\textcolor{deepred}{\textbf{63.8}} & \cellcolor[rgb]{ .949,  .949,  .949}81.7 & \cellcolor[rgb]{ .949,  .949,  .949}\textcolor{deepred}{\textbf{84.0}} \\

& Gemini-2.0-Pro \cite{gemini2.0-pro} &- & 91.8 & 36.0 &- &- &86.5 & 64.7 \\

& \cellcolor[rgb]{ .949,  .949,  .949}GPT-4o \cite{gpt4o-0513} &\cellcolor[rgb]{ .949,  .949,  .949}9.3 & \cellcolor[rgb]{ .949,  .949,  .949}74.6 & \cellcolor[rgb]{ .949,  .949,  .949}34.2 & \cellcolor[rgb]{ .949,  .949,  .949}23.6 & \cellcolor[rgb]{ .949,  .949,  .949}38.8 & \cellcolor[rgb]{ .949,  .949,  .949}87.2 & \cellcolor[rgb]{ .949,  .949,  .949}49.9 \\

& Claude-3.5-Sonnet \cite{claude-3-5-sonnet} & 16.0 & 78.3 & 33.8 & 20.3 & 50.8 & 88.3 & 65.0 \\

&\cellcolor[rgb]{ .949,  .949,  .949}  Deepseek-V3 \cite{liu2024deepseek} &\cellcolor[rgb]{ .949,  .949,  .949} 39.2 &\cellcolor[rgb]{ .949,  .949,  .949} 90.2 &\cellcolor[rgb]{ .949,  .949,  .949} 36.2 &\cellcolor[rgb]{ .949,  .949,  .949} 58.7 &\cellcolor[rgb]{ .949,  .949,  .949} 42.0 &\cellcolor[rgb]{ .949,  .949,  .949} 88.5 &\cellcolor[rgb]{ .949,  .949,  .949} 59.1  \\

 &Claude 3.7 Sonnet \cite{claude-3-7-sonnet}  & - & 82.2 & - & - & \textcolor{blue}{\textbf{62.3}}& -& 68.0 \\


\hline

\fontsize{7}{10}\selectfont\multirow{21}{*}{{\rotatebox{90}{\textbf{Reasoning LLMs}}}
} & \cellcolor[rgb]{ .949,  .949,  .949}Eurus-2-7B-PRIME \cite{cui2025process} & \cellcolor[rgb]{ .949,  .949,  .949}26.7 & \cellcolor[rgb]{ .949,  .949,  .949}79.2 & \cellcolor[rgb]{ .949,  .949,  .949}- & \cellcolor[rgb]{ .949,  .949,  .949}- &\cellcolor[rgb]{ .949,  .949,  .949}- & \cellcolor[rgb]{ .949,  .949,  .949}- & \cellcolor[rgb]{ .949,  .949,  .949}-  \\

& InternLM3-8B-Instruct \cite{cai2024internlm2} & 20.0 & 83.0 & 17.8 & - & - & 76.6 & 37.4 \\

& \cellcolor[rgb]{ .949,  .949,  .949}rStar-Math-7B \cite{guan2025rstarmathsmallllmsmaster} & \cellcolor[rgb]{ .949,  .949,  .949}46.7 & \cellcolor[rgb]{ .949,  .949,  .949}81.6 &\cellcolor[rgb]{ .949,  .949,  .949}- & \cellcolor[rgb]{ .949,  .949,  .949}- & \cellcolor[rgb]{ .949,  .949,  .949}- & \cellcolor[rgb]{ .949,  .949,  .949}82.7 & \cellcolor[rgb]{ .949,  .949,  .949}54.9 \\

  & STILL-2-32B \cite{Slow_Thinking_with_LLMs_2}& 46.7 & 90.2 & - & - &- & - & - \\
  
    & \cellcolor[rgb]{ .949,  .949,  .949}Redstar-code-math \cite{RedStar} & \cellcolor[rgb]{ .949,  .949,  .949}53.3 & \cellcolor[rgb]{ .949,  .949,  .949}91.2 &\cellcolor[rgb]{ .949,  .949,  .949}- &\cellcolor[rgb]{ .949,  .949,  .949}- &\cellcolor[rgb]{ .949,  .949,  .949}- &\cellcolor[rgb]{ .949,  .949,  .949}- &\cellcolor[rgb]{ .949,  .949,  .949}-  \\
    
 & Search-o1 \cite{li2025search}& 56.7 & 86.4 & 33.0 & - &- & - &63.6 \\
 
 & \cellcolor[rgb]{ .949,  .949,  .949}QwQ \cite{qwq-32b-preview} & \cellcolor[rgb]{ .949,  .949,  .949}50.0 & \cellcolor[rgb]{ .949,  .949,  .949}90.6 & \cellcolor[rgb]{ .949,  .949,  .949}41.9 & \cellcolor[rgb]{ .949,  .949,  .949}62.0 & \cellcolor[rgb]{ .949,  .949,  .949}- &\cellcolor[rgb]{ .949,  .949,  .949}- & \cellcolor[rgb]{ .949,  .949,  .949}54.5 \\
 
  & s1-32B \cite{muennighoff2025s1} & 56.7 & 93.0 & - &- & - &- & 59.6 \\
  
 &\cellcolor[rgb]{ .949,  .949,  .949}OpenAI o1-mini \cite{o1-mini} & \cellcolor[rgb]{ .949,  .949,  .949}63.6 & \cellcolor[rgb]{ .949,  .949,  .949}90.0 &\cellcolor[rgb]{ .949,  .949,  .949}53.8 & \cellcolor[rgb]{ .949,  .949,  .949}93.4 & \cellcolor[rgb]{ .949,  .949,  .949}41.6 & \cellcolor[rgb]{ .949,  .949,  .949}85.2 & \cellcolor[rgb]{ .949,  .949,  .949}60.0 \\
 
  & LIMO-32B \cite{ye2025limoreasoning} & 57.1 & 94.8 & - &- &- &- & 66.7 \\
  
& \cellcolor[rgb]{ .949,  .949,  .949}Kimi k1.5 long-CoT \cite{team2025kimi} & \cellcolor[rgb]{ .949,  .949,  .949}77.5 & \cellcolor[rgb]{ .949,  .949,  .949}96.2 & \cellcolor[rgb]{ .949,  .949,  .949}62.5 &\cellcolor[rgb]{ .949,  .949,  .949}94.0 & \cellcolor[rgb]{ .949,  .949,  .949}- &\cellcolor[rgb]{ .949,  .949,  .949}- & \cellcolor[rgb]{ .949,  .949,  .949}- \\

&OpenAI o3-mini \cite{o3-mini}  & \textcolor{blue}{\textbf{87.3}} & \textcolor{deepred}{\textbf{97.9}} & \textcolor{deepred}{\textbf{84.6}} & - & 49.3 & 86.9 & \textcolor{blue}{\textbf{79.7}} \\

 &\cellcolor[rgb]{ .949,  .949,  .949}Seed-Thinking-v1.5 \cite{seed2025seedthinkingv15advancingsuperbreasoning}  & \cellcolor[rgb]{ .949,  .949,  .949}86.7 &\cellcolor[rgb]{ .949,  .949,  .949} - &\cellcolor[rgb]{ .949,  .949,  .949} 64.9 &\cellcolor[rgb]{ .949,  .949,  .949} 55.0 &\cellcolor[rgb]{ .949,  .949,  .949} 47.0 &\cellcolor[rgb]{ .949,  .949,  .949} 87.0 &\cellcolor[rgb]{ .949,  .949,  .949} 77.3 \\
 
&DeepSeek-R1-Distill-Qwen-1.5B \cite{deepseekai2025deepseekr1incentivizingreasoningcapability}  & 28.9 & 83.9 & 16.9 & - & - & - & 33.8 \\

 &\cellcolor[rgb]{ .949,  .949,  .949}DeepSeek-R1-Distill-Qwen-7B\cite{deepseekai2025deepseekr1incentivizingreasoningcapability}  & \cellcolor[rgb]{ .949,  .949,  .949}55.5 &\cellcolor[rgb]{ .949,  .949,  .949} 92.8&\cellcolor[rgb]{ .949,  .949,  .949} 37.6 &\cellcolor[rgb]{ .949,  .949,  .949} - &\cellcolor[rgb]{ .949,  .949,  .949} - &\cellcolor[rgb]{ .949,  .949,  .949}-&\cellcolor[rgb]{ .949,  .949,  .949}49.1 \\

&DeepSeek-R1-Distill-Qwen-14B \cite{deepseekai2025deepseekr1incentivizingreasoningcapability}  & 69.7 & 93.9 & 53.1 & - & - & - & 59.1 \\

 &\cellcolor[rgb]{ .949,  .949,  .949}DeepSeek-R1-Distill-Qwen-32B\cite{deepseekai2025deepseekr1incentivizingreasoningcapability}  & \cellcolor[rgb]{ .949,  .949,  .949}72.6 &\cellcolor[rgb]{ .949,  .949,  .949} 94.3&\cellcolor[rgb]{ .949,  .949,  .949} 57.2&\cellcolor[rgb]{ .949,  .949,  .949} - &\cellcolor[rgb]{ .949,  .949,  .949} - &\cellcolor[rgb]{ .949,  .949,  .949}-&\cellcolor[rgb]{ .949,  .949,  .949}62.1\\

 &DeepSeek-R1-Distill-Llama-8B \cite{deepseekai2025deepseekr1incentivizingreasoningcapability}  & 50.4 & 89.1 & 39.6 & - & - & - & 49.0\\

&\cellcolor[rgb]{ .949,  .949,  .949}DeepSeek-R1-Distill-Llama-70B \cite{deepseekai2025deepseekr1incentivizingreasoningcapability}  &\cellcolor[rgb]{ .949,  .949,  .949} 70.0 &\cellcolor[rgb]{ .949,  .949,  .949} 94.5 &\cellcolor[rgb]{ .949,  .949,  .949} 57.5 & \cellcolor[rgb]{ .949,  .949,  .949}- & \cellcolor[rgb]{ .949,  .949,  .949}- & \cellcolor[rgb]{ .949,  .949,  .949}- & \cellcolor[rgb]{ .949,  .949,  .949}65.2 \\

& DeepSeek-R1 \cite{Deepseek-R1} & 79.8 & \textcolor{blue}{\textbf{97.3}} & \textcolor{blue}{\textbf{65.9}} & \textcolor{blue}{\textbf{96.3}} & 49.2 & \textcolor{blue}{\textbf{90.8}} & 71.5 \\
   
&\cellcolor[rgb]{ .949,  .949,  .949}OpenAI-o1 \cite{openai_o1} & \cellcolor[rgb]{ .949,  .949,  .949}79.2 & \cellcolor[rgb]{ .949,  .949,  .949}96.4 & \cellcolor[rgb]{ .949,  .949,  .949}63.4 & \cellcolor[rgb]{ .949,  .949,  .949}\textcolor{deepred}{\textbf{96.6}} & \cellcolor[rgb]{ .949,  .949,  .949}48.9 & \cellcolor[rgb]{ .949,  .949,  .949}\textcolor{deepred}{\textbf{91.8}} & \cellcolor[rgb]{ .949,  .949,  .949}75.7 \\

\bottomrule[1.2pt]
\end{tabular}
}
\end{table*}

\subsubsection{Task Types}

In terms of benchmark categories, mathematical reasoning typically uses two main metrics: \textit{Pass@k} and \textit{Cons@k}. The \textit{Pass@k} metric evaluates the model's ability to generate a correct solution within k attempts, measuring the likelihood of success within a limited number of tries. On the other hand, \textit{Cons@k} assesses whether the model consistently produces correct or logically coherent solutions, highlighting the stability and reliability of its reasoning capabilities. For code tasks, the key metrics are \textit{Elo} and \textit{Percentile}, both of which measure the relative skill in generating correct code compared to other models or human programmers. In scientific tasks, evaluation generally employs \textit{Exact Match (EM)} and \textit{Accuracy} for fill-in-the-blank and multiple-choice questions, respectively. 
The \textit{EM} metric judges whether the model's output exactly matches the expected solution, while \textit{Accuracy} measures the proportion of correct answers out of the total number of questions.

\subsubsection{Technical Proposals}

Based on technical routes, the schemes with ORM or PRM often leverage \textit{RM@k} and \textit{Best-of-N} two evaluation indicators. 
\textit{RM@k} measures whether the reward model can rank the good answer higher in the top k candidates according to reward score, and \textit{Best-of-N} chooses the solution with highest score from N generated reasoning trajectories. Methods for self-consistency are evaluated using \textit{Greedy Decoding}, \textit{Beam Search}, and \textit{Major@k}. \textit{Greedy Decoding} and \textit{Beam Search} control the randomness of the inference process by limiting the sampling range. \textit{Major@k} selects the solution with the most consistent results from k candidate solutions. In RL, metrics reflect both performance in achieving desired outcomes and the efficiency of the learning process. For example, \textit{Cumulative Reward} measures the total reward received by the agent over time, while \textit{Sample Efficiency} assesses the efficiency of the agent's sample usage during learning.

\subsubsection{Reasoning Paradigms}

For reasoning paradigm of the multi-turn solution generation in reasoning LLMs, \textit{Outcome Efficiency} and \textit{Process Efficiency} \cite{Tecent_2_plus_3} are proposed recently to evaluate the efficiency of long thinking specifically. \textit{Outcome Efficiency} metric empirically evaluates how effectively later solutions contribute to accuracy improvements, formulating as the ratio of efficient tokens that contribute to reaching the correct answer, to all output tokens. \textit{Process Efficiency} metric evaluates the contribution of later solutions to solution diversity empirically, concretely representing as the ratio of tokens of distinct solutions to all solution tokens. These two indicators reveal to the overthinking issue of existing reasoning LLMs to simple problems certainly.

\subsubsection{Summary} 

Most of the existing evaluation metrics are judged according to the final answer. 
It is imperative to develop a comprehensive assessment framework that considers various aspects of the reasoning process in view of the large inference computation consumption. 
Current popular evaluation frameworks, such as LMMs-Eval \cite{zhang2024lmmsevalrealitycheckevaluation}, OpenCompass \cite{2023opencompass}, and PRMBench \cite{song2025prmbench}, lack efficiency and their metrics do not adequately account for the computational and temporal efficiency of the reasoning process. 
To address these shortcomings, we highly recommend exploring more efficient proxy tasks as potential solutions. 
By identifying and utilizing tasks that better capture the nuances of long reasoning chains, we can develop more robust and effective evaluation metrics to enhance the overall assessment framework, ensuring that it not only measures the accuracy of the final output but also evaluates the efficiency and coherence of the reasoning process throughout.

\subsection{Performance Comparison}\label{performance_compare}

In this section, we compare the performance of different reasoning LLMs and their corresponding foundational LLMs on plain text benchmarks, such as math and code problems, as well as on multimodal benchmarks. 
The comprehensive real-time leaderboard is available on \href{https://lmarena.ai/?leaderboard}{Arena}.

\subsubsection{Performance on Plain Text Benchmarks}

As shown in Table \ref{tab:text_performance}, reasoning LLMs, such as DeepSeek-R1 \cite{Deepseek-R1} and OpenAI-o1/o3 \cite{openai_o1, o3-mini}, demonstrate exceptional performance across a wide range of tasks, including math, coding, and other general tasks. 
These models achieve high scores on multiple plain-text benchmarks, such as AIME 2024, MATH-500, and LiveCodeBench, showcasing their robust text-based reasoning abilities. 
In contrast, foundational LLMs, like GPT-4o \cite{openai2023gpt4}, Claude-3.5-Sonnet \cite{claude-3-5-sonnet}, and DeepSeek-V3 \cite{liu2024deepseek}, generally perform less effectively than reasoning LLMs, particularly in math and coding tasks (\emph{e.g.}, AIME 2024 and Codeforces). For example, OpenAI-o1 outperforms GPT-4o by 69.9\% and 73\% on these tasks, respectively. 
Moreover, DeepSeek-R1, based on the DeepSeek-V3 architecture, surpasses its predecessor on all benchmarks, further highlighting the advantages of the reasoning LLMs.

\subsubsection{Performance on Multimodal Benchmarks}

As shown in Table \ref{tab:vlm_performance}, reasoning LLMs continue to excel in multimodal tasks. OpenAI-o1 \cite{openai_o1} performs strongly in vision tasks, achieving the highest score of 78.2\% on MMMU and outperforming its corresponding foundational LLM, GPT-4o \cite{openai2023gpt4}, by 7.2\% on MathVista. However, the performance improvement in multimodal tasks is less pronounced compared to text-only tasks. This can be attributed in part to the limitations of current multimodal reasoning LLM techniques, as well as the lack of sufficient datasets to fully assess the multimodal capabilities of reasoning LLMs.

\begin{table}[tbp]
\centering
\caption{Performance of Models, including Basic LLMs and Reasoning LLMs, on Multimodal Benchmarks. The \textcolor{deepred}{\textbf{red}} denotes the highest result, and the \textcolor{blue}{\textbf{blue}} denotes the second highest result.}
\resizebox{0.98\linewidth}{!}{
\begin{tabular}{c|lcccc}
\toprule[1.2pt]
 \multicolumn{2}{c}{{\textbf{Model}}} & \textbf{MMMU} & \textbf{Mathvista} & \textbf{Mathvision} & \textbf{Olympiadbench} \\
\midrule
\fontsize{6}{6}\selectfont\multirow{6}{*}{\rotatebox{90}{{\textbf{Basic LLMs}}}}&\cellcolor[rgb]{ .949,  .949,  .949}Claude-3.5-Sonnet \cite{claude-3-5-sonnet} &\cellcolor[rgb]{ .949,  .949,  .949}68.3 &\cellcolor[rgb]{ .949,  .949,  .949} 65.3 &\cellcolor[rgb]{ .949,  .949,  .949} 38.0 &\cellcolor[rgb]{ .949,  .949,  .949} - \\

&GPT-4o \cite{gpt4o-0513} & 69.1 & 63.8 &30.4 & \textcolor{blue}{\textbf{25.9}} \\

&\cellcolor[rgb]{ .949,  .949,  .949}Gemini 2.0 Pro \cite{gemini2.0-pro} & \cellcolor[rgb]{ .949,  .949,  .949}72.7 & \cellcolor[rgb]{ .949,  .949,  .949}- &\cellcolor[rgb]{ .949,  .949,  .949}- & \cellcolor[rgb]{ .949,  .949,  .949}- \\

&Llama 4 Maverick \cite{noauthor_llama_nodate} & 73.4 & 73.7 & - & - \\

&\cellcolor[rgb]{ .949,  .949,  .949}Claude-3.7-Sonnet \cite{claude-3-7-sonnet} &\cellcolor[rgb]{ .949,  .949,  .949} 75.0 &\cellcolor[rgb]{ .949,  .949,  .949} - &\cellcolor[rgb]{ .949,  .949,  .949} - &\cellcolor[rgb]{ .949,  .949,  .949} - \\

&Gemini 2.5 Pro\cite{gemini2.5-pro}&\textcolor{deepred}{\textbf{81.7}} &- &- &-  \\

\hline

\fontsize{6}{6}\selectfont\multirow{12}{*}{\rotatebox{90}{{\textbf{Reasoning LLMs}}}}

&\cellcolor[rgb]{ .949,  .949,  .949}Skywork-R1V-38B \cite{skywork2025r1v} &\cellcolor[rgb]{ .949,  .949,  .949} 69.0&\cellcolor[rgb]{ .949,  .949,  .949} 67.5&\cellcolor[rgb]{ .949,  .949,  .949} -&\cellcolor[rgb]{ .949,  .949,  .949} -\\

& LLaVA-CoT \cite{xu2024llava} & - & 54.8 & - & - \\

&\cellcolor[rgb]{ .949,  .949,  .949}Kimi k1.5 long-CoT \cite{team2025kimi} &\cellcolor[rgb]{ .949,  .949,  .949} 70.0 &\cellcolor[rgb]{ .949,  .949,  .949} 74.9 &\cellcolor[rgb]{ .949,  .949,  .949} - &\cellcolor[rgb]{ .949,  .949,  .949} - \\

& Qwen2.5-VL-72B \cite{bai2025qwen2} & 70.2 & 74.8 & \textcolor{blue}{\textbf{38.1}} & - \\

&\cellcolor[rgb]{ .949,  .949,  .949}QvQ-72B-preview \cite{qvq-72b-preview} & \cellcolor[rgb]{ .949,  .949,  .949}70.3 & \cellcolor[rgb]{ .949,  .949,  .949}71.4 & \cellcolor[rgb]{ .949,  .949,  .949}35.9 & \cellcolor[rgb]{ .949,  .949,  .949}20.4 \\

&Kimi k1.6 preview &- &\textcolor{blue}{\textbf{80.0}} &\textcolor{deepred}{\textbf{53.3}}  &-  \\

&\cellcolor[rgb]{ .949,  .949,  .949}MMR1-Math-v0-7B \cite{MMR1-Math2025} &\cellcolor[rgb]{ .949,  .949,  .949}- &\cellcolor[rgb]{ .949,  .949,  .949}71.0 &\cellcolor[rgb]{ .949,  .949,  .949}30.2  &\cellcolor[rgb]{ .949,  .949,  .949}-  \\

&MM-EUREKA \cite{meng2025mm} & -& 64.2& 26.6& \textcolor{deepred}{\textbf{37.3}}\\

&\cellcolor[rgb]{ .949,  .949,  .949}R1-Onevision-7B \cite{yang_r1-onevision_2025} &\cellcolor[rgb]{ .949,  .949,  .949}- &\cellcolor[rgb]{ .949,  .949,  .949}64.1 &\cellcolor[rgb]{ .949,  .949,  .949}29.9  &\cellcolor[rgb]{ .949,  .949,  .949}-  \\

& Doubao-1.5-pro \cite{Doubao-1.5-pro} & 	73.8& 78.8 & 48.6 &48.5 \\

&\cellcolor[rgb]{ .949,  .949,  .949}OpenAI-o1 \cite{openai_o1} & \cellcolor[rgb]{ .949,  .949,  .949}78.2 & \cellcolor[rgb]{ .949,  .949,  .949}71.0 &\cellcolor[rgb]{ .949,  .949,  .949}- &\cellcolor[rgb]{ .949,  .949,  .949}- \\

&OpenAI o4-mini \cite{o4-mini} &\textcolor{blue}{\textbf{81.6}} &\textcolor{deepred}{\textbf{84.3}} &-  &-  \\

\bottomrule[1.2pt]
\end{tabular}
}
\label{tab:vlm_performance}
\end{table}

\subsubsection{Summary}

In summary, reasoning LLMs show strong performance across both plain text and multimodal benchmarks, particularly excelling in math and coding tasks, where they outperform foundational LLMs by a large margin. 
Although the improvement in multimodal tasks is not as pronounced as in text-only tasks, reasoning LLMs still surpass their counterparts, highlighting their potential for processing both image and text data. 
These results emphasize the versatility and effectiveness of reasoning LLMs across a broad spectrum of reasoning tasks, with potential for further advancements in multimodal reasoning techniques.

\section{Extended Techniques}
%
\label{extend_techniques}
\subsection{Advanced Infrastructure}
\label{advanced_infra}
In the realm of System-2 paradigms, reinforcement learning stands as a pivotal methodology to elevate the performance of reasoning models, finding ubiquitous employment in LLM, MLLM, and agentic applications such as Agent-R1 \cite{Agent-R1} and UI-R1 \cite{ui-r1}. Recent advancements have seen state-of-the-art RL frameworks like VeRL \cite{VeRL}, OpenRLHF \cite{OpenRLHF}, and ReaLHF \cite{ReaLHF} emerge as dominant tools in this domain. Despite their divergent architectural designs, these frameworks adhere to a canonical RLHF workflow comprising three core stages:

\begin{itemize}
 
\item {\textbf{Generative Rollout Phase}: The actor network generates outputs in response to a batch of input prompts. In agentic contexts, this process transcends mere static response generation, necessitating dynamic interactions with complex environments to facilitate adaptive decision-making.}  

\item {\textbf{Preprocessing Pipeline}: Post-actor inference, the framework undertakes a sophisticated derivation of auxiliary information critical for RL training. This includes computing log probabilities from the reference policy to quantify behavioral consistency, estimating value functions via the critic model to assess state desirability, and calculating advantage estimates to prioritize high-impact learning signals.}

\item {\textbf{Optimization Stage}: With the curated dataset from the generative and preprocessing stages—encompassing actor outputs, critic evaluations, and policy gradients—the system initiates iterative parameter updates. This stage harmonizes actor-critic dynamics to refine decision-making strategies, ensuring alignment with task objectives and environmental feedback.}
\end{itemize}

{This structured workflow underscores the synergy between generative modeling, value estimation, and policy optimization, cementing RLHF as a cornerstone for developing robust, adaptive intelligent systems.}

{In the Generative Rollout Phase, contemporary inference engines such as vLLM \cite{vLLM} and SGLang \cite{SGLang} are leveraged to facilitate high-throughput deployment, ensuring efficient realization of generative workflows.}

{For the Optimization Stage, diverse parallelism paradigms—including data parallelism, pipeline parallelism, and tensor parallelism—are systematically implemented. This is achieved through state-of-the-art distributed training frameworks like Megatron-LM \cite{Megatron-LM} and MegaScale \cite{MegaScale}, which incorporate sophisticated 3D parallelism. These architectural designs enable seamless orchestration of computational resources, enhancing both scalability and training efficiency across distributed systems.}

\subsubsection{Taxonomy of RL Infrastructure}

{While distinct RLHF frameworks broadly adhere to an analogous workflow, notable discrepancies emerge in their architectural configurations. The most pronounced differentiator resides in whether they employ a \textbf{Single-Controller} or \textbf{Multi-Controller} paradigm, a distinction that fundamentally shapes their operational dynamics and scalability profiles.}

{Within the \textbf{Single-Controller} paradigm, a centralized command hub governs the holistic execution trajectory of distributed programs, enabling users to craft core dataflow functionalities as an integrated workflow. However, the propagation of coordination messages from this central hub to all worker nodes engenders significant dispatch overhead during the execution of large-scale dataflow graphs across sprawling clusters. VeRL and ReaLHF exemplify this architectural model with notable prominence, illustrating both its cohesive design advantages and scalability challenges in high-scale distributed environments.}

{In the domain of \textbf{Multi-Controller} architectures, each device—hereafter referred to as a worker—is equipped with its own bespoke controller. Cutting-edge distributed systems for LLM training and inference demonstrate a pronounced preference for the multi-controller paradigm. This inclination is primarily attributable to its inherent scalability and the minimal scheduling overhead it incurs.} 

{In the context of implementing the RLHF workflow within a multi-controller architectural framework, practitioners are required to meticulously choreograph the integration of codebase governing inter-device collective communication protocols, computational logic frameworks, and point-to-point data transfer mechanisms across each device's operational runtime environment. This imperative orchestration gives rise to profoundly nested code architectures, wherein computational routines and data transmission operations become inextricably interwoven. The resultant architectural intricacy poses formidable challenges for development, maintenance, and performance optimization—given that the inter-dependencies between functional components necessitate elaborate debugging methodologies and precision tuning to ensure seamless cross-component synchronization and optimal resource allocation. OpenRLHF stands as a paradigmatic exemplar of such systems, embodying the complex interplay between modular device autonomy and centralized coordination required to sustain efficient RLHF pipeline execution.}

\subsubsection{Prospective Challenges of RL Infrastructure}

\begin{enumerate}[itemindent=0em]
\item {\textbf{Long-Text-Generation}. The prioritization of long-text generation in reasoning models presents substantial challenges for the generation stage. During Long-CoT generation, the substantial variability in response lengths across different prompts leads to significant GPU under-utilization due to prolonged idle periods. In the context of agentic reinforcement learning (Agentic-RL), the generation stage must not only efficiently produce lengthy responses but also interact dynamically with environments. As environments scale in complexity and diversity, this interaction imposes escalating challenges on generative efficiency and adaptability.  For long-text generation, numerous studies have introduced sophisticated rollout strategies to address these issues. For example, Kimi-1.5 \cite{team2025kimi} employs a partial-rollout framework, eschewing the need for monolithic response generation. Instead, it processes and stores textual segments incrementally, enabling the creation of significantly longer outputs while maintaining rapid iteration cycles. During training, selective exclusion of non-critical segments from loss computation optimizes the learning process, enhancing both operational efficiency and system scalability. Concurrently, Seed-Thinking-1.5 \cite{seed-thinking} introduces a Streaming Rollout System that decouples model evolution from runtime execution, facilitating dynamic adjustment of on/off-policy sampling. This architectural innovation ensures that generative processes remain responsive to environmental feedback without compromising computational efficiency, thus advancing the state-of-the-art in scalable long-text generation frameworks.}

\item {\textbf{Scalable Design for Agent-Environment Interaction}. In the realm of agent-environment interaction scenarios \cite{agent-arena}, the dynamic interplay between agents and their surroundings poses a formidable challenge, primarily stemming from the heterogeneous latencies inherent across diverse environments \cite{gui-r1, ui-r1}. To address this, we posit that leveraging an asynchronous message-queue architecture such as the producer-consumer paradigm becomes imperative for effectively managing the temporal discrepancies in agent-environment exchanges. This approach mitigates the adverse effects of variable delays, ensuring robust coordination even in systems characterized by unpredictable response times. Regarding environmental stability, our analysis, rooted in scaling-law principles, underscores the critical role of scalable environmental design in advancing LLM/MLLM-based agentic applications (e.g., DeepResearch \cite{Agentic-Reason}, DeepSearch \cite{Open-Deep-Search}, and autonomous tooling frameworks like OpenManus \cite{OpenManus}). Unlike traditional LLM/MLLM training pipelines, where data resides statically within file systems, agentic environments typically consist of distributed assemblages of HTTP servers. This architectural distinction introduces a nontrivial stability challenge for large-scale Agentic-RL training, as the reliability of such systems hinges on the seamless integration and resilience of networked components. Scaling environments to accommodate growing computational demands while maintaining operational consistency thus emerges as a foundational requirement for realizing the full potential of next-generation agentic systems.}

\item {\textbf{Protocol Design and Security Considerations}. Developing interfaces between disparate HTTP servers and LLMs can impose substantial overhead, as each integration often requires bespoke engineering efforts. A standardized protocol—such as the Model-Computer Protocol (MCP) \cite{MCP}—facilitates seamless interaction between LLMs/MLMs and tools without requiring intimate knowledge of tool-specific implementations. This framework streamlines development by mitigating the need for redundant interface design, thereby enhancing interoperability and reducing engineering latency. However, this architectural elegance introduces a critical security imperative: third-party MCP servers may harbor malicious code that could compromise the integrity of the training ecosystem. Such risks necessitate robust safeguards to prevent unauthorized access, code injection, or systemic sabotage, ensuring that the benefits of protocol standardization are not overshadowed by potential vulnerabilities. Proactive measures—including rigorous code audits \cite{CodeAuditing}, secure authentication mechanisms \cite{authentication_mechnism}, and runtime monitoring \cite{Runtime-Monitoring} are essential to maintain the reliability and safety of the integrated system, striking a balance between operational efficiency and defensive resilience.}

\end{enumerate}

{\subsection{Trustworthiness of Reasoning LLMs} \label{srs_safety_of_lrm}}

{As RLLMs grow more capable, their trustworthiness becomes increasingly critical—particularly in domains requiring transparency, factual grounding, and safety. Unlike general LLMs, RLLMs introduce unique trust challenges due to their explicit multi-step reasoning, reliance on external tools, and complex decision traces. Recent work has begun addressing these issues by enhancing interpretability, managing knowledge conflicts, improving robustness, grounding reasoning in evidence, and enabling controllability.}

\begin{enumerate}[itemindent=0em]

\item {\textbf{Interpretability of Reasoning Processes:} 
RLLMs often produce complex reasoning traces, making it hard to assess whether outputs reflect genuine internal decision-making. Faithfulness metrics have been introduced to evaluate the alignment between explanations and model behavior \cite{wei-jie-etal-2024-interpretable}. Mechanistic analyses explore how reasoning steps are encoded in neurons or attention heads\cite{dutta2024mechanistic, li2024happened}. Other studies examine how gradient dynamics differ when training models for slow, deliberate reasoning \cite{li2024gradient}. These efforts enhance transparency while preserving performance.}

\item {\textbf{Knowledge Conflicts} Knowledge inconsistencies arise when internal model memory conflicts with external context, undermining reasoning. Recent surveys categorize such conflicts and highlight their impact on reasoning accuracy \cite{xu-etal-2024-knowledge-conflicts, bi2024factuality}. Methods have been proposed to detect and resolve these conflicts using special prompt \cite{wang2024resolving}, decoding \cite{bi2024decoding, chuang2023dola} and alignment \cite{lin2024flame, bi2024context} technologies, while advanced editing approaches \cite{li2025reinforced, bi2024decoding, bi2024adaptive} allow updated facts to be incorporated into reasoning traces without full model retraining.}

\item {\textbf{Safety \& Robustness:} Reasoning enables models to reflect on safety but also opens new vulnerabilities. Alignment procedures often reduce reasoning performance, a trade-off termed "safety tax" \cite{huang2025safetytax}. Chain-of-thought attacks, such as prompt hijacking and input slowdown, expose structural vulnerabilities in reasoning flows \cite{guan2024deliberative, kumar2025overthink}. Backdoor-based jailbreaks further demonstrate that specific reasoning patterns can trigger harmful behavior \cite{kuo2025hcot, zhu2025bot}. To counter this, researchers propose deliberative safety alignment \cite{liu2025guardreasoner}, reasoning-based safeguard modules \cite{wen2025thinkguard}, and dedicated safe-CoT benchmarks \cite{jiang2025safechain}.}

\item {\textbf{External Grounding and Truthfulness:} RLLMs benefit from grounding reasoning in external evidence to mitigate hallucination. Retrieval-augmented frameworks dynamically adapt context during reasoning \cite{li2024chainofknowledge, bi2024struedit}. Other prompting techniques first extract chains of evidence from the input before generating an answer \cite{tran2024rare}. Reinforcement methods reward reasoning paths that include faithful, verifiable external content \cite{parvez2024chain}. Symbolic-verification modules also check reasoning consistency against retrieved data \cite{sun2025rearter}.}

\item {\textbf{Consistency and Controllability:} To improve reliability, RLLMs must generate consistent reasoning across runs and support external control. Dual-model self-reflection and critique techniques help detect and revise flawed reasoning \cite{li2025twoheads}. Prompting strategies that separate high-level plans from step-wise solutions improve coherence \cite{wang2024scot, ge2025innate, yang2025quantifying}. Studies also show that excessively long reasoning chains reduce accuracy, prompting step-length control mechanisms \cite{wu2025when}.}
\end{enumerate}

{Across interpretability, safety, grounding, and control, recent research demonstrates growing attention to the trustworthiness of RLLMs. Still, many challenges remain, particularly in balancing reasoning ability with safety, ensuring consistency under open-domain inputs, and aligning long reasoning chains with verified knowledge. Future efforts must continue integrating reasoning supervision, verification, and robust alignment into unified RLLM frameworks.}

{\subsection{Reasoning LLM as Agent} \label{agent_reason}}

{LLMs are increasingly employed as autonomous agents capable of complex reasoning and interaction with external environments. This involves enabling them to actively seek information, utilize diverse tools, and dynamically refine their reasoning. Recent advancements span enhanced search augmentation, broader tool integration, novel training methodologies, and evolving evaluation frameworks.}

{\subsubsection{Search-Augmented Reasoning}}
{Rather than relying solely on internal knowledge, these models are being trained to interact with external search engines to access up-to-date information. Reinforcement learning emerged as a prominent technique for training LLMs to autonomously generate relevant search queries as an integral part of their reasoning chains \cite{jin2025searchr1trainingllmsreason},\cite{chen2025researchlearningreasonsearch}, \cite{li2025searcho1agenticsearchenhancedlarge}. Frameworks were developed to incentivize models to invoke search proactively during reasoning, optimizing the interaction based on outcomes rather than predefined processes or distillation \cite{song2025r1searcherincentivizingsearchcapability}. Furthermore, research explored the synergistic potential of combining open-source reasoning LLMs with specialized search tools, demonstrating that such combinations can achieve performance competitive with larger, proprietary systems \cite{alzubi2025opendeepsearchdemocratizing}. The complexity of information retrieval was also addressed through collaborative approaches, where multiple LLM agents work together, potentially using sophisticated search strategies like MCTS, to tackle complex search-based reasoning tasks \cite{yang2025multillmcollaborativesearchcomplex}.}

{\subsubsection{Tool Integration Beyond Search}}
{Beyond web search, the scope of tool integration for LLM agents expanded considerably. Efforts focused on enabling LLMs to leverage a wider array of external tools, such as code interpreters and specialized solvers. Self-learning frameworks were introduced, allowing models to acquire tool-using skills through methods like hint-based learning and rejection sampling fine-tuning, effectively teaching them when and how to employ external functionalities \cite{li2025startselftaughtreasonertools}. This capability proved valuable in specialized domains; for instance, systems were developed to automate the complex process of modeling and solving Operations Research (OR) problems by translating natural language descriptions into formal mathematical models and generating executable code for solvers \cite{zhang2025orllmagentautomatingmodelingsolving}. More comprehensive agentic reasoning frameworks emerged, integrating multiple capabilities—including web search \cite{fourney2024magenticonegeneralistmultiagentsolving, sun2024genesis}, code execution \cite{zheng2025opencodeinterpreterintegratingcodegeneration}, and structured reasoning-context memory \cite{packer2024memgptllmsoperatingsystems, wu2025agenticreasoningreasoningllms,lei2025spider20evaluatinglanguage} (e.g., using a mind map structure)—to support more robust and multifaceted problem-solving.}

{\subsubsection{Training Methodologies}}
{Innovations in training methodologies were crucial for developing these sophisticated agentic capabilities. Research demonstrated the effectiveness of reinforcement learning techniques for enhancing reasoning abilities, even in the context of smaller LLMs, suggesting that advanced reasoning is achievable without massive parameter counts \cite{dang2025reinforcementlearningreasoningsmall}. Further emphasizing self-improvement through reinforcement learning, some approaches employ a two-stage paradigm involving initial format tuning to instill structured reasoning patterns incorporating meta-actions, followed by large-scale RL optimization enabling autoregressive search and self-correction with minimal initial supervision \cite{shen2025satorireinforcementlearningchainofactionthought}. RL-based training was successfully scaled to complex, specialized domains like software engineering, utilizing lightweight, rule-based reward signals to guide the learning process effectively \cite{wei2025swerladvancingllmreasoning}. Advanced RL strategies, such as explicit policy optimization over multiple interaction turns, were explored to improve strategic reasoning performance without requiring initial supervised fine-tuning phases \cite{liu2025epoexplicitpolicyoptimization}. Additionally, novel training paradigms were developed to cultivate complex reasoning patterns, such as long chains of thought. Bootstrapping methods, for example, enabled the development of these capabilities without relying heavily on advanced teacher models or extensive human annotations, offering a more scalable approach to training deep reasoning \cite{pang2025boltbootstraplongchainofthought}.}

{\subsubsection{Evaluation Benchmarks and Metrics}}
{As LLM agent capabilities advanced, evaluation methodologies evolved in tandem. New benchmarks were introduced specifically to assess the complex, multi-step reasoning and tool-use abilities required in real-world applications \cite{zhang2025datascibenchllmagentbenchmark}. Interactive platforms and "arenas" gained prominence, providing environments for systematically comparing the performance of different LLM agents on user-defined tasks, often incorporating community voting and feedback mechanisms for a more holistic evaluation \cite{chai2025a3androidagentarena}, \cite{agent-arena}, \cite{bonatti2024windowsagentarenaevaluating}, \cite{zhou2024webarenarealisticwebenvironment}. Alongside these new developments, established benchmarks measuring core reasoning abilities (e.g., AIME \cite{AIME2024}, GPQA \cite{rein2024gpqa}) and code generation/execution (e.g., LiveCodeBench \cite{jain2024livecodebench}, SWE-Bench \cite{jimenez2024swebench}) continued to serve as important yardsticks for tracking progress in foundational capabilities. }

{
\subsection{Efficient Reasoning LLM}
\label{efficient_lrm}
Despite LongCoT demonstrating strong generalization and reasoning abilities, the autoregressive paradigm of language models imposes a significant reasoning burden, limiting the application of these models in agent-based or edge scenarios \cite{OS-ATLAS}. Additionally, numerous studies have shown that language models often exhibit excessive reasoning, with a lot of redundant inference in the thought chains. As a result, various technical approaches have emerged to improve the reasoning efficiency of Reason LLMs. We categorize these approaches into the following categories based on their strategies:}

\begin{enumerate}[itemindent=0em]

\item {\textbf{Building reasoning-budget-sensitive LLMs:} CoD \cite{CoD} and TALE-EP \cite{TALE-EP} attempt to impose reasoning budget constraints on Reason LLMs to control the overall reasoning cost. Models like L1 \cite{L1}, TOPS \cite{TOPS}, O1-Pruner \cite{o1_pruner}, and Kimi-k1.5 \cite{team2025kimi} add length penalties.}

\item {\textbf{Building diverse reasoning-length data for post-training:} CoT-Valve \cite{CoT-Valve} synthesizes diverse-length thought chain data through data interpolation. TOPS samples corresponding versions using a budget-sensitive data model, and C3oT \cite{C3oT} compresses the original LLM output to obtain shorter Short-CoT and trains them jointly.}

\item {\textbf{Using external models or switching mechanisms \cite{Reasoning-in-Flux} to intervene in the reasoning scope:} Routellm \cite{RouteLLM} introduces multiple routers to find the most suitable reasoning model for a given problem, while Self-REF \cite{Self-REF} uses confidence to route the model's reasoning difficulty. Methods like JudgeLRM \cite{JudgeLRM, li2024generation, li2025preference} and DeepSeek-GRM \cite{DeepSeek-GRM} attempt to construct generalized Reward Models for efficient test-time scaling, eliminating irrelevant reasoning paths.}

\item {\textbf{Using efficient representations to execute reasoning:} TokenSkip selects data based on the importance of tokens and executes reasoning using compressed, more concise thought chain representations. COCONUT \cite{COCONUT} attempts to execute more efficient reasoning in the latent space. ICoT-KD \cite{CCoT} and CCoT \cite{CCoT} try to build more efficient reasoning strategies in the hidden space, while Token Assorted combines hidden space reasoning with text-based reasoning to balance interpretability and efficiency. Heima aims to build hidden-space reasoning for multimodal models.}

\item {\textbf{Non-autoregressive reasoning models:} Diffusion-LM avoids a large number of autoregressive prediction steps due to its non-autoregressive nature, enabling efficient reasoning. LLaDA \cite{LLaDA} constructs a diffusion-based LM with an enormous parameter size, achieving scalability based on a non-autoregressive language model architecture. Diffusion-of-Thoughts \cite{Diffusion-of-Thoughts} uses post-training strategies to transform an LLM into a denoising process for token prediction, significantly improving reasoning efficiency.}

\item {\textbf{New architectures for efficient reasoning:} One of the key bottlenecks in long thought-chain prediction is the computational cost of ultra-long context. Mamba \cite{Mamba} and RWKV \cite{RWKV}, among others, use linear attention or state space techniques to effectively enhance the model's reasoning efficiency.}

\end{enumerate}

\section{Challenges \& Future Directions}\label{future}

Despite the rapid advancements in reasoning LLMs, several challenges persist, limiting their generalizability and practical applicability. 
This section outlines these challenges and highlights potential research directions to address them.

\subsection{Efficient Reasoning LLMs}\label{efficient_srs}

While reasoning LLMs excel at solving complex problems via extended inference, their reliance on long autoregressive reasoning within large-scale architectures presents significant efficiency challenges \cite{liu_efficient_survey, foresightsampling_xu}. 
For example, many problems on platforms like Codeforces require over 10,000 tokens of reasoning, resulting in high latency. 
As noted in \cite{Tecent_2_plus_3}, even when a reasoning LLM identifies the correct solution early, it often spends considerable time verifying its reasoning. 
Recent reports, such as Deepseek-R1 \cite{Deepseek-R1}, suggest that self-improvement via RL is more effective in larger models, while smaller-scale large language models (SLMs) (\emph{e.g.}, 3B and 7B models as explored by \cite{simplerl_reason_blob} and \cite{yeo2025demystifying,tinyzero}) struggle to match performance in slow-thinking reasoning tasks.


Future research should focus on two key areas: (1) integrating external reasoning tools to enable early stopping and verification mechanisms, thus improving the efficiency of long inference chains, and (2) exploring strategies to implement slow-thinking reasoning capabilities in SLMs without sacrificing performance.

\subsection{Collaborative Slow \& Fast-thinking Systems}\label{slow-fast_srs}

A key challenge in reasoning LLMs is the loss of fast-thinking capabilities, which results in inefficiencies when simple tasks require unnecessary deep reasoning. 
Unlike humans, who fluidly switch between fast (\textit{System 1}) and slow (\textit{System 2}) thinking, current reasoning LLMs struggle to maintain this balance. 
While reasoning LLMs ensure deliberate and thorough reasoning, fast-thinking systems rely on prior knowledge for quick responses. 
Despite efforts such as the \textit{System 1-2} switcher \cite{VisualSlowAgent}, speculative decoding \cite{fast_speculative_slow_thinking, ning2023skeleton_of_thought, SpecInfer}, and interactive continual learning \cite{qi2024interactive, litesearch, mix_tree}, integrating both modes of thinking remains challenging. 
This often leads to inefficiencies in domain-specific tasks and underutilized strengths in more complex scenarios.

Future research should focus on developing adaptive switching mechanisms, joint training frameworks, and co-evolution strategies to harmonize the efficiency of fast-thinking systems with the precision of reasoning LLMs. 
Achieving this balance is crucial for advancing the field and creating more versatile AI systems.

\subsection{Reasoning LLMs For Science}\label{src_science}

Reasoning LLMs play a crucial role in scientific research \cite{zheng2024openresearcher, sun2025ScienceBoard}, enabling deep, structured analysis that goes beyond the heuristic-based fast-thinking models. 
Their value becomes especially clear in fields that demand complex reasoning, such as medicine and mathematics. 
In medicine, particularly in differential diagnosis and treatment planning, reasoning LLMs (\emph{e.g.}, inference-time scaling) enhance AI's step-by-step reasoning, improving diagnostic accuracy where traditional scaling methods fall short \cite{huang2025o1}. 
In mathematics, approaches like FunSearch \cite{romera2024mathematical} incorporate slow-thinking principles to push beyond previous discoveries, showcasing the potential of AI-human collaboration.

Beyond these fields, reasoning LLMs can foster advancements in physics, engineering, and computational biology by refining model formulation and hypothesis testing. 
Investing in reasoning LLMs research not only bridges the gap between AI's computational power and human-like analytical depth but also paves the way for more reliable, interpretable, and groundbreaking scientific discoveries.

\subsection{Deep Integration of Neural and Symbolic Systems}\label{llm_symbol}

Despite significant advancements in reasoning LLMs, their limited transparency and interpretability restrict their performance in more complex real-world reasoning tasks. 
The reliance on large-scale data patterns and lack of clear reasoning pathways makes it challenging to handle intricate or ambiguous problems effectively. 
Early symbolic logic systems, while less adaptable, offered better explainability and clearer reasoning steps, leading to more reliable performance in such cases.

A promising future direction is the deep integration of neural and symbolic systems. Google's AlphaGeometry \cite{trinh2024solving} and AlphaGeometry2 \cite{chervonyi2025gold} combine reasoning LLMs with symbolic engines, achieving breakthroughs in the International Olympiad in Mathematics (IMO). 
In particular, AlphaGeometry2 utilizes the Gemini-based model \cite{team2023gemini, team2024gemini, gemini2.0-pro} and a more efficient symbolic engine, improving performance by reducing rule sets and enhancing key concept handling \cite{LEGO-Prover, TRIGO, DS-Solver}. 
The system now covers a broader range of geometric concepts, including locus theorems and linear equations. 
A new search algorithm and knowledge-sharing mechanism accelerate the process. 
This system solved 84\% of IMO geometry problems (2000-2024), surpassing gold medalists' averages. 
In contrast, reasoning LLMs like OpenAI-o1 \cite{openai_o1} failed to solve any problems. 
The integration of neural and symbolic systems offers a balanced approach, improving both adaptability and interpretability, with vast potential for complex real-world reasoning tasks beyond mathematical geometry problems.

\subsection{Multilingual Reasoning LLMs}\label{mlan-srs}

Current reasoning LLMs perform well in high-resource languages like English and Chinese, demonstrating strong capabilities in tasks such as translation and various reasoning tasks \cite{DRT-o1, Marco_o1, zhang2024shifcon}. 
These models excel in environments where large-scale data and diverse linguistic resources are available. 
However, their performance in low-resource languages remains limited \cite{MGSM}, facing challenges related to data sparsity, stability, safety, and overall performance. 
These issues hinder the effectiveness of reasoning LLMs in languages that lack substantial linguistic datasets and resources.

Future research should prioritize overcoming the challenges posed by data scarcity and cultural biases in low-resource languages. 
Innovations such as parameter sharing across reasoning LLMs and the incremental injection of domain-specific knowledge could help mitigate these challenges, enabling faster adaptation of slow-thinking capabilities to a broader range of languages. 
This would not only enhance the effectiveness of reasoning LLMs in these languages but also ensure more equitable access to advanced AI technologies.




\subsection{Safe Reasoning LLMs}\label{srs_safety}

The rapid development of reasoning LLMs like OpenAI-o1 \cite{openai_o1} and DeepSeek-R1 \cite{Deepseek-R1} has led to the rise of superintelligent models capable of continuous self-evolution. 
However, this progress brings challenges in safety and control \cite{mei2024slang, mei2024hiddenguard, mei2024not}. 
RL, a key training method, introduces risks such as reward hacking, generalization failures, and language mixing, which can lead to harmful outcomes. 
Ensuring the safety of such systems like DeepSeek-R1 is urgent. While RL enhances reasoning, its uncontrollable nature raises concerns about safely guiding these models. 
SFT addresses some issues but is not a complete solution. 
A hybrid approach combining RL and SFT is needed to reduce harmful outputs while maintaining model effectiveness \cite{parmar2025challenges, GuardReasoner}.

As these models surpass human cognitive capabilities, ensuring their safe, responsible, and transparent use is crucial. 
This requires ongoing research to develop methods for controlling and guiding their actions, thereby balancing AI power with ethical decision-making.





\section{Conclusion}\label{conclusion}

This paper presents a comprehensive survey that advances research on reasoning LLMs. 
We begin with an overview of the progress in foundational LLMs and key early \textit{System 2} technologies, including symbolic logic, MCTS, and RL, exploring how each, when combined with foundational LLMs, has paved the way for reasoning LLMs. 
We then provide a detailed feature analysis of the latest reasoning LLMs, examining the core methods that enable their advanced reasoning capabilities and highlighting representative models. 
Through a review of mainstream reasoning benchmarks and performance comparisons, we offer valuable insights into the current state of the field. 
Looking ahead, we identify promising research directions and continue to track developments via our real-time \href{https://github.com/zzli2022/Awesome-Slow-Reason-System}{GitHub Repository}. 
This survey aims to inspire innovation and foster progress in the rapidly evolving field of reasoning LLMs.

\bibliography{reference}
\bibliographystyle{IEEEtran}



\vfill

\end{document}